\documentclass{article}
\usepackage{microtype}
\usepackage{graphicx}
\usepackage{booktabs}
\usepackage{hyperref}

\usepackage{arxiv_icml2023}

\usepackage{amsmath}
\usepackage{amssymb}
\usepackage{mathtools}
\usepackage{amsthm}
\usepackage{color}
\usepackage{comment}
\usepackage{wrapfig}
\usepackage{bm}
\usepackage[capitalize,noabbrev]{cleveref}
\usepackage{amsfonts}       
\usepackage{nicefrac}       
\usepackage{microtype}      
\usepackage[labelformat=simple]{subcaption}

\theoremstyle{plain}
\newtheorem{theorem}{Theorem}[section]
\newtheorem{proposition}[theorem]{Proposition}
\newtheorem{lemma}[theorem]{Lemma}

\theoremstyle{definition}
\newtheorem{definition}[theorem]{Definition}
\newtheorem{assumption}[theorem]{Assumption}
\theoremstyle{remark}

\newcommand{\req}[1]{Eq.~(\ref{#1})}
\newcommand{\rfig}[1]{Fig.~\ref{#1}}
\newcommand{\rtab}[1]{Tab.~\ref{#1}}

\newcommand{\argmax}{\mathrm{arg}\!\max}

\newcommand{\SCE}{\ell_{\mathrm{CE}}}
\usepackage[textsize=tiny]{todonotes}

\icmltitlerunning{One-vs-the-Rest Loss to Focus on Important Samples in Adversarial Training}

\begin{document}

\twocolumn[
  \icmltitle{One-vs-the-Rest Loss to Focus on Important Samples in Adversarial Training}

\begin{icmlauthorlist}
\icmlauthor{Sekitoshi Kanai}{yyy}
\icmlauthor{Shin'ya Yamaguchi}{yyy}
\icmlauthor{Masanori Yamada}{yyy}
\icmlauthor{Hiroshi Takahashi}{yyy}
\icmlauthor{Kentaro Ohno}{yyy}
\icmlauthor{Yasutoshi Ida}{yyy}
\end{icmlauthorlist}

\icmlaffiliation{yyy}{NTT}

\icmlcorrespondingauthor{Sekitoshi Kanai}{sekitoshi.kanai@ntt.com}
\icmlkeywords{Machine Learning, Adversarial Robustness, Deep Neural Networks}

\vskip 0.3in
]

\printAffiliationsAndNotice{}
\begin{abstract}
    This paper proposes a new loss function for adversarial training. 
    Since adversarial training has difficulties, e.g., necessity of high model capacity, 
    focusing on important data points by weighting cross-entropy loss has attracted much attention.
    However, they are vulnerable to sophisticated attacks, e.g., Auto-Attack.
    This paper experimentally reveals that the cause of their vulnerability is
    their small margins between logits for the true label and the other labels.
    Since neural networks classify the data points based on the logits, 
    logit margins should be large enough to avoid flipping the largest logit by the attacks.
    Importance-aware methods do not increase logit margins of important samples but
    decrease those of less-important samples compared with cross-entropy loss.
    To increase logit margins of important samples,  
    we propose switching one-vs-the-rest loss (SOVR), which switches from cross-entropy to one-vs-the-rest loss for important samples that have small logit margins.
    We prove that one-vs-the-rest loss increases logit margins two times larger than the weighted cross-entropy loss for a simple problem.
    We experimentally confirm that SOVR 
    increases logit margins of important samples unlike existing methods and
    achieves better robustness against Auto-Attack than importance-aware methods. \looseness=-1
  \end{abstract}
    \section{Introduction}
    For multi-class classification problems, deep neural networks have become the de facto standard method in this decade.
    They classify a data point into the label that has the largest logit, which is the input of a softmax function.
    However, the largest logit is easily flipped, and deep neural networks can misclassify slightly perturbed data points, which are called adversarial examples~\citep{szegedy2013intriguing}.
    Various methods have been presented to search the adversarial examples, and Auto-Attack~\citep{AutoAttack} is one of the most successful methods at finding the worst-case attacks.
    For trustworthy deep learning applications, classifiers should be robust against the worst-case attacks.
    To improve the robustness, many defense methods have also been presented~\citep{pgd,pgd2,Wang2020Improving,cohen2019certified}. 
     Among them, adversarial training is a promising method, which empirically achieves good robustness~\citep{carmon2019unlabeled,pgd,pgd2}. 
    However, adversarial training is more difficult than standard training,
    e.g., it requires higher sample complexity~\citep{schmidt2018adversarially,MART} and model capacity~\citep{GAIRAT}.
   
    To address these difficulties, several methods focus on the difference in importance of data points~\citep{MART,MAIL,GAIRAT}. 
    These studies hypothesize that data points closer to a decision boundary are
    more important for adversarial training~\citep{MART, GAIRAT,MAIL}.
    To focus on such data points,
    GAIRAT~\citep{GAIRAT} and MAIL~\citep{MAIL} use weighted softmax cross-entropy loss, which controls weights on the losses on the basis of the closeness to the boundary.
    As the measure of the closeness, GAIRAT uses the least number of steps at which the iterative attacks make models misclassify the data point.
    On the other hand, MAIL uses the measure based on the softmax outputs.
    However, these importance-aware methods tend to be more vulnerable to the robustness against Auto-Attack than na\"ive adversarial training.  
    Thus, it is still unclear \textit{whether focusing on the important training data points enhances the robustness in adversarial training}. \looseness=-1  
 
    To answer this question, we investigate the cause of the vulnerability of importance-aware methods via margins between logits for the true label and the other labels.
    Since neural networks classify the data into the largest logit class,
    logit margins should be large enough to avoid the flipping of the largest logit class by the attacks.
    Through the histogram of logit margins, we first confirmed that there is actually a difference in the data points when using na\"ive adversarial training (AT).
    AT has two peaks in the histogram, i.e., large and small logit margins.  
    This indicates that there are \textit{difficult samples} of which logit margins are difficult to be increased, and they correspond to \textit{important samples}
    since small logit margins indicate that data points are near the decision boundary.
    Next, we found that importance-aware methods only have one peak of small logit margins near zero in the histogram.
    Thus, importance-aware methods do not increase logit margins of important samples but decrease the logit margins of easy (less-important) samples.
    As a result, importance-aware methods are more vulnerable near less-important samples than AT.
    This implies that weighting cross-entropy used in importance-aware methods is not very effective strategy for focussing on important samples.
     
    To increase the logit margins of important samples, we propose switching one-vs-the-rest loss (SOVR), which switches between cross-entropy and one-vs-the-rest loss (OVR) for less-important and important samples, instead of weighting cross-entropy.
    We prove that OVR is always greater than or equal to cross-entropy on any logits.
    Furthermore, we theoretically derive the trajectories of logit margin losses in minimizing OVR and cross-entropy by using gradient flow on a simple problem.
    These trajectories reveal that OVR increases logit margins two times larger than weighted cross-entropy losses after sufficient training time. 
    Thus, SOVR increases logit margins of important samples while it does not decrease logit margins of less-important, unlike importance-aware methods.
    Experiments demonstrate that SOVR increases logit margins more than AT
    and outperforms AT, GAIRAT~\citep{GAIRAT}, MAIL~\citep{MAIL}, MART~\citep{MART}, MMA~\citep{MMA}, and EWAT~\citep{EWAT} in terms of robustness against Auto-Attack.
    In addition, we find that SOVR achieves a better trade-off than TRADES, and the combination of SOVR and TRADES achieves the best robustness. 
    Thus, focusing on important data points improves the robustness when increasing their logit margins by OVR.
    Furthermore, our method improves the performance of other recent optimization methods~\citep{AWP,SEAT} and data augmentation using 1M synthetic data~\citep{rebuffi2021fixing,gowal2021generated} for adversarial training. \looseness=-1
     \section{Preliminaries}
     \subsection{Adversarial Training}\label{ATSubsec}
    Given $N$ data points $\bm{x}_n\!\in\! \mathbb{R}^d$ and class labels $y_n\!\in\!\{1,\dots, K\}$, adversarial training~\citep{pgd2} attempts to solve the following minimax problem with respect to the model parameter $\bm{\theta}\!\in\!\mathbb{R}^m$:\looseness=-1
     \begin{align}\textstyle
      &\textstyle\min_{\bm{\theta}}\mathcal{L}_{\mathrm{AT}}(\bm{\theta})=\textstyle \min_{\bm{\theta}}\frac{1}{N}\sum_{n=1}^{N}\ell_{\mathrm{CE}}(\bm{z}(\bm{x}_n^\prime,\!\bm{\theta}),\!y_n),\label{AT}\\
      &\textstyle\bm{x}_n^\prime\textstyle=\bm{x}_n\!+\!\bm{\delta}_n\nonumber\\
      &~~~\textstyle=\!\bm{x}_n\!+\!\argmax_{|\!|\bm{\delta}_n|\!|_p\leq \varepsilon} \ell_{\mathrm{CE}}(\bm{z}(\bm{x}_n\!+\!\bm{\delta}_n,\bm{\theta}),\!y_n),\label{AdvEx}
    \end{align}
    where $\bm{z}(\bm{x},\bm{\theta})\!=\![z_1(\bm{x},\bm{\theta}),\dots,z_K(\bm{x},\bm{\theta})]^T$ and
     $z_k(\bm{x},\bm{\theta})$ is the $k$-th logit of the model, which is input of softmax:$f_k(\bm{x},\bm{\theta})\!=\!e^{z_k(\bm{x})}\!/\!\sum_i e^{z_i(\bm{x})}$.
    $\ell_{\mathrm{CE}}$ is a cross-entropy function, and 
    $||\cdot||_p$ and $\varepsilon$ are $L_p$ norm and the magnitude of perturbation $\bm{\delta}_n\!\in\!\mathbb{R}^d$, respectively.
    The inner maximization problem is solved by projected gradient descent (PGD)~\citep{pgd,pgd2}, which 
     updates the adversarial examples as
     \begin{align}\textstyle
      \!\!\!\!\bm{\delta }_t\!=\! \Pi_{\varepsilon}\!\left(\bm{\delta}\!+\!\eta\mathrm{sign}\left(\nabla_{\bm{\delta}_{t-1}}\!\ell_{\mathrm{CE}}\left(\bm{z}(\bm{x}\!+\!\bm{\delta}_{t-1},\bm{\theta}), y\right)\!\right)\!\right),
     \end{align}
     for $\mathcal{K}$ steps where $\eta$ is a step size. $\Pi_{\varepsilon}$ 
     is a projection operation into the feasible region $\{\bm{\delta}~|~\bm{\delta}\in \mathbb{R}^d, ||\bm{\delta}||_p\leq \varepsilon\}$.
     Note that we focus on $p\!=\!\infty$ since it is a common setting.
     For trustworthy deep learning, we should improve the true robustness: the robustness against the worst-case attacks in the feasible region.
    Thus, the evaluation of robustness should use crafted attacks, e.g., Auto-Attack~\citep{AutoAttack}, since PGD often fails to find the adversarial examples misclassified by models. \looseness=-1
    
     \subsection{Importance-aware Adversarial Training}
    GAIRAT (geometry aware instance reweighted adversarial training)~\citep{GAIRAT} and 
    MAIL (margin-aware instance reweighting learning)~\citep{MAIL} regard data points closer to the decision boundary of model $\bm{f}$ as important samples
    and assign higher weights to the loss for them: \looseness=-1
    \begin{align}\label{WeightedLoss}\textstyle
      \mathcal{L}_{\mathrm{weight}}(\bm{\theta})=\textstyle\frac{1}{N}\sum_{n=1}^N \bar{w}_n\ell_{\mathrm{CE}}(\bm{z}(\bm{x}_n^\prime,\bm{\theta}),\!y_n), 
    \end{align}
    where $\bar{w}_n\!\geq\!0$ is a weight normalized as $ \bar{w}_n\!=\!\frac{w_n}{\sum_l{w}_l}$ and $\sum_n \bar{w}_n\!=\!1$.
    GAIRAT determines the weights through the $w_n\!=\!\frac{1\!+\!\mathrm{tanh}(\lambda+5(1-2\kappa_n/\mathcal{K}))}{2}$
    where $\kappa_n$ is the least steps at which PGD succeeds at attacking models, and $\lambda$ is a hyperparameter.
    On the other hand, MAIL uses $w_n\!=\!\mathrm{sigmoid}(-\gamma (PM_n\!-\!\beta))$ where $PM_n\!=\!f_{y_n}(\bm{x}_n^\prime,\bm{\theta})\!-\!\max_{k\neq {y_n}}f_k(\bm{x}_n^\prime,\bm{\theta})$.
    $\beta$ and $\gamma$ are hyperparameters.
    MART (misclassification aware adversarial training)~\citep{MART} uses a similar approach.
    It regards misclassified samples as important samples and controls the difference between the loss on less-important and important samples. 
    MMA (max-margin adversarial training)~\citep{MMA} also adaptively changes the loss function and $\varepsilon$ for each data point,
    and thus, MMA also has a similar effect to the above methods. We collectively call the above methods \textit{importance-aware methods}.
    \subsection{Vulnerability of Importance-aware Methods}\label{AASec}
\begin{figure}
      \centering
      \begin{subfigure}[t]{0.187\linewidth}
        \centering
        \includegraphics[width=\linewidth]{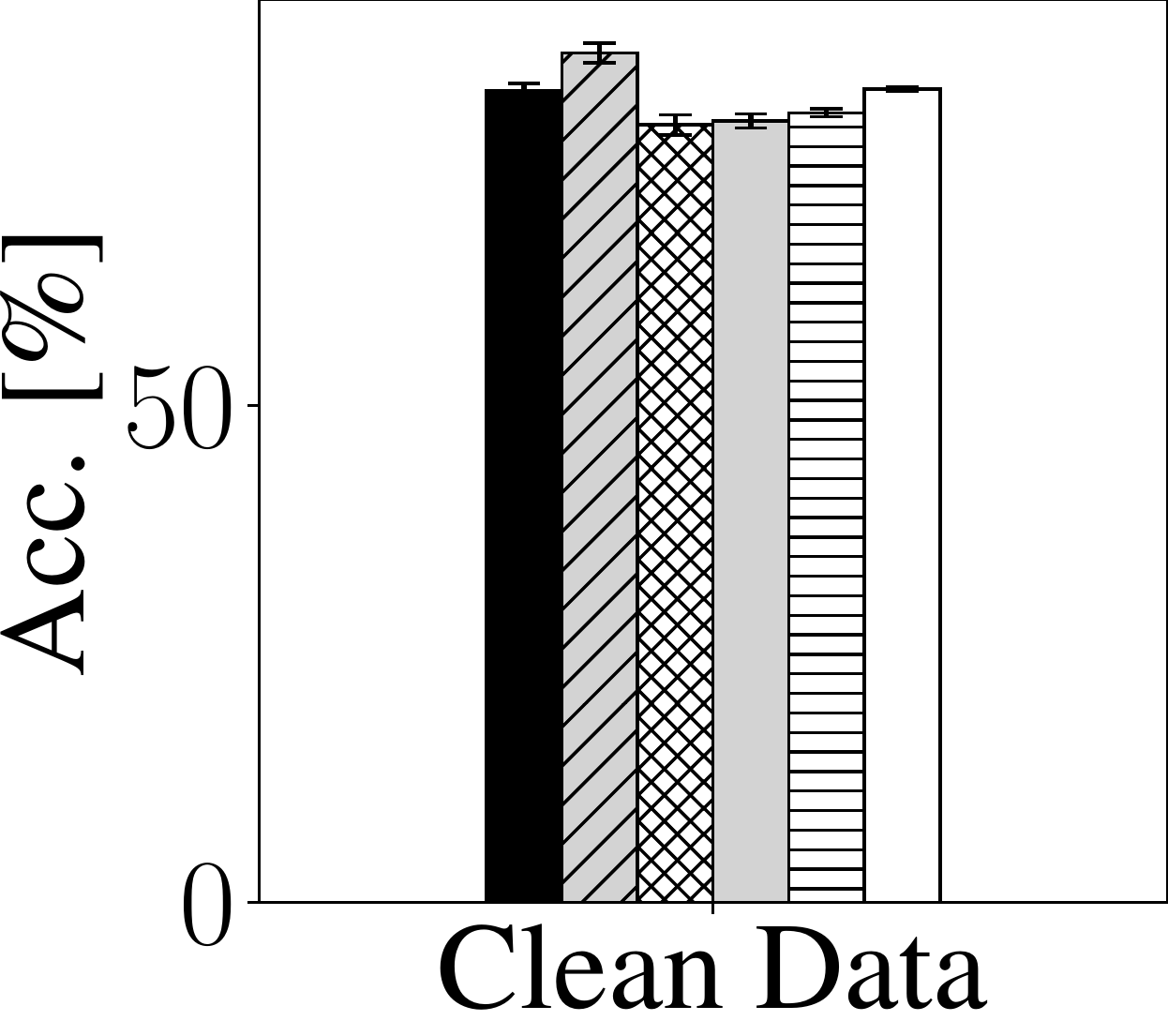}
        \end{subfigure}
        \begin{subfigure}[t]{0.8\linewidth}
          \centering
          \includegraphics[width=\linewidth]{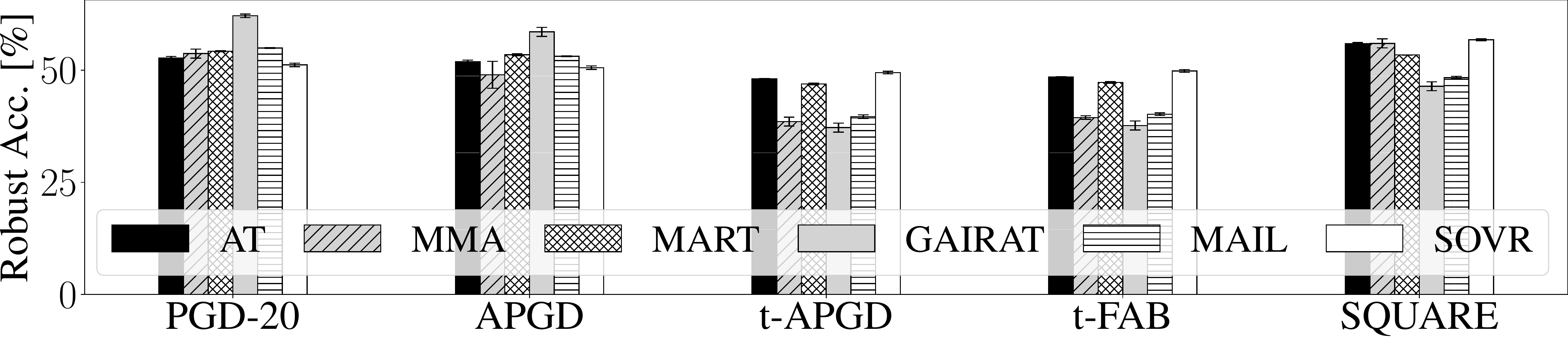}
        \end{subfigure}
      \caption{Robustness against PGD and components of Auto-Attack on CIFAR10~\citep{cifar} with PreActResNet18 (RN18). SOVR is our proposed method.} 
      \label{ATDis}
    \end{figure}

    \citet{hitaj2021evaluating,AutoAttack,EWAT} have reported that the robust accuracies of GAIRAT, MART, and MMA are lower than na\"ive adversarial training when using logit scaling attacks or Auto-Attack~\citep{AutoAttack}.
    Since Auto-Attack searches adversarial examples by using various attacks,
    it achieves a larger success attack rate than using one attack, e.g., PGD. 
    To clarify the vulnerabilities, 
    we individually evaluate the robustness against the components of Auto-Attack and PGD ($\mathcal{K}\!=\!20$) on CIFAR10 (\rfig{ATDis}).
    The training setup is the same as in Section~\ref{ExSec}, and we add the results of our method (SOVR) as a reference.
    This figure shows that almost all importance-aware methods can improve the robustness against PGD and APGD compared with na\"ive adversarial training (AT~\citep{pgd2}).
    However, they do not improve the true robustness; i.e., their robust accuracies against the worst-case attack are lower than that of AT.
    We investigate the reasons of this vulnerability in the next section.
    \vspace{-2pt}
    \section{Evaluation of Robustness via Logit Margin}\label{vulsec}
    We investigate the causes of the vulnerabilities of importance-aware methods by comparing histograms of logit margin losses. 
    First, we explain that logit margin losses determine the robustness. 
    Next, we experimentally reveal that logit margin losses of importance-aware methods concentrate on zero; i.e., their logit margins are smaller than AT.
    We use training data for empirical evaluation in this section
    because the goal of this section is to investigate the effect of importance-aware methods, which modify the loss function based on training data points.
    Experimental setups are provided in Appendix~\ref{ExTochuSec}. \looseness=-1
    \subsection{Potentially Misclassified Data Detected by Logit Margin Loss}\label{lipSec}
    To investigate the robustness of models near each data point, we apply \textit{logit margin loss}~\citep{MMA} to the models trained by importance-aware methods. 
    Logit margin loss is
    \begin{align}
    \textstyle  \ell_{\mathrm{LM}}(\bm{z}(\bm{x}_n, \bm{\theta}), y)& \textstyle=z_{k^*}(\bm{x}^\prime)-z_{y_n}(\bm{x}^\prime)\nonumber\\
    & \textstyle=\max_{k\neq y}z_{k}(\bm{x}^\prime)-z_{y}(\bm{x}^\prime),
    \end{align}
    where $k^*\!=\!\mathrm{arg}\!\max_{k\neq y}\!z_k(\bm{x}^\prime)$.
    Since the classifier infers the label of $\bm{x}$ as $\hat{y}\!=\!\argmax_k\!z_{k}(\bm{x})$,
    it correctly classifies $\bm{x}^\prime$ if $\ell_{\mathrm{LM}}\leq 0$. Thus, the logit margin loss on a difficult sample in adversarial training takes a value near zero.
    We refer to the absolute value of a logit margin loss $|\ell_{\mathrm{LM}}|$ as \textit{logit margin}.
    In contrast to $\mathrm{PM}_n$ of MAIL, $\ell_{\mathrm{LM}}$ is not bounded since $z_k(\bm{x})$ can take an arbitrary value in $\mathbb{R}$. \looseness=-1
    
    To  explain the effect of logit margins,
    we assume that
    the Lipschitz constant of the $k$-th logit function is $L_k$ as $|z_k(\bm{x}_1)\!-\!z_k(\bm{x}_2)|\!\leq\!L_k||\bm{x}_1-\bm{x}_2||_\infty$. 
    In this case, we have the following inequality:
    \begin{align}\textstyle\label{BoundLip}
    \max_k\!z_{k}(\bm{x}^\prime)\!-\!z_y(\bm{x}^\prime)
    &\!\textstyle\!\leq\!\max_{k}\!\left[\textstyle z_{k} (\bm{x})\!-\!z_y(\bm{x})\!+\!(L_{k}\!+\!L_y)\varepsilon \right]\nonumber\\
    &\!\textstyle\!\leq\!z_{k*} (\bm{x})\!-\!z_y(\bm{x})\!+\!(L_{\hat{k}}\!+\!L_y)\varepsilon,
    \end{align}
    where $\hat{k}=\argmax_{k}L_k$.
    From the above, we define the potentially misclassified sample:
    \begin{definition}\label{potentially}
    If a data point $\bm{x}$ satisfies 
        $z_{k^*}(\bm{x})\!-\!z_y(\bm{x})\!>\!-(L_{\hat{k}}\!+\!L_y)\varepsilon$,
    we call it a \textit{potentially misclassified sample}.
    \end{definition}
    By using the above definition, we can derive the following:
    \begin{proposition}\label{CertProp}
      If data points are not potentially misclassified samples, 
      models are guaranteed to have the certified robustness on them as 
      $y\!=\!\argmax_{k} z_k(\bm{x}+\bm{\delta})$
      for any $\bm{\delta}$ satisfying $||\bm{\delta}||_{\infty}\leq \varepsilon$.
    \end{proposition}\vspace{-5pt}
    All proofs are provided in Appendix~\ref{proofsec}.
    We can estimate the true robustness of each method by counting the number of potentially misclassified samples.
    Definition~\ref{potentially} and Proposition~\ref{CertProp} indicate
    that large logit margins $|\ell_{\mathrm{LM}}|$ or small Lipschitz constants $L_k$ are necessary for the robustness.
    Thus, the logit margin loss can be the metric of robustness, and we evaluate it in Section~\ref{HistSec}.
    In Section~\ref{miscriSec}, we provide the estimated number of potentially misclassified samples for each method. 
    \begin{figure*}[tb]
     \begin{subfigure}[t]{0.16\linewidth}
       \includegraphics[width=\linewidth]{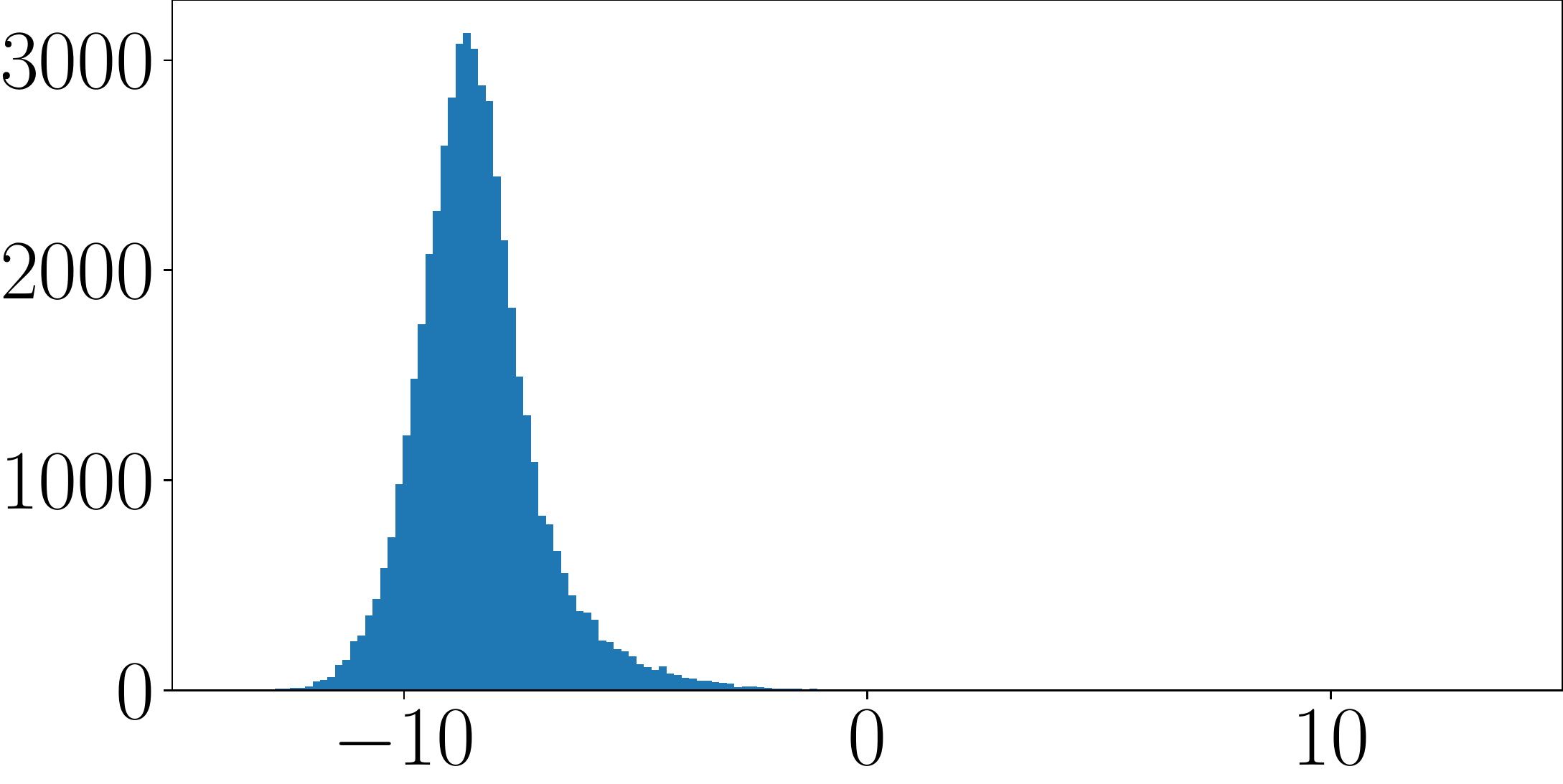}
     \end{subfigure}
     \begin{subfigure}[t]{0.16\linewidth}
       \includegraphics[width=\linewidth]{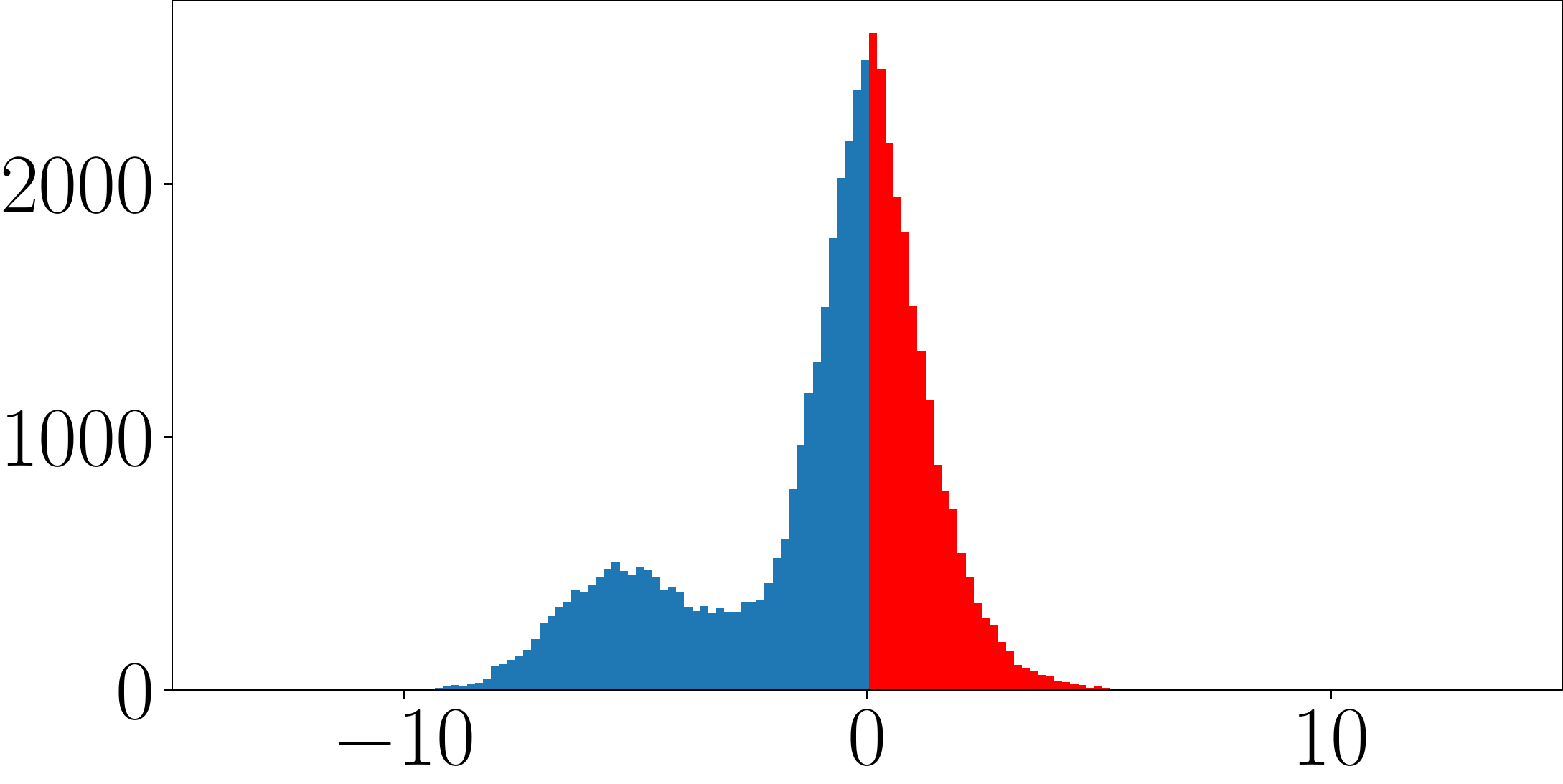}
   \end{subfigure}
   \begin{subfigure}[t]{0.16\linewidth}
     \includegraphics[width=\linewidth]{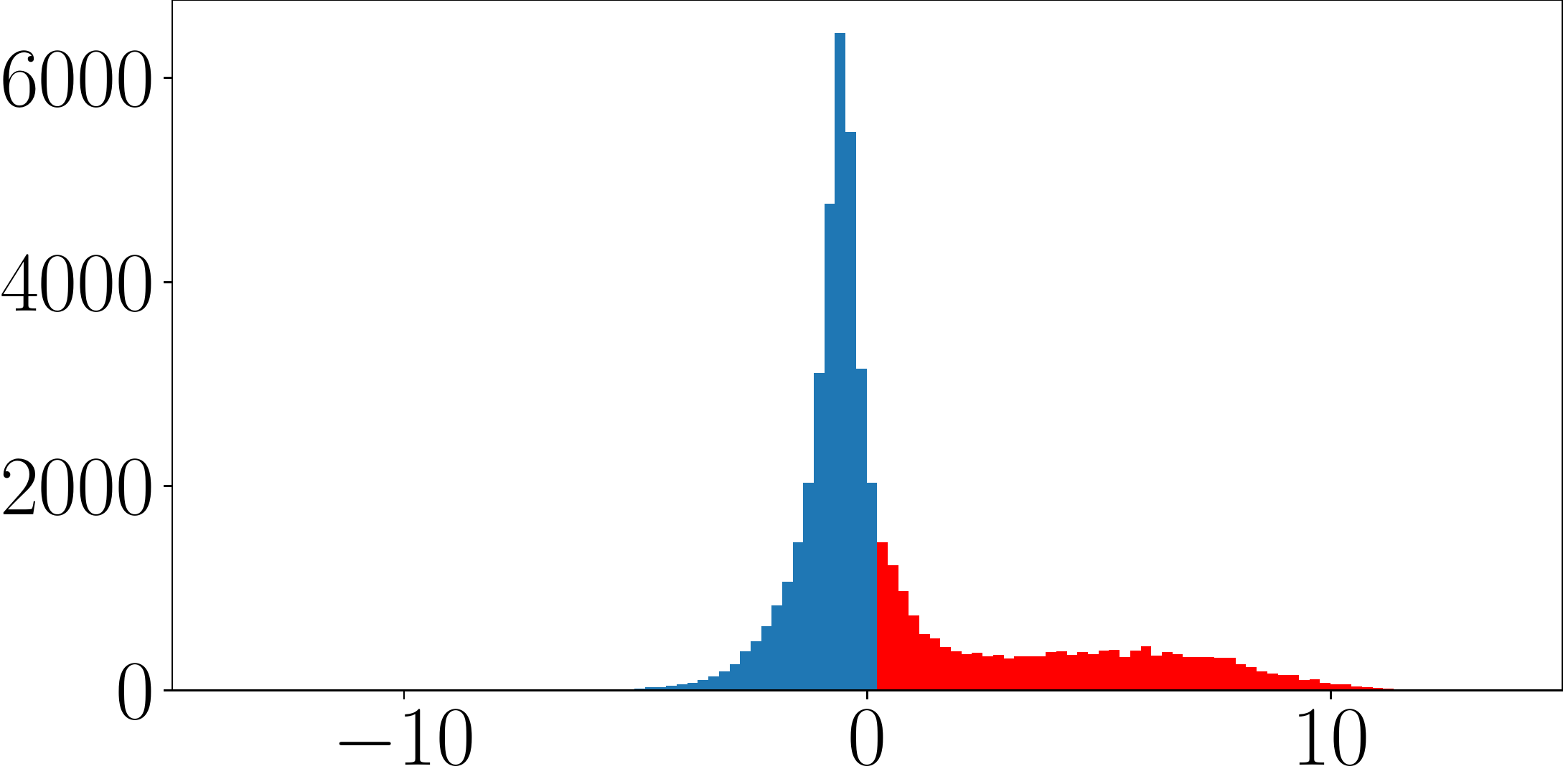}
   \end{subfigure}
   \begin{subfigure}[t]{0.16\linewidth}
     \includegraphics[width=\linewidth]{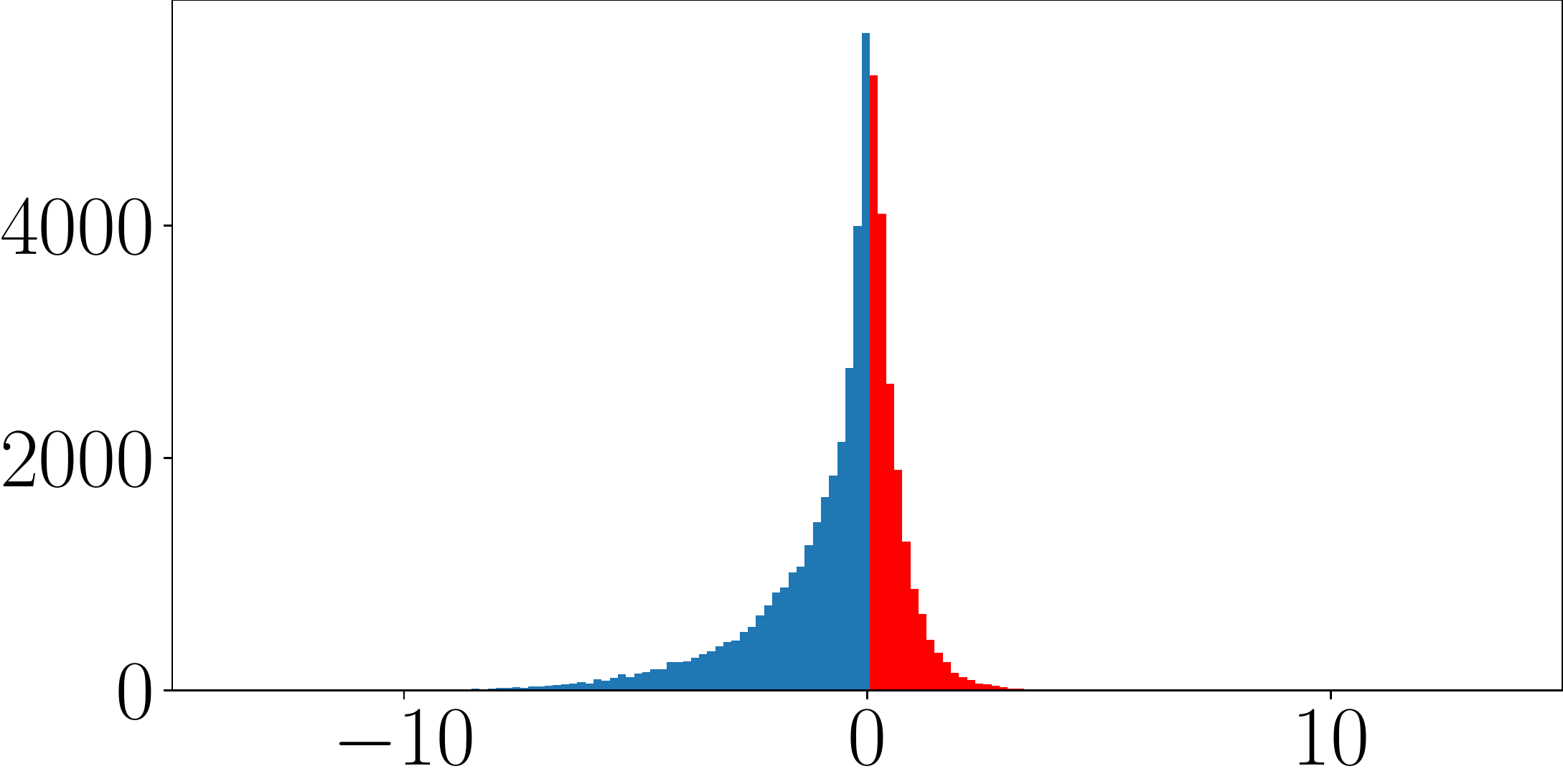}
   \end{subfigure}
   \begin{subfigure}[t]{0.16\linewidth}
     \includegraphics[width=\linewidth]{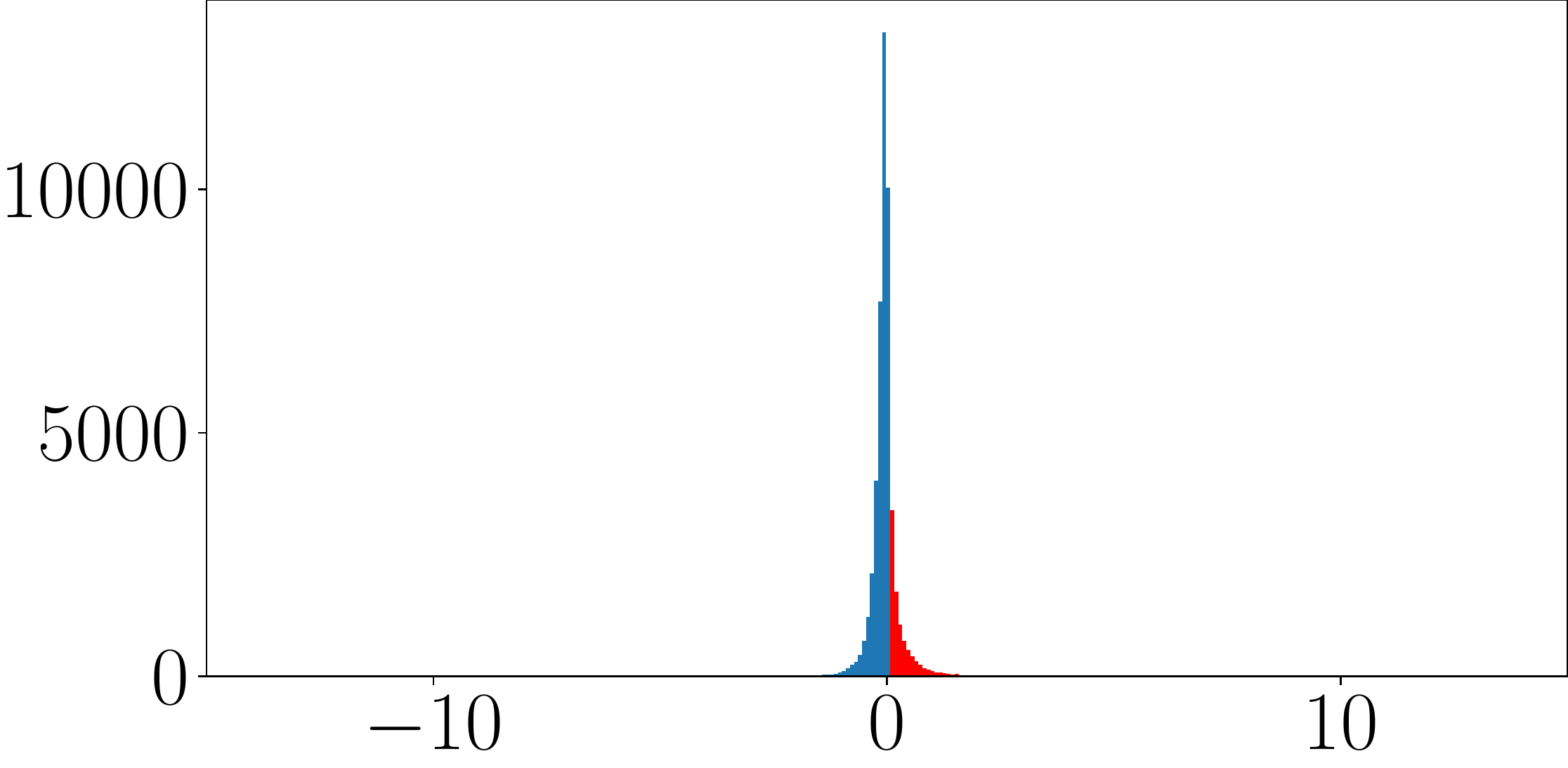}
   \end{subfigure}
   \centering
   \begin{subfigure}[t]{0.16\linewidth}
     \includegraphics[width=\linewidth]{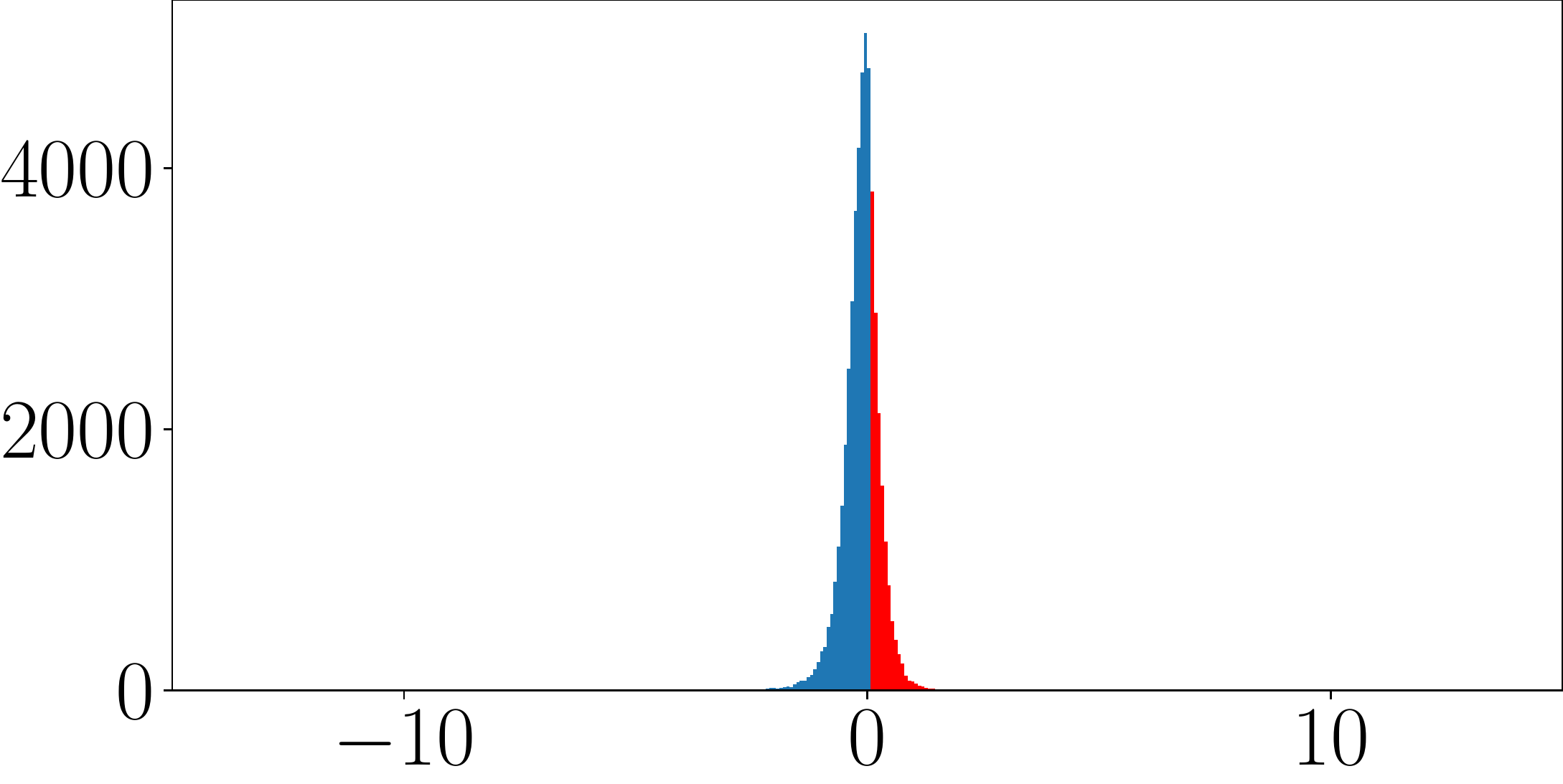}
   \end{subfigure}
     \begin{subfigure}[t]{0.16\linewidth}
       \includegraphics[width=\linewidth]{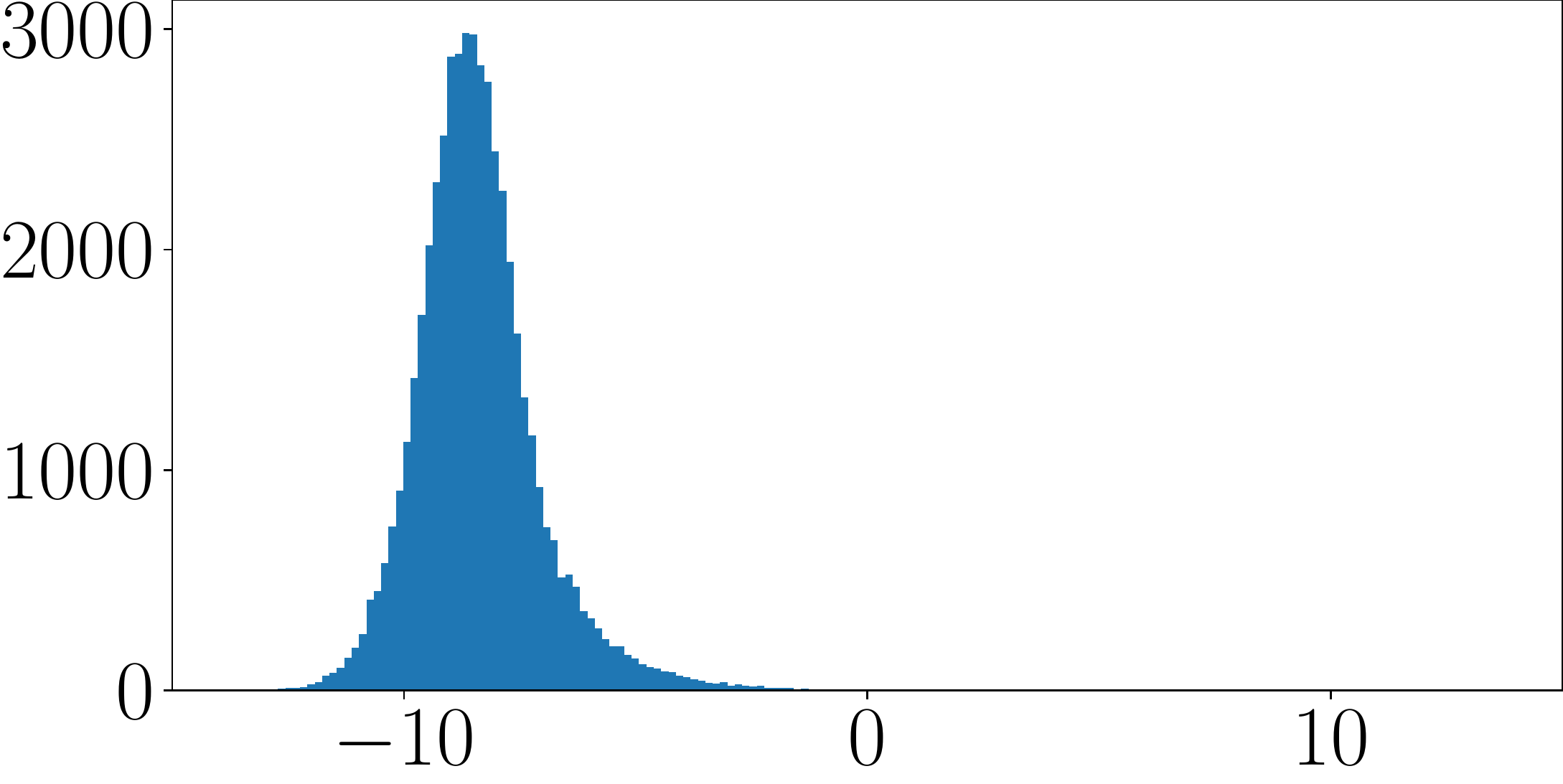}
         \caption{ST}\label{LMHist-ST}
     \end{subfigure}
     \begin{subfigure}[t]{0.16\linewidth}
       \includegraphics[width=\linewidth]{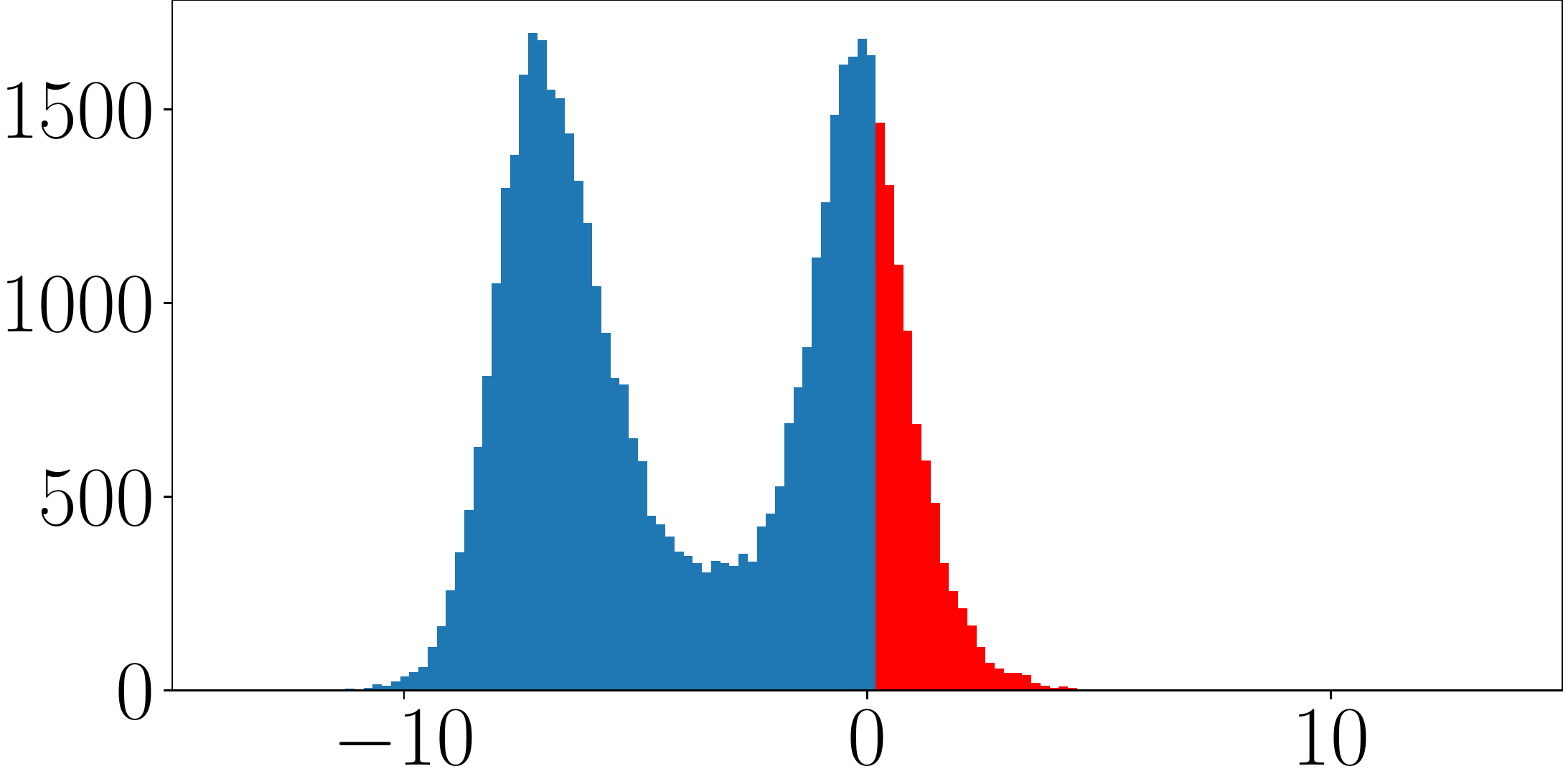}
       \caption{AT}\label{LMHist-AT}
   \end{subfigure}
   \begin{subfigure}[t]{0.16\linewidth}
     \includegraphics[width=\linewidth]{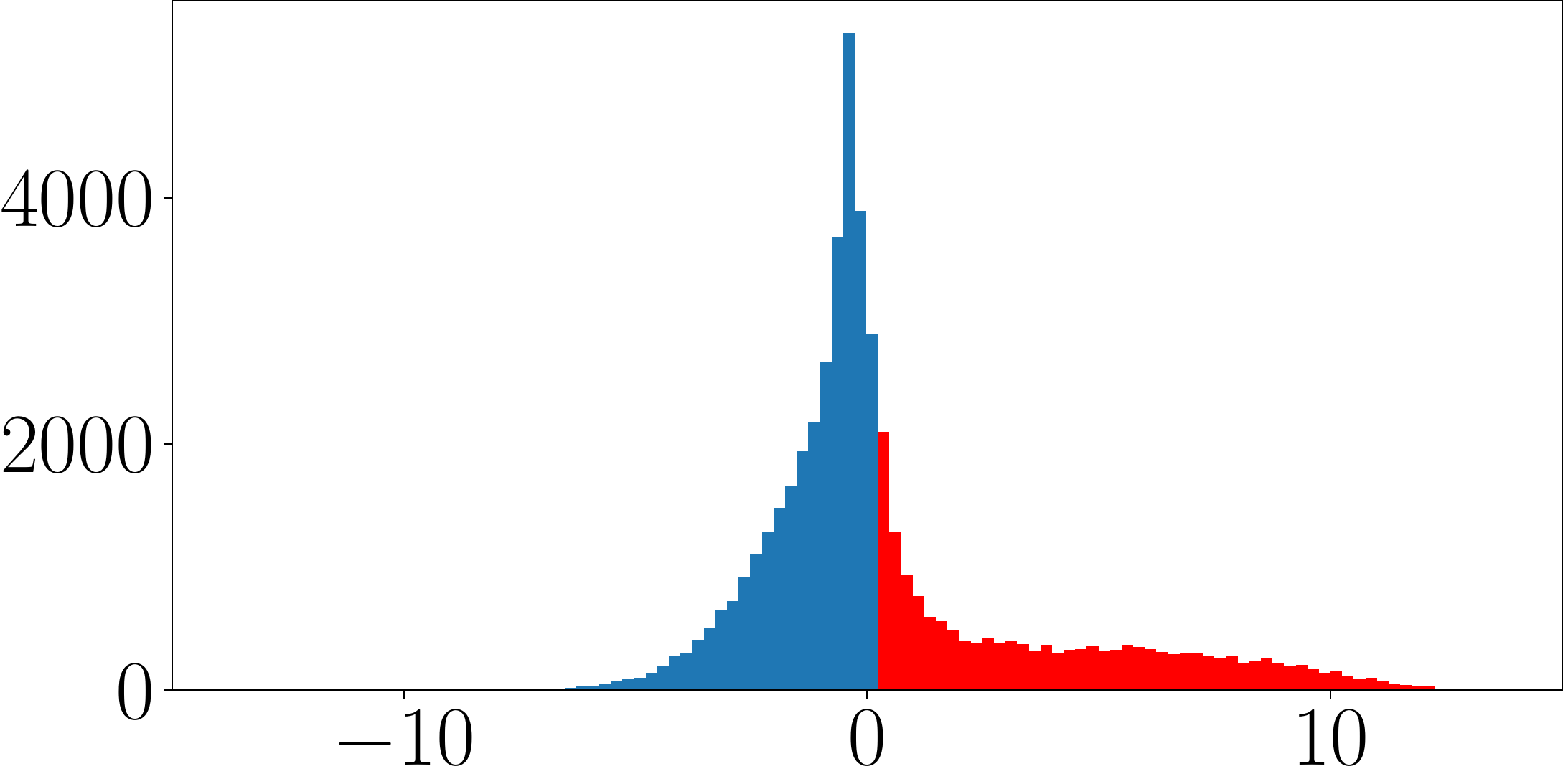}
     \caption{MMA}\label{LMHist-MMA}
   \end{subfigure}
   \begin{subfigure}[t]{0.16\linewidth}
     \includegraphics[width=\linewidth]{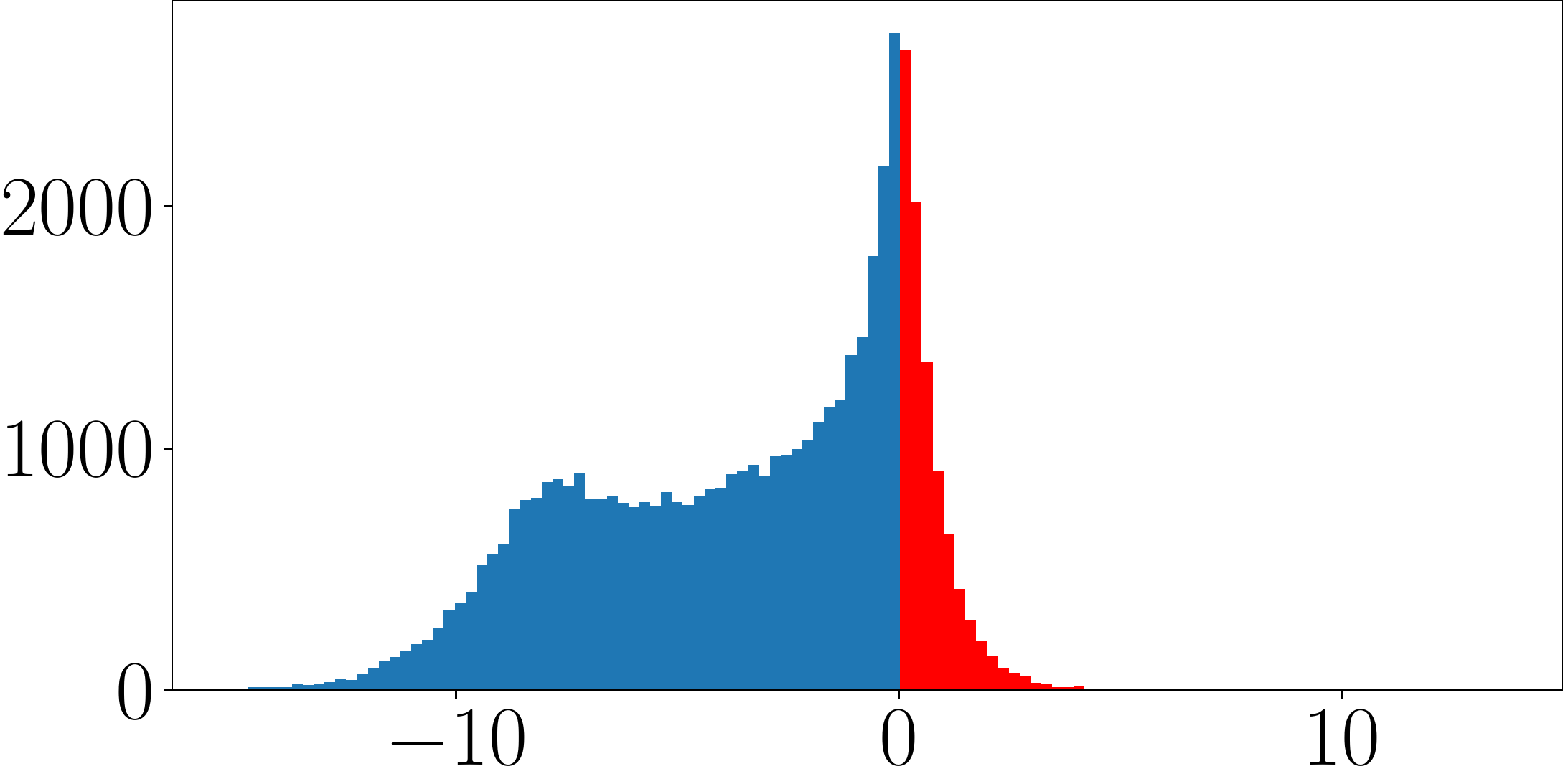}
     \caption{MART}\label{LMHist-MART}
   \end{subfigure}
   \begin{subfigure}[t]{0.163\linewidth}
     \includegraphics[width=\linewidth]{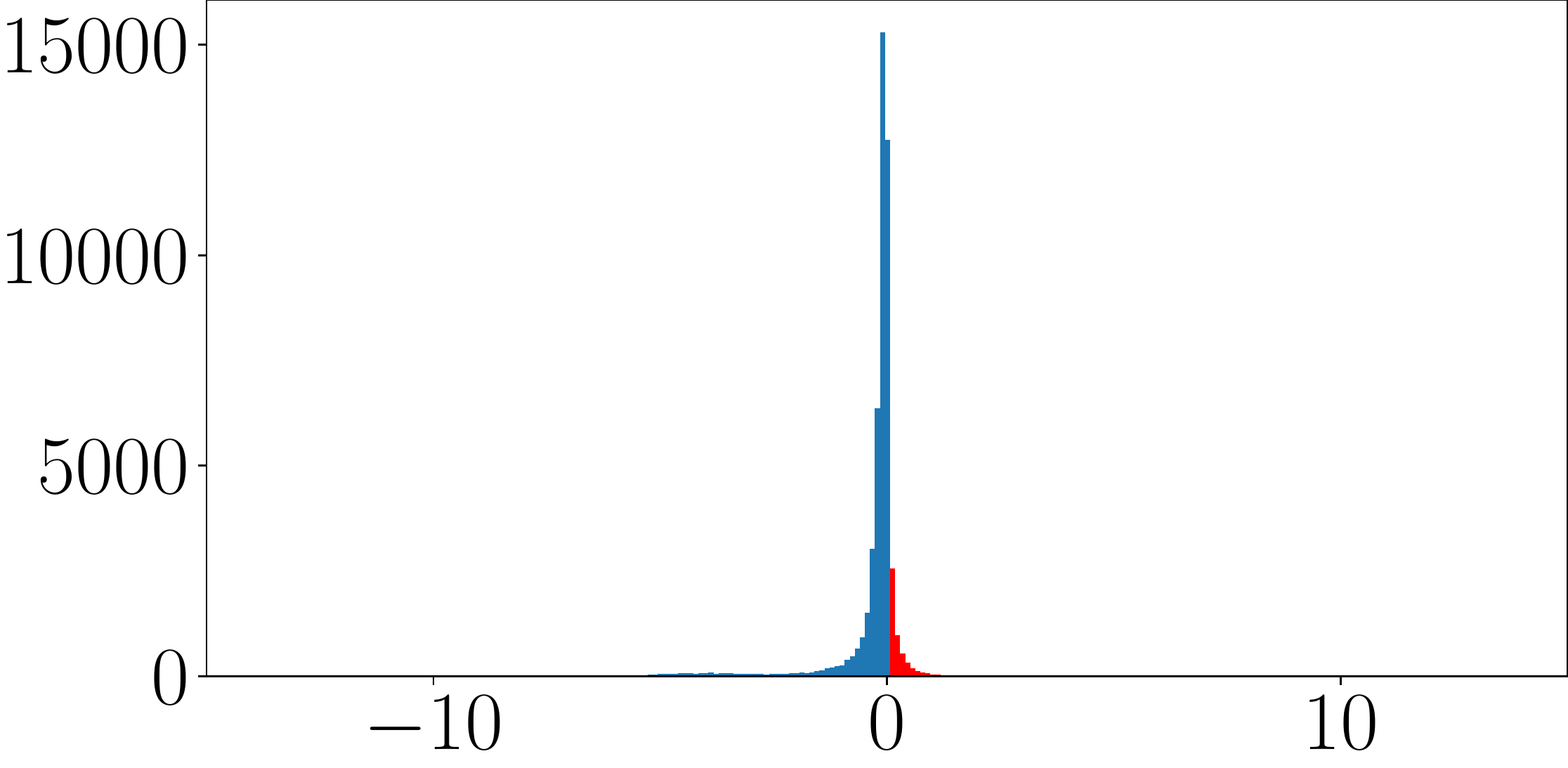}
     \caption{GAIRAT}\label{LMHist-GAIRAT}
   \end{subfigure}
   \begin{subfigure}[t]{0.16\linewidth}
     \includegraphics[width=\linewidth]{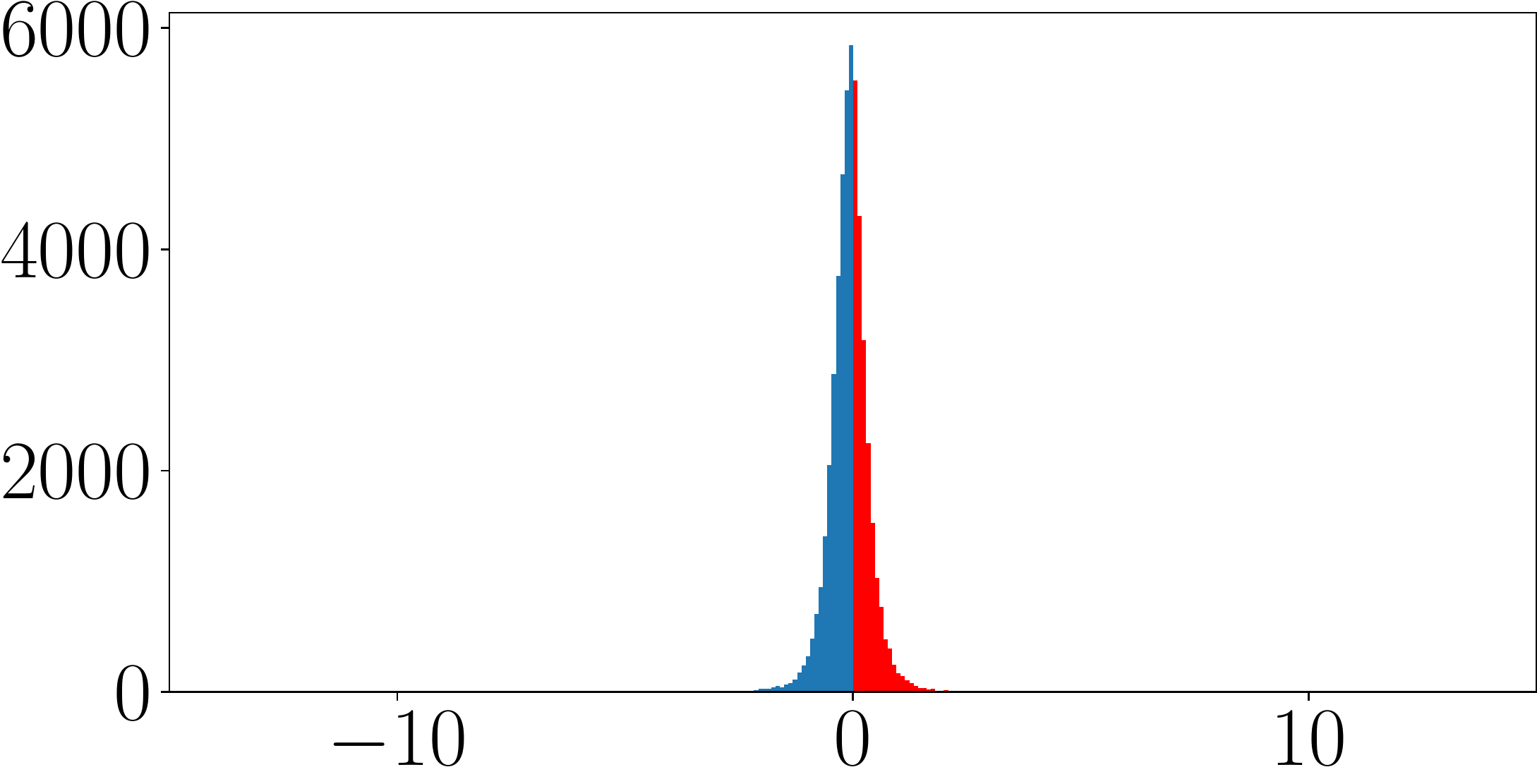}
     \caption{MAIL}\label{LMHist-MAIL}
   \end{subfigure}
   \caption{Histogram of $\ell_{\mathrm{LM}}$ for training data of CIFAR10 with RN18 at the best (top) and the last (bottom) epoch.
   Best epoch is the epoch when models achieved the best robust accuracy against PGD by early stopping.
   ST denotes standard training, i.e., training on clean data.
     For standard training, we use $\ell_{\mathrm{LM}}$ on clean data $\bm{x}$,
     while we plot that on adversarial examples $\bm{x}^\prime$ for the other methods. Blue bins are the correctly classified data points,
     and red bins are misclassified samples.}
     \label{LMHist}
   \end{figure*}
    \subsection{Histograms of Logit Margin Loss}\label{HistSec}
    Since logit margin losses determine the number of potentially misclassified samples,
    we show the histogram of them for each method on CIFAR10 at the last epoch in \rfig{LMHist}.
    Comparing AT (\rfig{LMHist-AT}) with standard training (ST, \rfig{LMHist-ST}), AT has two peaks in the histogram.
    This indicates that the levels of difficulty in increasing the margins in AT are roughly divided into two: \textit{difficult samples} (right peak) and \textit{easy samples} (left peak).
    Difficult samples correspond to the data close to the boundary; i.e., \textit{important samples}.
    Next, comparing AT (\rfig{LMHist-AT}) with importance-aware methods (Figs.~\ref{LMHist-MMA}-\subref{LMHist-MAIL}),
    their logit margin losses $\ell_{\mathrm{LM}}$ concentrate on zero, and their peaks are sharper than that of AT.
    This indicates that importance-aware methods fail to increase the logit margins $|\ell_{\mathrm{LM}}|$ for not only important samples but also less-important (easy) samples
    because the weights for less-important samples are relatively small.
    Thus, it is necessary to increase the small logit margins for important samples without decreasing those of easy samples.
    Appendix~\ref{HisSub} provides results under various settings, which show similar tendencies.\looseness=-1
    \section{Proposed Method}\label{PropSec}
    In Section~\ref{HistSec}, we observe that (a) training samples are roughly divided into two types via logit margins; difficult and easy (important and less-important) samples,
    and (b) importance-aware methods reduce the logit margins on less-important samples to focus on important samples.
    From these observations, our method is based on two ideas:
    (i) we switch from cross-entropy to an alternative loss for important samples by the criterion of the logit margin loss,
    and (ii) the alternative loss increases the logit margins of important samples more than weighted cross-entropy. \looseness=-1
    \subsection{One-Vs-the-Rest Loss (OVR)} 
    The logit margin $|z_{k^*}(\bm{x})-z_{y}(\bm{x})|$ should be large while keeping Lipschitz constants of logit functions small values.
    To this end, we need a loss function to penalize small logit margins.
    The logit margin loss can be an intuitive candidate as such a loss function. 
    However, the logit margin loss only considers the pair of the largest logit $z_{k^*}$ and the logit for the true label $z_{y}$,
    and this is not sufficient for the robustness because $k^*$ and $\hat{k}$ in \req{BoundLip} are not necessarily the same.
    Moreover, the logit margin loss does not have the desirable property for multi-class classification: infinite sample consistent (ISC)~\citep{SAForMC,bartlett2003convexity,lin2002support}.
    To consider the logits for all classes and satisfy ISC,
    our proposed method uses the one-vs-the-rest loss (OVR):
    \begin{align}\textstyle
      \ell_{\mathrm{OVR}}(\bm{z}(\bm{x},\bm{\theta}),y)=\phi(z_y(\bm{x}))+\sum_{k\neq y}\phi(-z_k(\bm{x})).\label{OVRRaw}
    \end{align}
    When $\phi$ is a differentiable non-negative convex function and satisfies $\phi (z)\!<\!\phi (-z)$ for $z\!>\!0$, OVR satisfies ISC~\citep{SAForMC}.
    To satisfy ISC, we set $\phi(z)\!=\!\mathrm{log}(1\!+\!e^{-z})$ and use \looseness=-1
    \begin{align}\textstyle
    \!\!\ell_{\mathrm{OVR}}(\bm{z}(\bm{x},\bm{\theta}),y)&\textstyle\!=\!\mathrm{log} (1\!+\!e^{-z_{y}(\bm{x})})\!\!+\!\!\sum_{k\neq y}\!\mathrm{log}(1\!+\!e^{z_{k}(\bm{x})})\nonumber\\
    &\textstyle\!=\!-z_{y}(\bm{x})\!+\!\sum_{k}\mathrm{log}(1\!+\!e^{z_{k}(\bm{x})}).\label{OVR}
    \end{align}
    We provide explanation of ISC and the detailed reason for this selection of $\phi(z)$ in Appendix~\ref{PhiSelect}.
    \subsection{Behavior of Logit Margin Losses by OVR}\label{OVRBehavSec}
   To show the effectiveness of OVR in increasing logit margins, we theoretically discuss the difference between OVR and cross-entropy. 
    First, OVR has the following property compared with cross-entropy: \looseness=-1
    \begin{theorem}\label{CEVsOVR}
      If we use OVR (\req{OVR}) and softmax as $f_k(\bm{x})=e^{z_k(\bm{x})}/\sum_i e^{z_i(\bm{x})}$,
       we have 
      \begin{align}\textstyle
        0\leq \ell_{\mathrm{CE}}(\bm{z}(\bm{x}),y)\leq \ell_{\mathrm{OVR}}(\bm{z}(\bm{x}),y),~~\forall (\bm{x},y).
      \end{align}
      When $z_y(\bm{x})\!\rightarrow\!+\infty$ and $z_k(\bm{x})\!\rightarrow\!-\infty$ for $k\!\neq\!y$,
      we have $\ell_{\mathrm{OVR}}(\bm{z}(\bm{x}),y)\!\rightarrow\!0$ and $\ell_{\mathrm{CE}}(\bm{z}(\bm{x}),y)\!\rightarrow\!0$.
    \end{theorem}
    Thus, OVR is always larger than or equal to cross-entropy, and OVR and cross-entropy approach asymptotically to zero when $|\ell_{\mathrm{LM}}|$ grows to infinity.
    In fact, we observed that $\ell_{\mathrm{OVR}}(\bm{z},y)$ is about four times greater than $\SCE(\bm{z},y)$ for random logits $\bm{z}\!\sim\!\mathcal{N}(\bm{0},\bm{I})$ and randomly selected $y$ from $\{1,\dots,10\}$.
    Thus, we expect OVR to penalize the small logit margin more strongly than cross-entropy. \looseness=-1
    
   Besides this general result, we further investigate the effect of OVR in logit margin losses by using a simple problem.
   To analyze the behavior of logit margins, we formulate the following problem:
    \begin{align}\textstyle
     \min_{\bm{z}} w\ell_{*}(\bm{z},y)\label{ToyProb},
    \end{align}
   where
   $\ell_{*}$ is set to $\ell_\mathrm{OVR}$ or $\ell_\mathrm{CE}$.
   $\bm{z}\!\in\! \mathbb{R}^K$ is a logit vector for a data point $\bm{x}$, and we assume that we can directly move it in this problem. 
   $w\!\in\! \mathbb{R}$ is a weight of the loss, which appears in \req{WeightedLoss}. 
    To analyze the dynamics of training on \req{ToyProb}, we use the following assumption.
     \begin{assumption}  \label{Assum}
     A logit vector $\bm{z}$ follows the following gradient flow to solve \req{ToyProb}:
       \begin{align}\label{GF}\textstyle
         \frac{d \bm{z}}{dt}=-\nabla_{\bm{z}} w\ell_{*}(\bm{z},y),
        \end{align}
        where $t$ is a time step of training.
       We assume that $\bm{z}$ is initialized to zeros $\bm{z}=\bm{0}$ at $t=0$. 
   \end{assumption}
 
    Equation~(\ref{GF}) is a continuous approximation of gradient descent $\bm{z}^{\tau\!+\!1}\!=\!\bm{z}^{\tau}\!\!-\!\eta \nabla_{\bm{z}}\!w\ell_{*}$ and matches it in the limit as $\eta\!\rightarrow\!0$.
    It is a commonly used method to analyze the training dynamics~\citep{kunin2020neural,elkabetz2021continuous}.
 
    Under Assumption~\ref{Assum}, we have the following lemmas about the logits in the training of \req{ToyProb}:
    \begin{lemma}\label{ovrlem}
     If we use OVR $\ell_{\mathrm{OVR}}(\bm{z},y)$ in \req{ToyProb}, the $k$-th logit $z_k (t)$ at time $t$ is 
     \begin{align}\textstyle
       z_k(t)=\begin{cases}\textstyle
         wt+1-W(e^{wt+1})&~~k\!=\!y,        \\ \textstyle
         -wt-1-W(e^{wt+1})&~~k\!\neq\!y,
       \end{cases}
     \end{align}
     where $W$ is Lambert $W$ function, which is a function satisfying $x=W(xe^x)$ \citep{Wfunction1}.
   \end{lemma}
    \begin{lemma}\label{celem}
     If we use cross-entropy $\ell_{\mathrm{CE}}(\bm{z},y)$ in \req{ToyProb}, the $k$-th logit $z_k (t)$ at time $t$ is 
     \begin{align}\textstyle
       \!\!z_k(t)\!=\!\begin{cases}\textstyle
        \!\frac{Kwt+1}{K}\!-\!\frac{K-1}{K}W(\frac{1}{K-1}e^{\frac{Kwt+1}{K-1}})&\!\!\!k\!=\!y,       \\ \textstyle
        \!-\frac{Kwt+1}{K(K-1)}\!+\!\frac{1}{K}W(\frac{1}{K-1}e^{\frac{Kwt+1}{K-1}})&\!\!\!k\!\neq\!y,.
       \end{cases}
     \end{align}
   \end{lemma}
   These lemmas give the trajectories of logit vectors in minimization of OVR and cross-entropy, respectively.
   Both methods increase the logit for the true labels $z_y$ and decrease the others, but their speeds are different.
    From the above lemmas, we derive the trajectory of the logit margin loss:\looseness=-1
    \begin{theorem}\label{lmthm}
     Logit margin losses for the logit vector $\bm{z}^{OVR}$ in the minimization of weighted OVR and logit vector $\bm{z}^{CE}$ in the minimization of weighted cross-entropy at time $t$ are
     \begin{align}\textstyle
       \ell_{\mathrm{LM}}(\bm{z}^{OVR}(t))&\textstyle=-2w_1t-2+2W(e^{w_1t+1}),\label{LMOVR} \\\textstyle
       \ell_{\mathrm{LM}}(\bm{z}^{CE}(t))&\textstyle=-\frac{Kw_2t+1}{K-1}+W(\frac{1}{K-1}e^{\frac{Kw_2t+1}{K-1}}),\label{CEOVR}
     \end{align}
     where $w_1\!\in\!\mathbb{R}$ and $w_2\!\in\!\mathbb{R}$ are weights for OVR and cross-entropy, respectively.
     For large $t$, they are approximated by\looseness=-1
     \begin{align}\textstyle
      \!\!\!\!\!\ell_{\mathrm{LM}}(\bm{z}^{OVR}(t))&\textstyle\!\approx\!-\log(w_1t+1)^2,\label{LMOVRApp} \\\textstyle
      \!\!\!\!\!\ell_{\mathrm{LM}}(\bm{z}^{CE}(t))&\textstyle\!\approx\!-\log(Kw_2t+1\!-\!\log(K\!-\!1)^{K\!-\!1}\!),\label{CEOVRApp}
     \end{align}
     and $\lim_{t\rightarrow \infty}\!\frac{\ell_{\mathrm{LM}}(\bm{z}^{OVR})(t)}{\ell_{\mathrm{LM}}(\bm{z}^{CE})(t)}\!=\!2$ for any fixed $w_1$, $w_2$, and $K$.\looseness=-1
    \end{theorem}
    This theorem shows the difference in trajectories of logit margin losses between OVR and cross-entropy under Assumption~\ref{Assum}.
    Regardless of the values of weights $w_1$ and $w_2$, 
    cross-entropy does not increase the logit margins as large as OVR for sufficiently large $t$.
    Thus, OVR increases the small logit margins more effectively than GAIRAT and MAIL, which use weighted cross-entropy (\req{WeightedLoss}). \looseness=-1
 
\begin{figure*}[tb]
       \centering
       \begin{minipage}{0.75\linewidth}
         \begin{subfigure}[t]{0.66\linewidth}
         \centering
     \includegraphics[width=\linewidth]{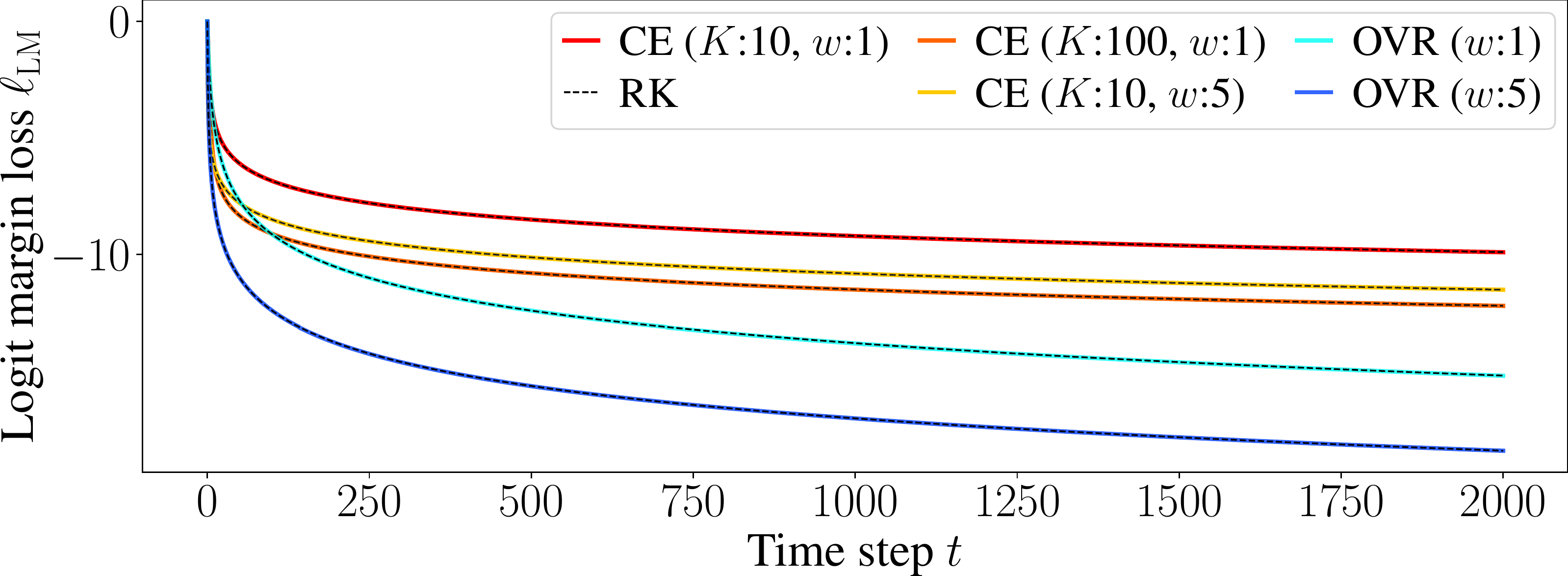}
     \caption{Minimization of \req{ToyProb}} \label{LMSim}
   \end{subfigure}\hfill
     \begin{subfigure}[t]{0.32\linewidth}
       \centering
       \includegraphics[width=\linewidth]{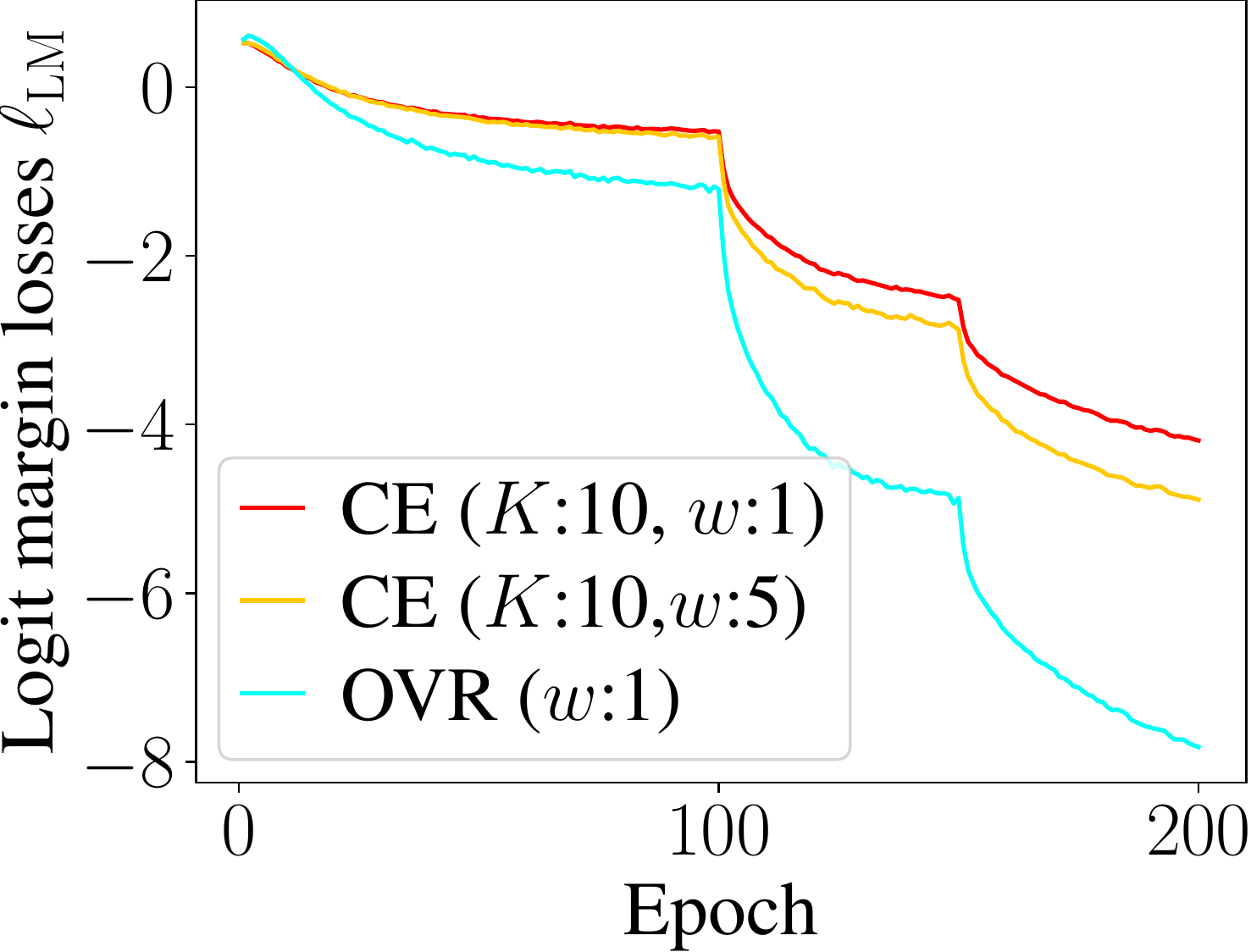}
       \caption{CIFAR 10 (RN18)}      \label{TrajC10}
     \end{subfigure}
     \caption{Trajectories of $\ell_{\mathrm{LM}}$. CE denotes cross-entropy.
     In (a), RK denotes 4-th order Runge–Kutta method with the step size of 0.1. (b) is trajectory in adversarial training on CIFAR10, and its setup is provided in Appendix~\ref{ATLMTrajSec}}
     \end{minipage}
     \hfill
     \begin{minipage}{0.23\linewidth}
       \centering
       \begin{subfigure}{.8\linewidth}
         \includegraphics[width=\linewidth]{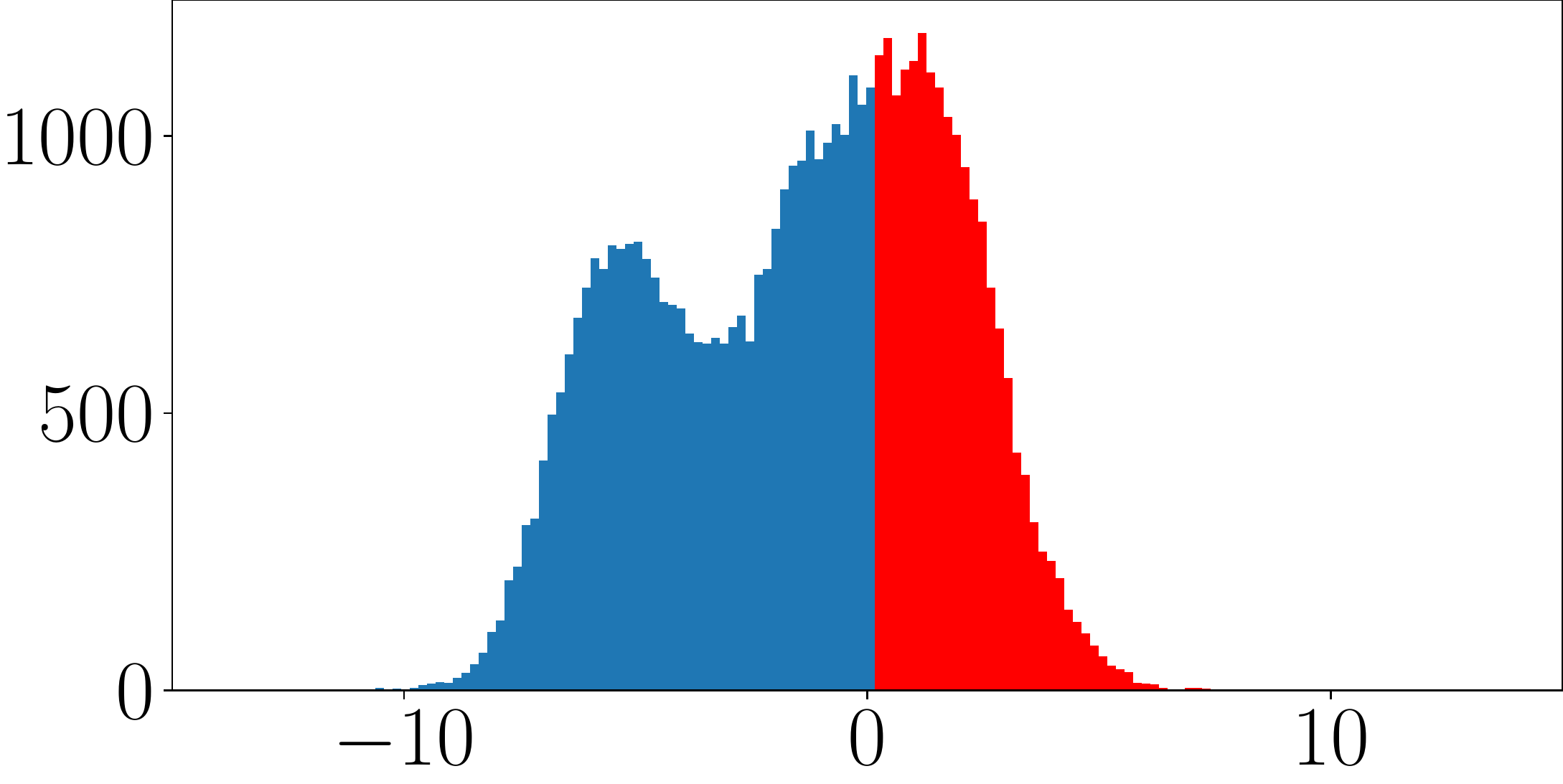}
       \end{subfigure}
       \begin{subfigure}{.8\linewidth}
       \includegraphics[width=\linewidth]{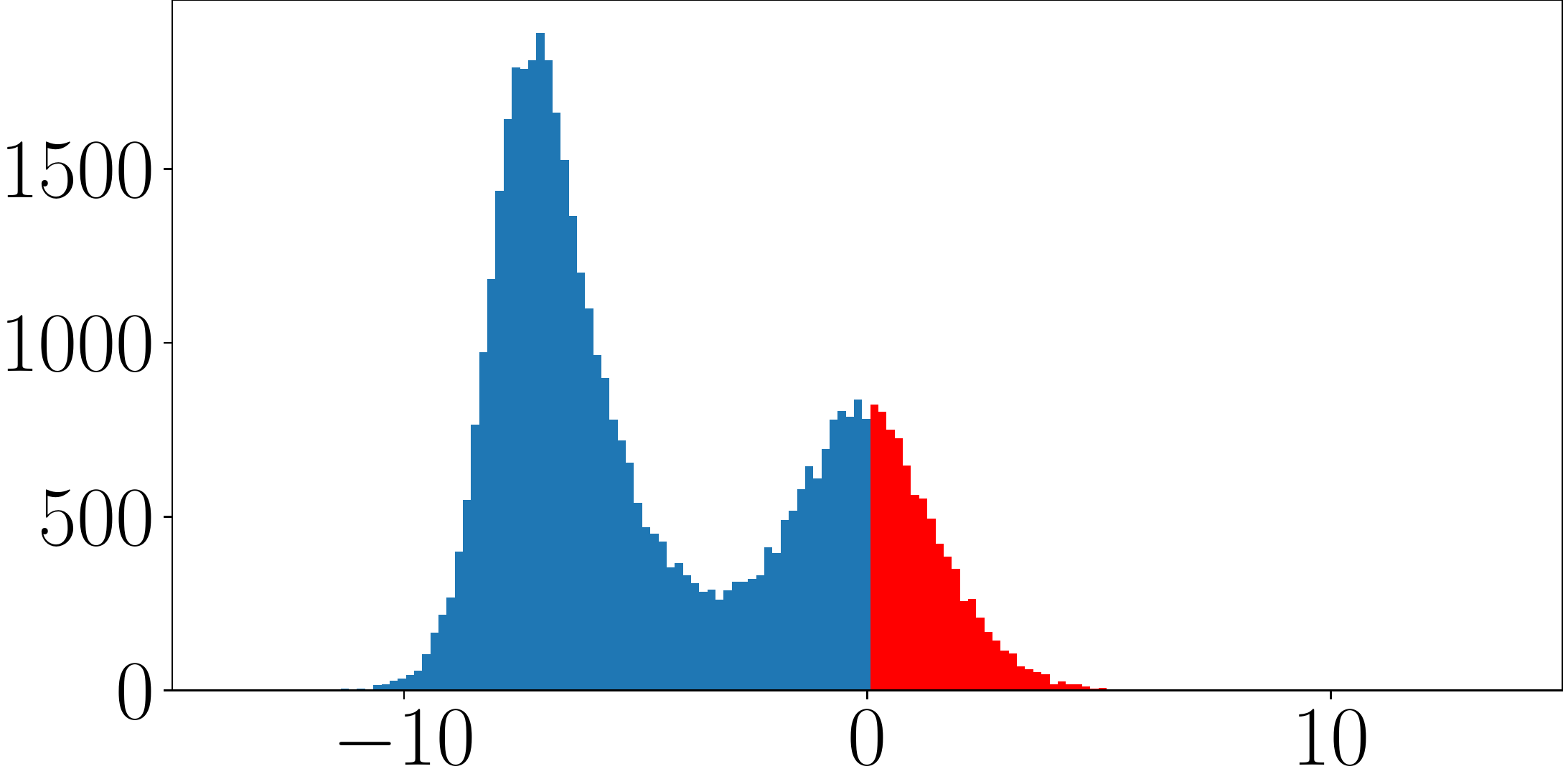}
     \end{subfigure}
     \caption{Histogram of $\ell_{\mathrm{LM}}$ of SOVR at the best (top) and the last (bottom) epoch.}\label{LMHist-SOVR}
     \end{minipage}
   \end{figure*}

    Figure~\ref{LMSim} plots the trajectory of logit margin losses $\ell_{\mathrm{LM}}$ in the minimization of \req{ToyProb}.
    This figure shows the solutions in Theorem~\ref{lmthm} (solid lines): we use Eqs.~(\ref{LMOVR}) and (\ref{CEOVR}) unless overflow occurs due to exponential functions
    and use Eqs.~(\ref{LMOVRApp}) and (\ref{CEOVRApp}) when it occurs.
    It also plots numerical solutions of \req{GF} by using gradients and the Runge–Kutta method as a reference (dashed lines).
   In \rfig{LMSim}, Eqs.~(\ref{LMOVR})-(\ref{CEOVRApp}) exactly match the numerical solutions of RK, and thus, logit margins follow Theorem~\ref{lmthm}.
   In addition, \rfig{LMSim} shows that OVR decreases logit margin losses more than cross-entropy against $t$ regardless of $K$ and $w$.
    Thus, using OVR is more suitable for increasing the logit margin on important samples than previous weighting approaches like GAIRAT and MAIL.
   Figure~\ref{TrajC10} plots trajectories of $\ell_{\mathrm{LM}}$ in adversarial training (\req{AT}) on CIFAR10.
   It shows that the logit margin $|\ell_{\mathrm{LM}}|$ of OVR is about twice as large as that of cross-entropy at the last epoch ($\frac{\ell_{\mathrm{LM}}(\bm{z}^{OVR})}{\ell_{\mathrm{LM}}(\bm{z}^{CE})}\!=\!1.87$ for $w\!=\!1$), like the case of Theorem~\ref{lmthm}.
   Thus, problem \req{ToyProb} is simple but precise enough to explain the difference in the logit margins between OVR and cross-entropy on a real dataset.
    In fact, $\frac{\ell_{\mathrm{LM}}(\bm{z}^{OVR})}{\ell_{\mathrm{LM}}(\bm{z}^{CE})}$  at the last epoch is in [1.5, 2] on other datasets, including CIFAR100 ($K\!=\!100$) (Appendix~\ref{ATLMTrajSec}). \looseness=-1
   
   As above, we prove that OVR is more effective for increasing logit margins of important samples than weighting cross-entropy.
    In the next section, we compose the objective functions switching between OVR and cross-entropy for important and less-important samples.
 
    \subsection{Proposed Objective Function: SOVR}\label{SOVRDef}
    Our proposed objective function is 
    \begin{align}\textstyle
      \mathcal{L}_{\mathrm{SOVR}}(\bm{\theta})
      =\frac{1}{N}&\textstyle\left[\sum_{(\bm{x},y)\in \mathbb{S}}\ell_{\mathrm{CE}}(\bm{z}(\bm{x}^\prime,\bm{\theta}),y)\right.\nonumber\\
      &\textstyle\left.+\lambda \sum_{(\bm{x},y)\in \mathbb{L}}\ell_{\mathrm{OVR}}(\bm{z}(\bm{x}^\prime,\bm{\theta}),y)\right],
    \end{align}
    where $\mathbb{S}$ is a set where logit margin losses $\ell_{LM}$ are smaller than those in the set $\mathbb{L}$, and we have $|\mathbb{S}|\!+\!|\mathbb{L}|=N$.
    These sets correspond to easy and difficult samples in \rfig{LMHist-AT}. 
    In our method, we select the top $M$~\% data points in minibatch of SGD as $\mathbb{L}$.
    $\lambda$ is a hyperparameter to balance the loss, and $\bm{x}^\prime$ is an adversarial example generated by \req{AdvEx}.
    The proposed algorithm is shown in Appendix~\ref{AlgSec}.
    Since we do not additionally generate the adversarial examples for $\ell_{\mathrm{OVR}}$,
    the overhead of our method is negligible: $O(b\log \frac{M}{100}b)$ where $b$ is minibatch size. 
   In the same way to Section~\ref{HistSec}, we evaluate the histograms of logit margin losses for SOVR on CIFAR10 in \rfig{LMHist-SOVR}.
   It shows that SOVR succeeds at increasing the left peak compared with AT (\rfig{LMHist-AT}).
    This is because OVR strongly penalizes important samples in the right peak and moves them into the left peak.
    \begin{figure}
      \centering
      \begin{subfigure}[t]{0.32\linewidth}
        \includegraphics[width=\linewidth]{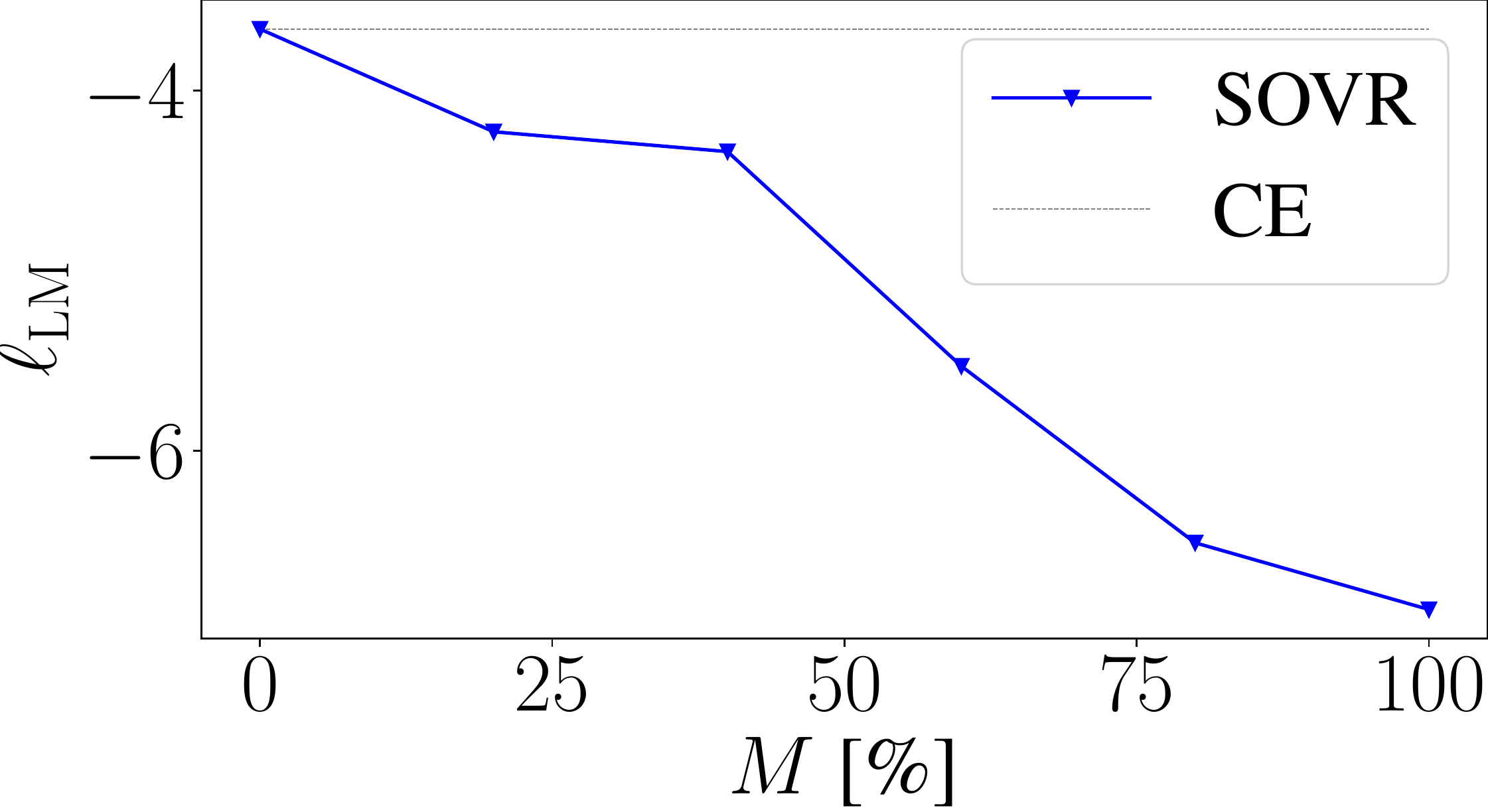}
        \caption{$\ell_{\mathrm{LM}}(\bm{x}^\prime)$}\label{MVsLMPaper}\vspace{-1pt}
      \end{subfigure}  
      \hfill
        \begin{subfigure}[t]{.32\linewidth}\centering
          \includegraphics[width=\linewidth]{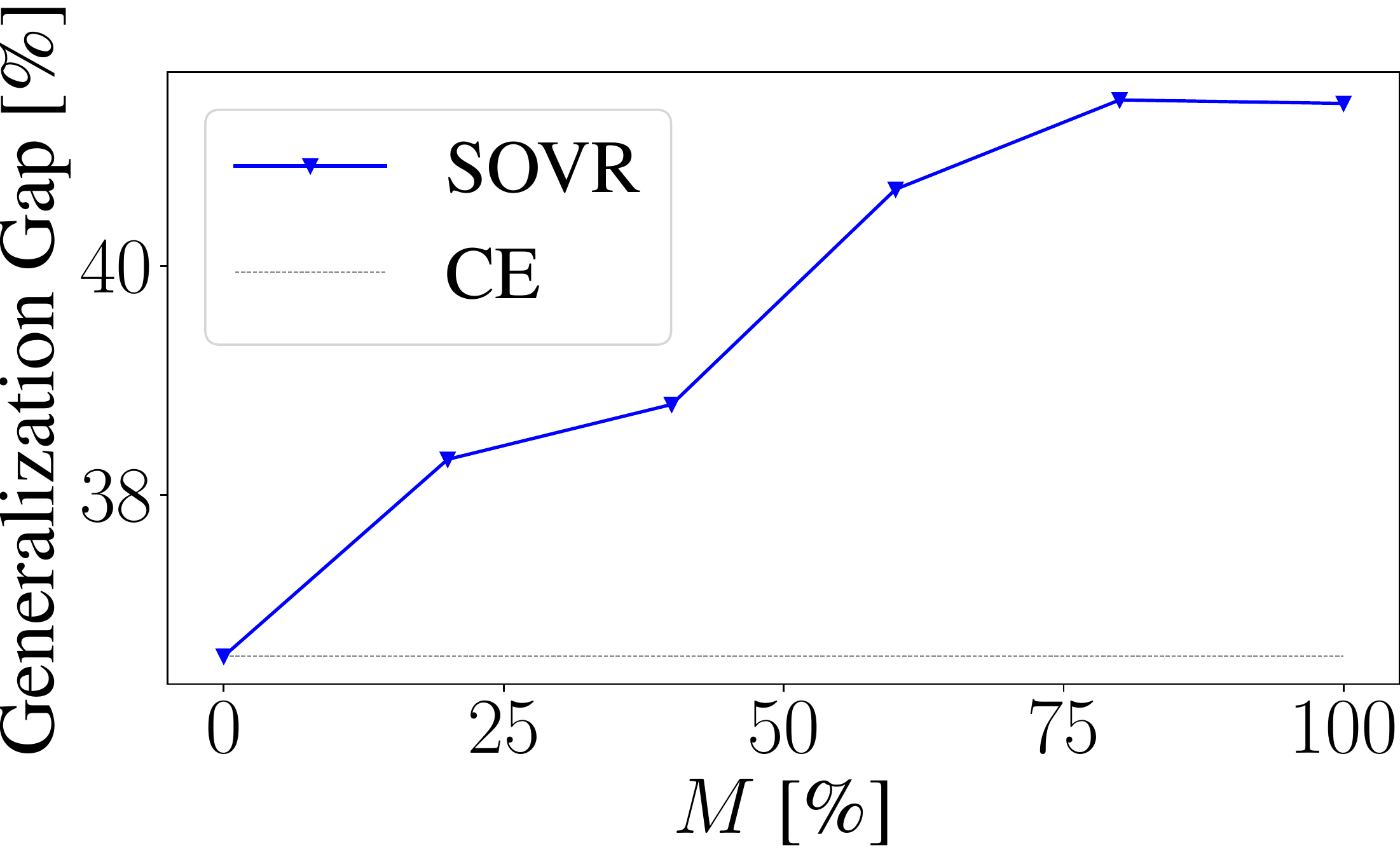}
          \caption{\scalebox{0.9}{Generalization Gap}}\label{MVsGap}\vspace{-1pt}
        \end{subfigure}
        \hfill
              \begin{subfigure}[t]{0.32\linewidth}
        \includegraphics[width=\linewidth]{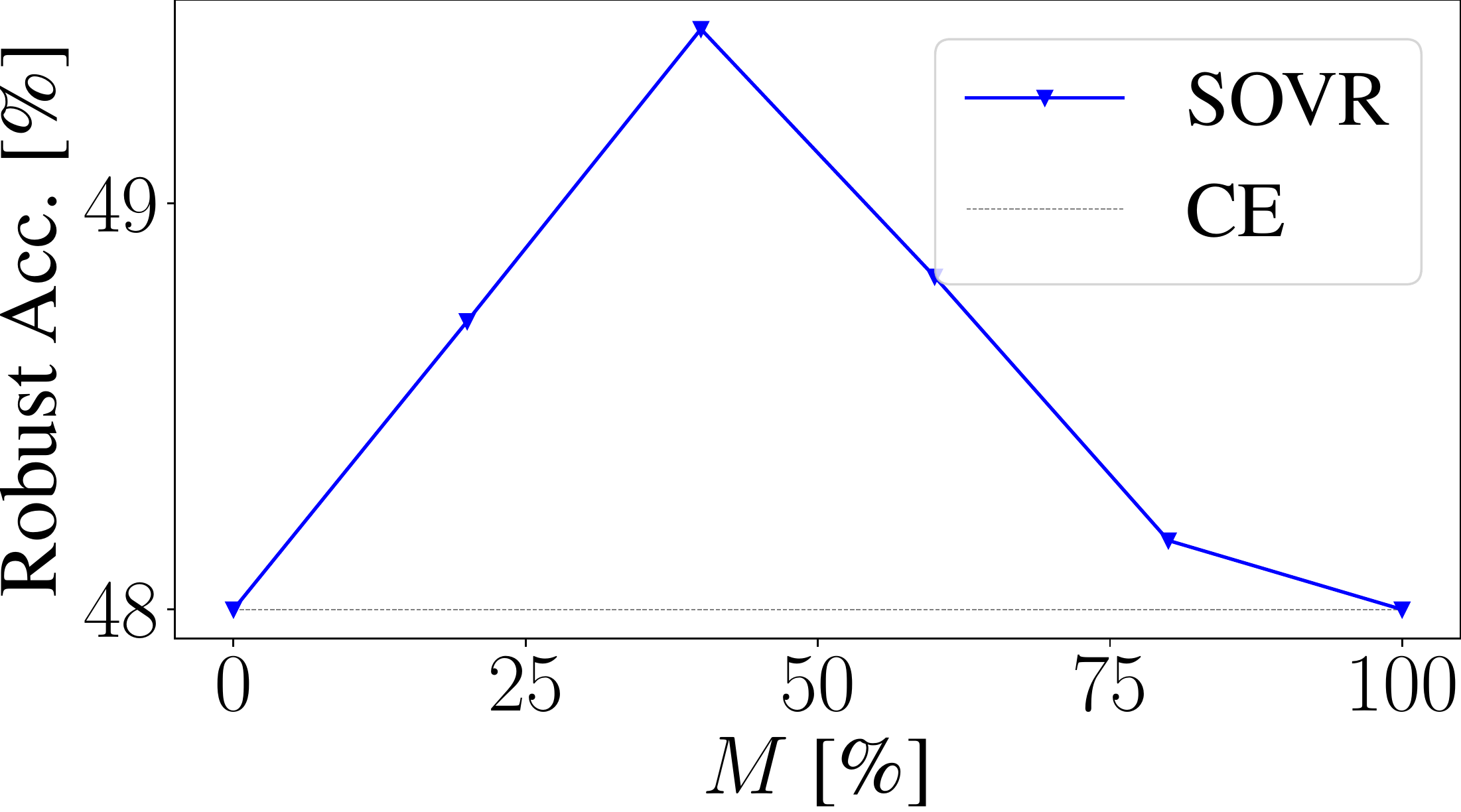}
        \caption{\scalebox{0.9}{Robust Acc.}}\label{MVsAccPaper}\vspace{-1pt}
      \end{subfigure}
      \caption{The effect of rate $M$ of applying OVR. $\lambda$ is set to 0.4. $M\!=\!0$ corresponds to the result of AT with cross-entropy.
      Generalization gap is a gap between training and test robust accuracies against PGD ($\mathcal{K}$=20) at the last epoch.
      Robust Acc. is robust accuracy against Auto-Attack.}\vspace{-2pt}
      \label{MPara}
    \end{figure}
   Though OVR increases logit margins as explained in Section~\ref{OVRBehavSec}, we found that OVR ($M\!=\!100$) is inferior to SOVR
   because OVR for less-important samples can cause overfitting. 
   Figure~\ref{MPara} plots the effect of $M$ in terms of $\ell_{\mathrm{LM}}$ at the last epoch, generalization gap at the last epoch, and robust accuracy against Auto-Attack on CIFAR10.
   It shows that $\ell_{\mathrm{LM}}$ monotonically decreases, i.e., robustness improves, when increasing $M$.
    However, the generalization gap increases at the same time. Robust accuracy takes the largest value at $M\!=\!40$.
    Thus, it is necessary to focus on important samples by switching losses, like existing importance-aware methods.
    We provide the evaluation of effects of $\lambda$ in Appendix~\ref{hypSub}, which shows the similar tendencies. \looseness=-1
    \subsection{Extension for Other Defense Methods}
    Since SOVR only modifies the objective function,
    it can be used with the optimization algorithms for robustness, e.g.,
    adversarial weight perturbation (AWP)~\citep{AWP} or self-ensemble adversarial training (SEAT)~\citep{SEAT},
    and generative data augmentation~\citep{gowal2021generated,rebuffi2021fixing},
    which improve generalization performance in adversarial training.
    However, SOVR is difficult to use with TRADES~\citep{TRADES}
    because TRADES also modifies the objective function.
    To combine our method with TRADES, 
    we propose TSOVR, which uses SOVR instead of cross-entropy for clean data: \vspace{-1pt} 
    \begin{align}\label{TRADESOVR}\textstyle
      &\!\!\!\!\!\textstyle\mathcal{L}_{\mathrm{TS}}\!(\bm{\theta})\!=\!\frac{1}{N}[\mathcal{L}_{\mathrm{SOVR}}\!+\!\beta_T\!\sum_{n}\!\mathrm{KL}(\bm{f}(\bm{x}_n,\!\bm{\theta}),\!\bm{f}(\bm{x}^\prime_n,\!\bm{\theta})\!)]\\
      &\!\!\!\!\!\textstyle \mathcal{L}_{\mathrm{SOVR}}\!=\!\sum_{(\bm{x},y)\in \mathbb{S}}\ell_{\mathrm{CE}}(\bm{z}(\bm{x},\bm{\theta}), y)\nonumber\\
      &~~~~~~~~~~~~~~\textstyle+\lambda \sum_{(\bm{x},y)\in \mathbb{L}}\ell_{\mathrm{OVR}}(\bm{z}(\bm{x},\bm{\theta}),y)
    \end{align}
    where $\lambda$ and $\beta_T$ are hyperparameters.
    $\bm{x}^\prime_n$ is obtained by
    $\bm{x}^\prime_n= \max_{||\bm{x}^\prime_n-\bm{x}||_p\leq \varepsilon}\!\mathrm{KL}(\bm{f}(\bm{x}_n,\bm{\theta}),\bm{f}(\bm{x}_n^\prime,\bm{\theta}))$.
    We evaluate the combinations of SOVR with AWP~\citep{AWP}, SEAT~\citep{SEAT},
    and data augmentation using synthetic data~\citep{gowal2021generated} in the experiments. 
    \section{Related Work}
    The difference in importance of data points in adversarial training has been investigated in several studies~\citep{MART,FAT,sanyal2021how,dong2022exploring}.
    \citet{FAT} investigated the effect of difficult samples on natural generalization performance, i.e., generalization performance on clean data.
    They presented FAT that improves the robustness without compromising the natural generalization performance.
    Unlike FAT, SOVR focuses on the robust performance.
    \citet{sanyal2021how}~and~\citet{dong2022exploring} have investigated the effect of memorization of difficult samples in the generalization performance of adversarial training.
    Whereas they focused on reducing generalization gap by regularization,
    our method reduces robust error on test data by reducing training robust error; i.e., mitigating underfitting more than overfitting.
    \citet{parseval,lmt,pmlr-v139-zhang21b} 
   have shown the relation between robustness, logit margins and Lipschitz constants similar to our justification (Section~\ref{vulsec}).
   They used them to present certified defense methods. 
   However, their empirical robustness is not as well as that of AT.
   Though MMA~\citep{MMA} is the most well-known method to increase logit margins empirically, our experiment shows that MMA does not necessarily increase logit margins.
   OVR has been used as the loss function of multi-class classification~\citep{SAForMC}, but there are few studies using it as a loss function of deep neural networks.
    \citet{padhy2020revisiting} and \citet{saito2021ovanet} used OVR for OOD detection and open-set domain adaptations, respectively.
   However, they did not discuss the effect of OVR in logit margins, 
   and our study is the first work to reveal the effect of OVR in adversarial robustness and logit margins theoretically and experimentally.
   \looseness=-1
    Studies of \citep{hitaj2021evaluating,AutoAttack,EWAT} have reported
    the vulnerabilities of some importance-aware methods to logit scaling attacks or Auto-Attack
    but few studies discuss the causes.
    \citet{EWAT} have pointed out that the cause is high entropy in GAIRAT and MART and
    presented EWAT (entropy-weighted adversarial training), which imposes a higher weight on the higher entropy.
   However, 
    the logit margin is more related to robustness than entropy as discussed in Section~\ref{vulsec}.
   Furthermore, weighted cross-entropy in EWAT is less effective than OVR as shown Theorem~\ref{lmthm}. 
    \section{Experiments}\label{ExSec}\vspace{-1pt}
    \subsection{Setup}\label{setupsec}
    We conducted the experiments for evaluating SOVR.
    We first compare SOVR and TSOVR with Madry's AT~\citep{pgd2}, MMA~\citep{MMA}, MART~\citep{MART},
    GAIRAT~\citep{GAIRAT}, MAIL~\citep{MAIL}, and EWAT~\citep{EWAT}. 
    Additionally, we evaluated TRADES as another baseline that does not consider the importance of data points.
    We used three datasets: CIFAR10, SVHN, and CIFAR100~\citep{cifar,svhn}.
    Next, we evaluate the combination of SOVR and AWP~\citep{AWP}, SEAT~\citep{SEAT}, and data augmentation using 1M synthetic data by DDPM (1M DDPM)~\citep{rebuffi2021fixing,gowal2021generated}.
    Our experimental codes are based on source codes provided by \citep{AWP,MART,MMA,rade2021pytorch}, 
    and 1M synthetic data is provided in \citep{gowal2021generated}.
    We used PreActResNet-18 (RN18)~\citep{resnet} for all datasets and WideResNet-34-10 (WRN)~\citep{WRN} for CIFAR10.
    We used PGD ($\mathcal{K}\!=\!10$, $\eta\!=\!2/255$, $\varepsilon\!=\!8/255$) in training.
    We used early stopping by evaluating test robust accuracies against PGD with $\mathcal{K}\!=\!10$.
    For AWP, SEAT, and 1M DDPM, we use the original public codes~\citep{AWP,SEAT,rebuffi2021fixing}.\footnote{We could not reproduce the results reported in \cite{AWP,SEAT} even though we did not modify their codes. This might be because we report the averaged values for reproducibility.}
    For AT+AWP and SOVR+AWP, we use the training setup that is used for TRADES+AWP in \cite{AWP}
    by changing losses because we found that it achieves a better result.    
    We trained models three times and show the average and standard deviation of test accuracies.
    We set $(M,\lambda)$ in SOVR to $(40, 0.4)$ for CIFAR10 (RN18) and SOVR+AWP, $(30, 0.4)$ for CIFAR10 (WRN) and $(50, 0.6)$ for CIFAR100, $(20, 0.2)$ for SVHN,
    $(40, 0.2)$ for SOVR+1M DDPM. 
    We set $(M, \lambda)$ in TSOVR to $(80, 0.8)$ for CIFAR10 (RN18),
     $(20,0.8)$ for CIFAR10 (WRN) and SVHN, $(50, 0.5)$ for CIFAR100,
    $(100,1.2)$ for TSOVR+AWP, and $(100, 1.6)$ for SOVR+1M DDPM. 
    We use Auto-Attack to evaluate the robust accuracy on test data.
    In Appendix~\ref{SetupSec} and \ref{AddSec}, we provide the
    details of setups and additional results, e.g.,
    evaluation using other various attacks.
    To determine the statistical significant difference, we use t-test with p-value of 0.05.
    \looseness=-1
    \subsection{Results}
    \begin{table*}[tb]
     \centering
     \caption{Robust accuracy against Auto-Attack and clean accuracy on test datasets.}
     \resizebox{\linewidth}{!}{
     \begin{tabular}{ccccccccccc}\toprule
        \multicolumn{10}{c}{Robust Accuracy against Auto-Attack ($L_\infty$, $\varepsilon\!=\!8/255$)}\\ \cmidrule(r){2-2}\cmidrule(r){3-8}\cmidrule(r){9-10}
       &AT&MART&MMA&GAIRAT&MAIL&EWAT&SOVR&TRADES &TSOVR\\ \midrule
     CIFAR10 (RN18)&48.0$\pm$0.2&46.9$\pm$0.3  &37.2$\pm$0.9&37.7$\pm$ 1&39.6$\pm$0.4&48.2$\pm$0.7&$49.4\pm 0.3$&48.8$\pm$ 0.3&$\bm{49.8\pm 0.1}$ \\
     CIFAR10 (WRN)&51.9$\pm$0.5&50.44$\pm$0.09  &43.1$\pm$1&41.8$\pm$0.6&43.3$\pm$0.1&51.6$\pm$0.3&$53.1\pm0.2$&52.9$\pm$ 0.3&$\bm{53.3\pm 0.1}$ \\
     SVHN (RN18)&45.6$\pm$0.4&46.9$\pm$0.3  &41.0$\pm$1&37.6$\pm$0.6&41.2$\pm$0.3&47.6$\pm$0.4&$48.5\pm0.4$&49.4$\pm$ 0.3&$\bm{49.87\pm 0.02}$  \\
     CIFAR100 (RN18)&23.7$\pm$0.3&23.9$\pm$0.1 &18.4$\pm$0.2&19.8$\pm$0.5&16.7$\pm$0.3 &23.52$\pm$0.06&$24.3\pm0.2$&23.2$\pm$ 0.1&$\bm{24.61\pm 0.07}$\\
      \midrule\midrule \multicolumn{10}{c}{Clean Accuracy}\\\midrule
    CIFAR10 (RN18)&81.6$\pm$0.5&78.3$\pm$1  &$\bm{85.5\pm0.7}$&78.7$\pm$ 0.7&79.5$\pm$0.4&82.8$\pm$0.4&81.9$\pm$0.2&82.5$\pm$0.2&81.4$\pm$0.2  \\
   CIFAR10 (WRN)&85.6$\pm$0.1&81.5$\pm$1  &$\bm{87.8\pm1}$&83.0$\pm$0.7 &82.2$\pm$0.4&86.0$\pm$0.5&85.0$\pm$0.2&84.2$\pm$0.7&83.1$\pm$0.2 \\
   SVHN (RN18)&89.8$\pm$0.6&86.9$\pm$0.6  &$\bm{93.9\pm0.4}$&89.9$\pm$0.4 &89.4$\pm$0.4&90.2$\pm$0.6&90.0$\pm$1&88.7$\pm$1&88.5$\pm$0.5\\
   CIFAR100 (RN18)&53.0$\pm$0.7&49.2$\pm$0.1  &$\bm{60.6\pm0.6}$&52.0$\pm$0.5 &46.5$\pm$0.5&54.2$\pm$1&52.1$\pm$0.8&53.8$\pm$0.2&51.6$\pm$0.2\\
     \bottomrule
     \end{tabular}
     }
     \label{PropAA}
   \end{table*}\vspace{-1pt}
   \begin{table*}[tb]
    \centering
    \vspace{-2pt}
    \caption{Robust accuracy against Auto-Attack ($L_\infty$, $\varepsilon=8/255$) on test dataset of CIFAR10.
    SEAT uses WideResNet32-10 following~\cite{SEAT}, and 1M DDPM uses WideResNet28-10 following~\cite{gowal2021generated,rebuffi2021fixing}\looseness=-1}
  \resizebox{\linewidth}{!}{
  \begin{tabular}{ccccccccccccccc}\toprule
    &\multicolumn{4}{c}{AWP (WRN34-10)}&\multicolumn{2}{c}{SEAT}&\multicolumn{4}{c}{+1M DDPM~\cite{gowal2021generated,rebuffi2021fixing}}\\ \cmidrule(r){2-5} \cmidrule(rl){6-7}\cmidrule(l){8-11}
   &AT+AWP&SOVR+AWP&TRADES+AWP&TSOVR+AWP&SEAT&+SOVR&AT&TRADES&SOVR&TSOVR \\ \midrule
 Robust Acc. &55.0$\pm$0.2&56.03$\pm$0.07&55.6$\pm$0.4&56.4$\pm$0.2&54.9$\pm$0.1&55.5$\pm$0.3&59.3$\pm$0.9&60.13$\pm$0.5&60.9$\pm$0.1&60.9$\pm$0.2\\
Clean Acc. &87.8$\pm$0.1&86.9$\pm$0.4&84.9$\pm$0.5&84.9$\pm$0.4&86.8$\pm$0.4&86.5$\pm$0.1&88.46$\pm$0.05&86.1$\pm$0.4&88.05$\pm$0.08&86.08$\pm$0.07 \\
  \bottomrule
  \end{tabular}  
  }
  \label{PropAWP}
\end{table*}
    We list the robust accuracy against Auto-Attack on all datasets in \rtab{PropAA}.
    Comparing SOVR with importance-aware methods and AT, 
    SOVR outperforms them in terms of the robustness against Auto-Attack and the difference is statistically significant.
    This is because SOVR increases the logit margins $|\ell_{\mathrm{LM}}|$ by using OVR.
    In fact, \rfig{LMHist-SOVR} shows that SOVR increases the logit margins $|\ell_{\mathrm{LM}}|$ for important samples.
   MART improved robustness on SVHN and CIFAR100. This might be because MART does not just impose weights on the loss.
   However, its improvement is less than SOVR.
   EWAT also achieves higher robust accuracies than AT on several datasets in \rtab{PropAA}.
    However, EWAT is not as robust as SOVR because EWAT employs weighted cross-entropy, which is less effective at increasing logit margins than OVR (Theorem~\ref{lmthm}).
    In \rtab{PropAA}, SOVR slightly sacrifices clean accuracies under some settings.
    We provide the histograms of logit margin losses on all datasets in Appendix~\ref{HisSub}, which also show SOVR increases margins.
    Comparing SOVR with TRADES and TSOVR, SOVR achieves comparable robust accuracy to TRADES, 
    and TSOVR achieves the best robust accuracy.
    In addition, SOVR achieves the largest value of robust accuracy + clean accuracy in almost all settings, which 
    is discussed in Section~\ref{Sec:trade-off}. 
    \subsubsection{Empirical Evaluation of Potentially Misclassified Samples}\label{miscriSec}
\begin{figure}
      \centering
      \includegraphics[width=\linewidth]{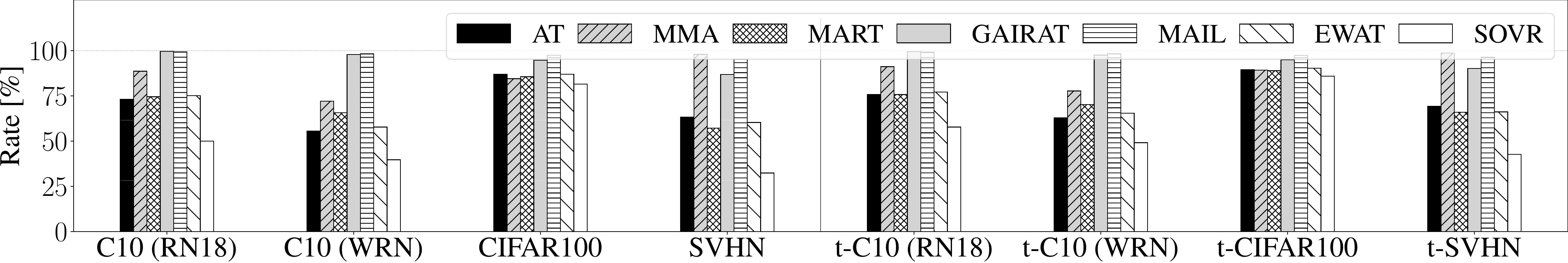}
      \caption{Rate of data satisfying \req{misccri}. C10 and t- represent CIFAR10 and test data, respectively.}\vspace{-4pt}
      \label{Poten}\vspace{-2pt}
    \end{figure}

    As discussed in Section~\ref{lipSec}, we can evaluate the robustness near each data point via logit margins and Lipschitz constants of the class-wise logit functions.
    In this section, we estimate the number of potentially misclassified samples for each method.
    Since Lipschitz constants for deep neural networks are difficult to compute  due to the complexity, 
    we compute the gradient norm of the logit function instead of Lipschitz constants.
    This is because the gradient norm satisfies $\sup_{\bm{x}}||\nabla_{\bm{x}}z_{k}(\bm{x})||_{1}\!=\!L_k$ for $L_k$ such as $|z_{k}(\bm{x}_1)\!-\!z_{k}(\bm{x}_2)|\!\leq\!L_k||\bm{x}_1\!-\!\bm{x}_2||_{\infty}$~\citep{jordan2020exactly},
    and we have
    \begin{proposition}\label{PotenProp}\vspace{-1pt}
      If a data point $\bm{x}$ satisfies\vspace{-2pt}
      \begin{align}\label{misccri}\textstyle
        z_{k^*}\!(\bm{x})\!-\!z_y(\bm{x})\!>\!- (\max_k||\nabla_{\bm{x}}z_{k}(\bm{x})||_{1}\!+\!||\nabla_{\bm{x}}z_{y}(\bm{x})||_{1})\varepsilon,
      \end{align}\vspace{-5pt}
      it is a potentially misclassified sample.
    \end{proposition}\vspace{-4pt}
    Thus, we can empirically estimate the number of potentially misclassified samples on each method 
    by the gradient norms.
    Figure~\ref{Poten} plots the rate of data points that satisfy \req{misccri}, which are potentially misclassified samples.
   Comparing \rfig{Poten} with \rtab{PropAA}, when the methods have the large rate on test data, they have low robust accuracies against Auto-Attack.
    This indicates that this rate is a reasonable metric for estimating robustness though it uses gradient norms instead of Lipschitz constants.
   Whereas most importance-aware methods have higher rates than AT due to small logit margins, SOVR has lower rate.
   This is because SOVR increases logit margins without increasing gradient norms by using OVR, which is more effective in increasing logit margins than weighted cross-entropy (Section~\ref{OVRBehavSec}).
    In \rfig{Poten}, EWAT does not necessarily increase the rate because EWAT uses weighted cross-entropy.
   We discuss the reason the rate of AT gets close to SOVR on CIFAR100 in Appendix~\ref{C100Sec}.
    \subsubsection{Extension for Other Methods}
    To improve the robust accuracy on test data, 
    our method mostly focuses on improving the robustness around training data points rather than regularization.
    Even so, SOVR can be used with recent regularization methods~\citep{AWP,SEAT} and data augmentation~\citep{rebuffi2021fixing,gowal2021generated}.
    We evaluated the combination of SOVR and AWP, SOVR and SEAT, and SOVR and data augmentation using 1M synthetic data by DDPM (1M DDPM). 
    Table~\ref{PropAWP} lists the robust accuracy of the combinations against Auto-Attack
    and shows that SOVR improved the performance of other recent methods. Thus, SOVR and these methods complementarily improve the performance.
    \subsubsection{Evaluation of Trade-off}\label{Sec:trade-off}
\begin{figure}[tb]
        \centering
      \includegraphics[width=.5\linewidth]{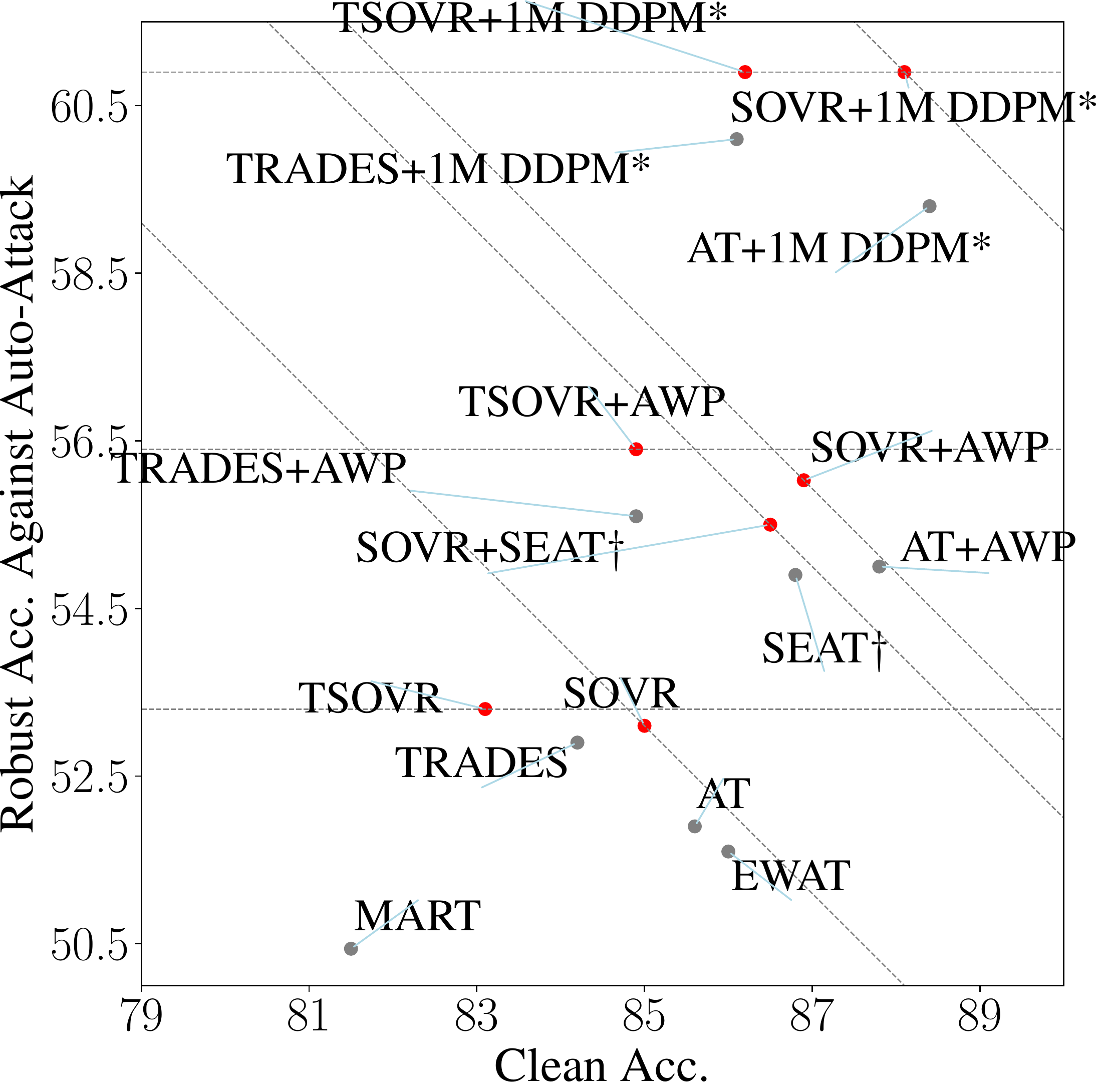}
      \caption{Trade-off between clean and robust accuracies (WRN).
      $*$ use WRN28-10 and $\dagger$ use WRN32-10.}    \vspace{-10pt}
      \label{TO2}
  \end{figure}

    We evaluate the trade-off between robustness and clean accuracy.
    Figure~\ref{TO2} plots clean accuracy and robust accuracy against Auto-Attack when using WRN on CIFAR10. Diagonal lines are lines satisfying $\mathrm{Clean Acc.}\!+\!\mathrm{Robust Acc.}\!=\!\mathrm{Const}$ through SOVR, SOVR+AWP, SOVR+SEAT, and SOVR+1M DDPM.
    SOVR achieves a good trade-off: Robust Acc. + Clean Acc. of SOVR, SOVR+AWP, and SOVR+DDPM are the best values under each condition.
    If we require the most robust models with sacrificing clean accuracies, TSOVR achieves the best robust accuracy against Auto-Attack.
    Therefore, SOVR and TSOVR are better objective functions than TRADES and cross-entropy in terms of trade-off and robustness, respectively.\looseness=-1
    \section{Conclusion}
    \vspace{-2pt}
    We investigated the reason importance-aware methods fail to improve the robustness against Auto-Attack.
    Our empirical results showed the reason to be that they decrease logit margins of less-important samples besides those of important samples.
    From the observation, we proposed SOVR, which switches from cross-entropy to OVR in order to focus on important samples.
    We proved that OVR increases logit margins more than cross-entropy for a simple problem 
    and experimentally showed that SOVR increases the margins of important samples and improves the robustness. 
    \bibliography{CameraBib}
    \bibliographystyle{icml2023}
        
    \newpage
    \appendix
    \onecolumn
   
    \section{Proofs}\label{proofsec}
    \subsection{Proof of Proposition~\ref{CertProp}}
    \begin{proof}
      From the definition of Lipschitz constants, we have 
      \begin{align}
        |z_{k}(\bm{x}+\bm{\delta})-z_{k}(\bm{x})|\leq L_k||\bm{x}+\bm{\delta}-\bm{x}||_{\infty}=L_k\varepsilon,
      \end{align}
     Thus, we have $z_{k}(\bm{x}+\bm{\delta})\leq z_{k}(\bm{x})+L_k\varepsilon$ if $z_{k}(\bm{x}+\bm{\delta})\geq z_{k}(\bm{x})$
     and  $z_{k}(\bm{x}+\bm{\delta})\geq z_{k}(\bm{x})-L_k\varepsilon$ if $z_{k}(\bm{x}+\bm{\delta})\leq z_{k}(\bm{x})$.
      Therefore, the following inequalities hold for the not potentially misclassified samples:
        \begin{align}
          \max_{k\neq y}z_k(\bm{x}+\bm{\delta})-z_y(\bm{x}+\bm{\delta})
          &\leq z_{k^\prime}(\bm{x})+L_{k^\prime}\varepsilon-(z_y(\bm{x})-L_y \varepsilon)\nonumber\\
          &\leq z_{k^*}(\bm{x})+L_{\hat{k}}\varepsilon-z_y(\bm{x})+L_y \varepsilon\nonumber\\
          &\leq z_{k^*}(\bm{x})-z_y(\bm{x}) +(L_{\hat{k}}+L_y)\varepsilon,\label{CertEq}
          \end{align}
          where $k^\prime\!=\!\mathrm{arg}\max_{k\neq y}z_k(\bm{x}+\bm{\delta})$, $k^*\!=\!\mathrm{arg}\max_{k\neq y}z_k(\bm{x})$, and $\hat{k}\!=\!\mathrm{arg}\max_{k}L_k$.
          From $z_{k^*}(\bm{x})-z_y(\bm{x})\leq -(L_{\hat{k}}+L_y)\varepsilon$ for not potentially misclassified samples and \req{CertEq}, we have $\max_{k\neq y}z_k(\bm{x}+\bm{\delta})-z_y(\bm{x}+\bm{\delta})\leq 0,~~\forall\bm{\delta} \in \{||\bm{\delta}||_{\infty}\leq \varepsilon\}$.
          Thus, models are guaranteed to classify adversarial examples of these data points accurately.
        \end{proof} 
    \subsection{Proof of Theorem~\ref{CEVsOVR}}
    \begin{proof}
      By using logit functions, $\SCE$ can be written as
      \begin{align}\textstyle
        \ell_{\mathrm{CE}}(\bm{z}(\bm{x},\bm{\theta}),y)=-z_{y}(\bm{x})+\mathrm{log}\sum_{k}e^{z_{k}(\bm{x})}.\label{SCE}
      \end{align}
      when a model uses a softmax function.
      Compared with OVR
       $\ell_{\mathrm{OVR}}(\bm{z}(\bm{x},\bm{\theta}),y)\textstyle=-z_{y}(\bm{x})+\sum_{k}\mathrm{log}(1+e^{z_{k}(\bm{x})})$, the difference is only the second term.
      Thus, $\ell_{\mathrm{OVR}}(\bm{z}(\bm{x},\bm{\theta}),y)-\ell_{\mathrm{CE}}(\bm{z}(\bm{x},\bm{\theta}),y)$ can be written as 
      \begin{align}\textstyle
        \ell_{\mathrm{OVR}}(\bm{z}(\bm{x},\bm{\theta}),y)-\ell_{\mathrm{CE}}(\bm{z}(\bm{x},\bm{\theta}),y)&\textstyle=\sum_{k}\mathrm{log}(1+e^{z_{k}})- \mathrm{log}\sum_{k}e^{z_{k}},\\
        &\textstyle=\mathrm{log}\prod_k (1+e^{z_{k}})- \mathrm{log}\sum_{k}e^{z_{k}}.
      \end{align}
      Since a logarithm is a strictly increasing function, we have 
      \begin{align}\label{signeq}\textstyle
        \prod_k (1+e^{z_{k}})-\sum_{k}e^{z_{k}}\geq 0\Rightarrow
       \mathrm{log}\prod_k (1+e^{z_{k}})- \mathrm{log}\sum_{k}e^{z_{k}}\geq 0.
      \end{align}
      Since $e^{z_k}\geq 0$ for any $z_k\in \mathbb{R}$, we have 
      \begin{align}\textstyle
       \prod_k (1+e^{z_{k}})-\sum_{k}e^{z_{k}}=1+\sum_{k}e^{z_{k}}+R(e^{z_k})-\sum_{k}e^{z_{k}}=1+R(e^{z_k})\geq 0
      \end{align}
      where $R(e^{z_k})$ is the second or higher order terms of $e^{z_{k}}$, and it takes a positive value because $e^{z_{k}}\!\geq\!0$.
    Thus,
    the left hand side of \req{signeq} holds, and we have $\mathrm{log}\prod_k (1+e^{z_{k}})- \mathrm{log}\sum_{k}e^{z_{k}}\geq 0$.
    Therefore, we have $\ell_{\mathrm{OVR}}(\bm{z}(\bm{x},\bm{\theta}),y)-\ell_{\mathrm{CE}}(\bm{z}(\bm{x},\bm{\theta}),y)\geq 0$: i.e., 
    $0\leq \ell_{\mathrm{CE}}(\bm{z}(\bm{x},\bm{\theta}),y)\leq \ell_{\mathrm{OVR}}(\bm{z}(\bm{x},\bm{\theta}),y)$ since $\ell_{\mathrm{CE}}(\bm{z}(\bm{x},\bm{\theta}),y)\geq 0$.
    Next, when $z_k(\bm{x})\rightarrow -\infty$ for $k\neq y$, 
    we have $e^{z_k}\rightarrow 0$ and
    \begin{align}
      \lim_{\substack{z_y\rightarrow +\infty,\\z_k\rightarrow -\infty~\mathrm{for}~k\neq y}}\!\!\!\!\!\!\ell_{\mathrm{CE}}(\bm{z}(\bm{x},\bm{\theta}),y)=
      \lim_{\substack{z_y\rightarrow +\infty,\\z_k\rightarrow -\infty~\mathrm{for}~k\neq y}}\!\!\!\!\!\!-z_{y}(\bm{x})+\mathrm{log}\sum_{k}e^{z_{k}(\bm{x})}
      = -z_y+\mathrm{log}(e^{z_y})=0.
    \end{align}
    On the other hand, when $z_ y(\bm{x})\rightarrow +\infty$ and $z_k(\bm{x})\rightarrow -\infty$ for $k\neq y$,
    we have $\mathrm{log}(1+e^{z_y})\rightarrow z_y$ and $\mathrm{log}(1+e^{z_k})\rightarrow 0$.
    Thus, we have
    \begin{align}
      \lim_{\substack{z_y\rightarrow +\infty,\\z_k\rightarrow -\infty~\mathrm{for}~k\neq y}}\!\!\!\!\!\!\ell_{\mathrm{OVR}}(\bm{z}(\bm{x},\bm{\theta}),y)=\lim_{\substack{z_y\rightarrow +\infty,\\z_k\rightarrow -\infty~\mathrm{for}~k\neq y}}
      \!\!\!\!\!\!-z_{y}(\bm{x})+\sum_{k}\mathrm{log}(1+e^{z_{k}(\bm{x})})
     =-z_y+z_y=0,
    \end{align}
      which completes the proof.
\end{proof} 
    \subsection{Proof of Lemma~\ref{ovrlem}}
   \begin{proof}
     From the assumption, we consider the following ordinary differential equation (ODE):
    \begin{align}
     \frac{dz_k}{dt}=-\frac{\partial w\ell_{OVR}(\bm{z},y)}{\partial z_k}.
 \end{align}
 The initial condition is $\bm{z}(0)=\bm{0}$.
 For the correct label $y$, the gradient of OVR is given by
 \begin{align}
   \frac{\partial \ell_{OVR}(\bm{z},y)}{\partial z_y}=-1+\frac{e^{z_y}}{e^{z_y}+1}.
 \end{align}
 Thus, ODE becomes 
 \begin{align}
   \frac{dz_y}{dt}&=w(1-\frac{e^{z_y}}{e^{z_y}+1})\\
   (1+e^{z_y})d{z_y}&=wdt\\
   z_y+e^{z_y} &=wt+c\\ 
   e^{z_y+e^{z_y}}&=e^{wt+c},
 \end{align}
 where $c$ is a constant, which is determined by the initial condition.
 We apply the Lambert $W$ function \citep{Wfunction1} for both sides and use $\log W(x)=\log x-W(x)$ for $x>0$ as
 \begin{align}
   e^{z_y}&=W(e^{wt+c}),\\ 
   z_y&=wt+c -W(e^{wt+c}).
 \end{align}
 From the assumption, we have $z_y(0)=0$, and thus, $c$ satisfies the following equality:
 \begin{align}
   c -W(e^{c})=0.
 \end{align}
 From $W(xe^x)=x$, we have $c=1$ and the logit of the correct label is given by
 \begin{align}
   z_y&=wt+1-W(e^{wt+1}).
 \end{align}
 Next, we consider the logit of another label $z_k$ for $k\neq y$.
 Since the gradient for the logit of incorrect label is
 $\frac{\partial \ell_{OVR}(\bm{z},y)}{\partial z_k}=\frac{e^{z_k}}{e^{z_k}+1}$, 
 we have 
 \begin{align}
   \frac{dz_k}{dt}&=-\frac{we^{z_k}}{1+e^{z_k}}\\
   (e^{-z_k}+1)dz_k&=-w dt.
 \end{align}
 It is solved in the same way as $z_y$, and we have
 \begin{align}
   z_k&=-wt-1+W(e^{wt+1}),
 \end{align}
 for $k\neq y$, which completes the proof.
 \end{proof}
    \subsection{Proof of Lemma~\ref{celem}}
   \begin{proof}
 We first solve the ODE for the logit of the correct label $z_y$.
   The gradient of cross-entropy is 
 \begin{align}
   \frac{\partial \ell_{CE}(\bm{z},y)}{\partial z_k}=-\delta_{ky}+\frac{e^{z_k}}{\sum_{m} e^{z_m}}.\label{CEGrads}  
 \end{align}
 From \req{CEGrads}, we have the following ODE:
    \begin{align}\label{CEODEy}
     \frac{dz_y}{dt}&=w\frac{\sum_{m\neq y} e^{z_m}}{\sum_{m\neq y} e^{z_m}+e^{z_y}}.
 \end{align}
 Since $\bm{z}=\bm{0}$ at $t=0$, we have $z_i=z_j$ for $\forall i,j\neq y$.
 In addition, we have $\sum_i \frac{dz_i(t)}{dt}=\sum_i\frac{w\partial \ell_{CE}(\bm{z},y)}{\partial z_i}=0$ for $\forall t$.
 Thus, logits satisfy the following equality:
 \begin{align}
   z_y&=-(K-1)z_k,\label{LogitRel}
 \end{align}
 for $k\neq y$.
 From \req{LogitRel}, \req{CEODEy} becomes
 \begin{align}
   \frac{dz_y}{dt}&=w\frac{(K-1) e^{-\frac{1}{K-1} z_y}}{(K-1) e^{-\frac{1}{K-1} z_y}+e^{z_y}}\\
   (K-1+e^{\frac{K}{K-1}z_y})dz_y&=w(K-1)dt\\
   \frac{K}{K-1}z_y+\frac{1}{K-1}e^{\frac{K}{K-1}z_y}&=\frac{K}{K-1}wt+c\\
   \frac{1}{K-1}e^{\frac{K}{K-1}z_y}e^{\frac{1}{K-1}e^{\frac{K}{K-1}z_y}}&=\frac{1}{K-1}e^{\frac{K}{K-1}wt+c}\\
   \frac{1}{K-1}e^{\frac{K}{K-1}z_y}&=W(\frac{1}{K-1}e^{\frac{K}{K-1}wt+c})\\
   \frac{K}{K-1}z_y&=\log\left\{ (K-1)W(\frac{1}{K-1}e^{\frac{K}{K-1}wt+c})\right\}\\
   z_y&=wt+\frac{K-1}{K}c- \frac{K-1}{K}W(\frac{1}{K-1}e^{\frac{K}{K-1}wt+c})\label{LogYCE}
 \end{align}
 where $c$ is a constant, which is determined by the initial condition.
 From Assumption, we have
 \begin{align}
   z_y(0)&=\frac{K-1}{K}c- \frac{K-1}{K}W(\frac{1}{K-1}e^{c})=0\\
 W(\frac{1}{K-1}e^{c})&=c.\label{constant}
 \end{align}
 Thus, we have $c=\frac{1}{K-1}$ since $W(xe^x)=x$. 
 From Eqs.~(\ref{LogitRel}) and (\ref{LogYCE}), we have 
 \begin{align}
   z_y(t) &=wt+\frac{1}{K}- \frac{K-1}{K}W(\frac{1}{K-1}e^{\frac{K}{K-1}wt+\frac{1}{K-1}})\\
   z_k(t) &=-\frac{1}{K-1}wt-\frac{1}{K(K-1)}+\frac{1}{K}W(\frac{1}{K-1}e^{\frac{K}{K-1}wt+\frac{1}{K-1}})~~\mathrm{for}~~~k\neq y
 \end{align}
 which completes the proof.
 \end{proof}
    \subsection{Proof of Theorem~\ref{lmthm}}
 From Lemmas~\ref{ovrlem} and \ref{celem}, we have 
    \begin{align}
     \ell_{\mathrm{LM}}(\bm{z}^{OVR}(t))&=-2w_1t-2 +2W(e^{w_1t+c}),\label{OVRlamb}\\
     \ell_{\mathrm{LM}}(\bm{z}^{CE}(t))&=-\frac{K}{K-1}w_2t-\frac{1}{K-1}+W(\frac{1}{K-1}e^{\frac{K}{K-1}w_2t+\frac{1}{K-1}}).\label{CElamb}
   \end{align}
 Since $W(x)= \log(x)-\log(\log(x))+O(1)$ for large $x$~\citep{Wfunction2}, 
 we have
 \begin{align}
   \ell_{\mathrm{LM}}(\bm{z}^{OVR}(t))\approx&-2w_1t-2+2(w_1t+1-\log(w_1t+1))\\
   =&-\log(w_1t+1)^2\\
   \ell_{\mathrm{LM}}(\bm{z}^{CE}(t))\approx&-\frac{K}{K-1}w_2t-\frac{1}{K-1}\\
   &-\log(K-1)+\frac{K}{K-1}w_2t+\frac{1}{K-1}\\
   &-\log(-\log(K-1)+\frac{K}{K-1}w_2t+\frac{1}{K-1})\\
   =&-\log(Kw_2t+1-(K-1)\log(K-1)).
 \end{align}
 From the above, we have
 \begin{align}
   \lim_{t\rightarrow \infty} \frac{\ell_{\mathrm{LM}}(\bm{z}^{OVR})(t)}{\ell_{\mathrm{LM}}(\bm{z}^{CE})(t)}&=\lim_{t\rightarrow \infty} \frac{\log(w_1t+1)^2+O(1)}{\log(Kw_2t+1-(K-1)\log(K-1))+O(1)},\\
   &=\lim_{t\rightarrow \infty} \frac{2\log t+2\log(w_1+t^{-1})+O(1)}{\log t+\log(Kw_2+t^{-1}(1-(K-1)\log(K-1)))+O(1)},\\
   &=\lim_{t\rightarrow \infty} \frac{2+\frac{2}{\log t}\log(w_1+t^{-1})+\frac{O(1)}{\log t}}{1+\frac{\log(Kw_2+t^{-1}(1-(K-1)\log(K-1)))}{\log t}+\frac{O(1)}{\log t}},\\
   &=2,
 \end{align}
 which completes the proof.
 \subsection{Proof of Proposition~\ref{PotenProp}}
 \begin{proof}
   Since we have $\sup_{\bm{x}}||\nabla_{\bm{x}}z_{k}(\bm{x})||_q= L_k$ for an $L_k$-Lipschitz function such as $|z_k(\bm{x}_1)-z_k(\bm{x}_2)|\leq L_k||\bm{x}_1-\bm{x}_2||_{p}$ where $1/q+1/p=1$~\citep{jordan2020exactly},
    the following inequality holds if $\bm{z}_{k^*}(\bm{x})-\bm{z}_y(\bm{x})>- (\max_k||\nabla_{\bm{x}}z_{k}(\bm{x})||_1+||\nabla_{\bm{x}}z_{y}(\bm{x})||_1)\varepsilon$ and $p=\infty$:
   \begin{align}
     \bm{z}_{k^*}(\bm{x})-\bm{z}_y(\bm{x})&>-(\max_k||\nabla_{\bm{x}}z_{k}(\bm{x})||_1+||\nabla_{\bm{x}}z_{y}(\bm{x})||_1)\varepsilon\nonumber\\
     &\geq -(\max_k\sup_{x}||\nabla_{\bm{x}}z_{k}(\bm{x})||_1+\sup_{x}||\nabla_{\bm{x}}z_{y}(\bm{x})||_1)\varepsilon\geq -(L_{\hat{k}}+L_y)\varepsilon,
   \end{align}
   because $||\nabla_{\bm{x}}z_{k}(\bm{x})||_1\leq L_k$ for $p=\infty$.
   Thus, we have $z_{k^*}(\bm{x})-z_y(\bm{x})>-(L_{\hat{k}}+L_y)\varepsilon$ on this condition, which completes the proof.
 \end{proof} 
    \section{Algorithm}\label{AlgSec}
    The proposed algorithm is shown in Algorithm~\ref{Alg}.
    We first generate $\bm{x}^\prime$ in Line~3 and compute the $\ell_{LM}$ for them in Line 4.
    In Line~6, we select the top $M$~\% samples in minibatch and add them to $\mathbb{L}$.
    Finally, we compute the objective $\mathcal{L}_{\mathrm{SOVR}}$ and its gradient to update $\bm{\theta}$.
    Since we do not additionally generate the adversarial examples for $\ell_{\mathrm{OVR}}$,
    the overhead of our method is $O(|\mathbb{B}|\log \frac{M}{100}|\mathbb{B}|)$, which is the computation cost of the heap sort for selecting $\mathbb{L}$ in Line 6.
   It is negligible in the whole computation since deep models have huge parameter-size compared with batch-size $|\mathbb{B}|$. \looseness=-1
    \begin{algorithm}[tb]
      \caption{Switching one-vs-the-rest by the criterion of a logit margin loss}
      \label{Alg}
      \begin{algorithmic}[1]
          \STATE Select the minibatch $\mathbb{B}$
          \FOR{$\bm{x}_n \in \mathbb{B}$}
              \STATE Generate adversarial examples $\bm{x}_n'=\argmax_{||\bm{x}_n^\prime-\bm{x}_n||_{\infty}\leq \varepsilon} \ell_{\mathrm{CE}}(\bm{z}(\bm{x}_n^\prime,\bm{\theta}), y_n)$ by PGD
              \STATE $\ell_{\mathrm{LM}}(\bm{z}(\bm{x}_n^\prime,\bm{\theta}),y_n)=\max_{k\neq y_n}z_k(\bm{x}_n^\prime)-z_{y_n}(\bm{x}_n^\prime)$
          \ENDFOR
          \STATE Select top $\frac{M}{100}|\mathbb{B}|$ samples of $(\bm{x}_n^\prime, y_n)$ in terms of $\ell_{\mathrm{LM}}(\bm{z}(\bm{x}_n^\prime,\bm{\theta}),y_n)$ and add them to $\mathbb{L}$
          \STATE $\mathcal{L}_{\mathrm{SOVR}}(\bm{\theta})=\frac{1}{|\mathbb{B}|}\left[\sum_{(\bm{x},y)\in \mathbb{B}\backslash \mathbb{L}} \ell_{\mathrm{CE}}(\bm{z}(\bm{x}^\prime,\bm{\theta}),y)+\lambda\sum_{(\bm{x}, y)\in \mathbb{L}}\ell_{\mathrm{OVR}}(\bm{z}(\bm{x}^\prime,\bm{\theta}), y)\right]$
          \STATE Update the parameter $\bm{\theta}\leftarrow\bm{\theta}-\eta\nabla_{\bm{\theta}} \mathcal{L}_{\mathrm{SOVR}}(\bm{\theta})$
      \end{algorithmic}
    \end{algorithm}\vspace{-5pt}
    \section{Additional Explanation about Previous Methods} \label{prevSec}
    \subsection{MART~\citep{MART}}
    MART~\citep{MART} uses a similar approach to importance-aware methods. It regards misclassified samples as important samples
    and minimizes
    \begin{align}\textstyle
      \ell_{\mathrm{MART}}(\bm{x}^\prime,y,\bm{\theta})&\textstyle=\mathrm{BCE}(\bm{f}(\bm{x}^\prime,\bm{\theta}),y)+\lambda \mathrm{KL}(\bm{f}(\bm{x},\bm{\theta}),\bm{f}(\bm{x}^\prime,\bm{\theta}))\cdot (1-f_{y}(\bm{x},\bm{\theta})),
      \textstyle   
   \end{align}
    where 
    $\mathrm{BCE}(\bm{f}(\bm{x}^\prime,\bm{\theta}),y)\!=\!-\mathrm{log}(f_{y}(\bm{x}^\prime ,\bm{\theta}))\!-\!\mathrm{log}(1\!-\!\max_{k\neq y}f_k(\bm{x}^\prime ,\bm{\theta}))$ and 
    $\mathrm{KL}$ is Kullback-Leibler divergence.
    MART controls the difference between the loss on less-important and important samples 
    via $1-f_y(\bm{x}_n,\bm{\theta})$: MART tends to ignore the second term when the model is confident in the true label. 
 
    \subsection{MMA~\citep{MMA}}
    MMA~\citep{MMA} attempts to maximize the distance\footnote{For clarity, we use the term ``margin" only for the distance between logits of the true labels and of the label that has the largest logit except for the true label,
     not for the distance between data points and the decision boundary.} between data points and the decision boundary for robustness.
    MMA regards $\min_{\bm{\delta}}||\bm{\delta}_n||_{\infty}$ subject to $\{\ell_{\mathrm{LM}}(\bm{z}(\bm{x}+\bm{\delta},\bm{\theta}),y)\geq 0 \}$
    as the distance. 
    By using this distance, MMA minimizes the following loss:
    \begin{align}
      &\textstyle\mathcal{L}(\bm{\theta})\textstyle=\frac{1}{3}\sum_{n=1}^{N}\ell_{\mathrm{CE}}(\bm{z}(\bm{x}_n,\bm{\theta}),y_n)+\frac{2}{3}\mathcal{L}_{\mathrm{MMA}}(\bm{\theta})\\
      &\textstyle\mathcal{L}_{\mathrm{MMA}}(\bm{\theta})\textstyle\!=\!\sum_{(\bm{x},y)\in \mathcal{S^+}\cap \mathcal{H}}\ell_{\mathrm{CE}}(\bm{z}(\bm{x}\!+\!\bm{\delta}_{\mathrm{MMA}},\bm{\theta}),y)\!+\!\sum_{(x,y)\in \mathcal{S^-}}\ell_{\mathrm{CE}}(\bm{z}(\bm{x},\bm{\theta}),y)\\
      &\textstyle\bm{\delta}_{\mathrm{MMA}}\textstyle=\mathrm{arg}\min_{\ell_{\mathrm{SLM}}(\bm{z}(\bm{x}+\bm{\delta},\bm{\theta}),y)\geq 0}||\bm{\delta}||_{\infty}\label{distance}\\
      &\textstyle \ell_{\mathrm{SLM}}(\bm{z}(\bm{x},\bm{\theta}),y)\textstyle =\mathrm{log}\sum_{k\neq y}e^{z_k(\bm{x})}-z_y(x)
    \end{align}
    where $\mathcal{S^+}$ is a set of correctly classified data points,
    and $\mathcal{S^-}$ is a set of misclassified samples.
    $\mathcal{H}$ is a set of data points that have a smaller distance than threshold $d_{\mathrm{max}}$ as $\mathcal{H}=\{(x_n,y_n)| \min_{\bm{\delta}_n} ||\bm{\delta}_n||_{\infty}\leq d_{\mathrm{max}}\}$.
    Since MMA uses $\bm{\delta}$ whose magnitude $||\bm{\delta}||_{\infty}$ depends on data points as \req{distance},
    we consider that it has similar effects to the importance-aware methods.
    In MMA, \citet{MMA} use $\ell_{\mathrm{SLM}}(\bm{z}(\bm{x},\bm{\theta}),y)$ as an approximated differentiable logit margin loss by changing $\max$ into differentiable function $\mathrm{log}\sum_{k\neq y}e^{z_k(\bm{x})}$.
    Comparing OVR with $\ell_{\mathrm{SLM}}(\bm{z}(\bm{x},\bm{\theta}),y)$, we have the following: \looseness=-1
    \begin{align}\textstyle
      \ell_{\mathrm{SLM}}(\bm{z}(\bm{x},\bm{\theta}),y)\leq \ell_{\mathrm{CE}}(\bm{z}(\bm{x},\bm{\theta}),y)\leq \ell_{\mathrm{OVR}}(\bm{z}(\bm{x},\bm{\theta}),y).
    \end{align}
    This is because we have $\ell_{\mathrm{CE}}(\bm{z}(\bm{x}),y)-\ell_{\mathrm{SLM}}(\bm{z}(\bm{x}),y)\!=\!\mathrm{log}\sum_{k}e^{z_k(\bm{x})}\!-\!\mathrm{log}\sum_{k\neq y}e^{z_k(\bm{x})}\!=\!\mathrm{log}\left(1+e^{z_y(\bm{x})}/\sum_{k\neq y}e^{z_k(\bm{x})}\right)\geq 0$
    and $\ell_{\mathrm{CE}}(\bm{z}(\bm{x}),y)\!\leq \!\ell_{\mathrm{OVR}}(\bm{z}(\bm{x}),y)$ from Theorem~\ref{CEVsOVR}.
    Thus, we expect that OVR more strongly penalizes the small logit margins than $\ell_{\mathrm{SLM}}(\bm{x},y)$.
    Note that training algorithms of MMA is also different from those of the other importance-aware methods~\citep{MMA}.
    \subsection{EWAT~\citep{EWAT}}
    EWAT uses a weighted cross-entropy like GAIRAT and MAIL, but it is added to cross-entropy as 
    \begin{align}\textstyle
      \mathcal{L}_{\mathrm{weight}}(\bm{\theta})=\textstyle\frac{1}{N}\sum_{n=1}^N \left(1+\bar{w}_n\right)\ell_{\mathrm{CE}}(\bm{z}(\bm{x}_n^\prime,\!\bm{\theta}),\!y_n), 
    \end{align}
    where $\bar{w}_n\geq 0$ is a weight divided by batch-mean of the weight $w_n$ as $ \bar{w}_n=\frac{|\mathbb{B}|w_n}{\sum_{l=1}^{|\mathbb{B}|}{w}_l}$.
    EWAT determines the weights by using entropy as 
    \begin{align}\textstyle
      w_n=-\sum_{k=1}^K f_k (\bm{x}^\prime_n,\bm{\theta})\mathrm{log}(f_k (\bm{x}^\prime_n,\bm{\theta}))
    \end{align}
    where $f_k(\bm{x}_n,\bm{\theta})$ is the $k$-th softmax output, and thus, it can be regarded as the probability for the $k$-th class label.
    EWAT is based on the observation that importance-aware methods tend to have high entropy,
    and it causes their vulnerability.
     Our theoretical results about logit margins and experiments seem to indicate that a logit margin loss is a more reasonable criterion to evaluate the robustness and improve the robustness by using it than entropy.
    Furthermore, Theorem~\ref{lmthm} shows that the weighted cross-entropy is less effective than OVR at increasing logit margins.
   \section{Selection of $\phi$ in OVR}\label{PhiSelect}
   \subsection{Infinite Sample Consistency}
 Infinite-sample consistency (ISC, also known as \textit{classification calibrated} or \textit{Fisher consistent}) is a desirable property for multi-class classification problems~~\citep{SAForMC,bartlett2003convexity,lin2002support}.
 We first introduce ISC in this section.
 Let $\bm{f}(\bm{x})$ be a model and $c$ be the classifier $c(\bm{x})=\argmax_k f_k(\bm{x})$.
 The classification error $\ell_{*}$ is
 \begin{align}\textstyle
   \ell_{*}(c(\cdot)):=\mathbb{E}_{\bm{x}}\sum_{k=1,k\neq c(\bm{x})}^K a_k p(y=k|\bm{x})
   \label{ClassEr}
 \end{align}
 where $a_k$ is a weight for the $k$-th label and is usually set to one.
 The optimal classification rule called a Bayes rule is given by \looseness=-1
 \begin{align}\textstyle
   c_*(\cdot)=\max_{k\in \{1,\dots,K\}} a_k p(y=k|\bm{x}).
 \end{align}
 Since \req{ClassEr} is difficult to minimize directly, we use a surrogate loss function $\ell$.
 In classification problems, we obtain the model $\hat{f}(\cdot)$ by the minimization of the empirical risk using $\ell$ as 
 \begin{align}\textstyle
   \hat{f}(\cdot)=\mathrm{arg}\max_{\bm{f}(\cdot)}\frac{1}{n}\sum_{i=1}^{N} \ell(\bm{f}(\bm{x}_i),y_i).
 \end{align}
 On the other hand, the true risk using $\ell$ is written by
 \begin{align}\textstyle
   \mathbb{E}_{\bm{x},y}\ell(\bm{f},y)&\textstyle=\mathbb{E}_{\bm{x}}W(\bm{p}(\cdot |\bm{x}),\bm{f}(\bm{x}))    \label{TrueRisk}
   \\\textstyle
   W(\bm{q},\bm{f})&\textstyle:=\sum_{k=1}^{K}q_k\ell(\bm{f},k)
 \end{align}
 where $\bm{p}(\cdot|\bm{x})=[p(1|\bm{x}),\dots,p(K|\bm{x})]$ and 
 $\bm{q}$ is a vector in the set $\Lambda_K$:
 \begin{align}\textstyle
   \Lambda_K :=\left\{ \bm{q} \in \mathbb{R}^K: \sum_{k=1}^{K} q_k=1, q_k\geq 0\right\}.
 \end{align}
 $W(\bm{q}, \bm{f})$ is the point-wise true loss of model $\bm{f}$ with the conditional probability $\bm{q}$.
 By using the above, ISC is defined as the following definition:
 \begin{definition}{\citep{SAForMC}}
   We say that the formulation is infinite-sample consistent (ISC) on a set $\Omega \subseteq \mathbb{R}^K$ with respect to \req{ClassEr} if the following condition holds:
   \begin{itemize}
       \item For each $k$, $\ell(\cdot ,k):\Omega \rightarrow \mathbb{R}$ is bounded below and continuous
       \item $\forall \bm{q}\in\Lambda_K$ and $k\in \{1,\dots,K\}$ such that $a_k q_k<\sup_i a_i q_i$, we have
       \begin{align}\textstyle
           W^*(\bm{q}):=\inf_{\bm{f}\in \Omega }W(\bm{q},\bm{f})<\inf\left\{ W(\bm{q},\bm{f})|\bm{f}\in \Omega,f_{k}=\sup_i f_i \right\}
       \end{align}
   \end{itemize}
 \end{definition}
 This definition indicates that the optimal solution of $W(\bm{q}, \cdot)$ leads to a Bayes rule with respect to classification error~\cite{SAForMC}:
 the minimizer of \req{TrueRisk} becomes the minimizer of classification error $\ell_{*}$ (\req{ClassEr}).
 Thus, surrogate loss functions $\ell$, e.g., cross-entropy or OVR, should satisfy ISC to minimize the classification error.
 \subsection{Evaluation of $\phi$}
 It is known that ISC is satisfied when  $\phi$ in \req{OVRRaw} is a differentiable non-negative convex function and satisfies $\phi(z) < \phi(-z)$ for $z > 0$.
 Among common nonlinear functions used in deep neural networks, 
 $e^{-z}$ and $\log (1+e^{-z})$ satisfy this condition. We first evaluated $e^{-z}$ and observed that $e^{-z}$ causes numerical unstability.
 On the other hand, $\log (1+e^{-z})$ tends to be stable in computation.
 This is because $\log (1+e^{-z})$ asymptotically closes to $\max (-z, 0)$.
 In addition, let the conditional probability $p(y|\bm{x})$ for the class $k$ given $\bm{x}$ be 
 \begin{align}
   p(k|\bm{x})=\frac{1}{1+e^{-z_k(\bm{x})}}
 \end{align}
 when we choose $\phi(z)=\log (1+e^{-z})$.
 We have
 \begin{align}\textstyle
   \ell_{\mathrm{OVR}}(\bm{z}(\bm{x},\bm{\theta}),y)\textstyle\!&=\!\mathrm{log} (1\!+\!e^{-z_{y}(\bm{x})})+\sum_{k\neq y}\mathrm{log}(1\!+\!e^{z_{k}(\bm{x})})\\
   &\textstyle=\!- \mathrm{log} p(y|\bm{x})+\sum_{k\neq y}-\mathrm{log}(1-p(k|\bm{x}))
 \end{align}
 and we can regard models as $K$ independent binary classifier~\citep{padhy2020revisiting}.
 \section{Experimental Setups}\label{SetupSec}
    We conducted the experiments for evaluating our proposed method.
    We first compared our method with baseline methods; Madry's AT~\citep{pgd2}, MMA~\citep{MMA}, MART~\citep{MART},
    GAIRAT~\citep{GAIRAT}, MAIL~\citep{MAIL}, and EWAT~\citep{EWAT} on three datasets; CIFAR10, SVHN, and CIFAR100~\citep{cifar,svhn}.
    Next, we evaluated the combination of our method with TRADES~\citep{TRADES}, AWP~\citep{AWP}, and SEAT~\citep{SEAT}.
    Our experimental codes are based on source codes provided by \cite{AWP,MART,MMA}.
    We used PreActResNet-18 (RN18)~\citep{resnet} and WideResNet-34-10 (WRN)~\citep{WRN} following~\cite{AWP}.
    The $L_\infty$ norm of the perturbation was set to $\varepsilon\!=\!8/255$, and all elements of $x_i+\delta_i$ were clipped so that they were in [0,1].
    We used early stopping by evaluating test robust accuracies against 20-step PGD.
    To evaluate TRADES, AWP, and SEAT, we used the original public code~\citep{AWP,SEAT}.
    We trained models three times and show the average and standard deviation of test accuracies.
    We used Auto-Attack to evaluate the robust accuracy on test data.
    We used one GPU among NVIDIA \textregistered V100 and NVIDIA\textregistered A100
    for each training in experiments.
    We trained models three times and show the average and standard deviation of test accuracies.
    For MART, we used \textit{mart\_loss} in the original code~\citep{MART}\footnote{\url{https://github.com/YisenWang/MART}} as the loss function.
    $\lambda$ of MART was set to 6.0.
    For GAIRAT and MAIL, we also used the loss functions in the original codes~\citep{GAIRAT,MAIL},\footnote{\url{https://github.com/zjfheart/Geometry-aware-Instance-reweighted-Adversarial-Training}}\footnote{\url{https://github.com/QizhouWang/MAIL}}
    and thus, hyperparameters of their loss functions were based on them.
    $\lambda$ of GAIRAT was set to $\infty$ until the 50-th epoch and then set to 3.0.
    $(\gamma,\beta)$ of MAIL was set to $(10, 0.5)$.
    For all settings, the size of minibatch was set to 128.
    The detailed setup for each dataset was as follows.
    \subsection{MMA}
    We trained models by using MMA based on the original code~\citep{MMA}\footnote{\url{https://github.com/BorealisAI/mma_training}}.
    Thus, the learning rate schedules and hyperparameters of PGD for MMA were different from those for other methods
    because the training algorithm of MMA is different from the other methods.
    The step size of PGD in MMA was set to $\frac{2.5\varepsilon}{10}$ in training by following~\cite{MMA}.
    For AN-PGD in MMA, the maximum perturbation length was 1.05 times the hinge threshold $\varepsilon_{\mathrm{max}}=1.05d_{\mathrm{max}}$,
    and $d_{\mathrm{max}}$ was set to 0.1255. 
    The learning rate of SGD was set to 0.3 at the 0-th parameter update, 0.09 at the 20000-th parameter update, 0.03 at  the 30000-th parameter update, and 0.009 at the 40000-th parameter update.
    \subsection{CIFAR10}
    For PreActResNet18, the learning rate of SGD was divided by 10 at the 100-th and 150-th epoch except for EWAT, and the initial learning rate was set to 0.05 for SOVR and 0.1 for others.
    We tested the initial learning rate of 0.05 for the other methods and found that the setting of 0.1 achieved better robust accuracies against Auto-Attack than the setting of 0.05 when using ResNet18.
    For EWAT, we divided the learning rate of SGD at the 100-th and 105-th epoch following~\cite{EWAT} after we found that the division at the 100-th and 150-th epoch was worse than the division at the 100-th and 105-th epoch.
    When using WideResNet34-10, we set the initial learning rate to 0.1 and divided by 10 at the 100-th and 150-th epoch.
    We used momentum of 0.9 and weight decay of 0.0005 and early stopping by evaluating test accuracies. 
    We standardized datasets by using mean = [0.4914, 0.4822, 0.4465] and std = [0.2471, 0.2435, 0.2616] as the pre-process.
    $(M,\lambda)$ was tuned by grid search over $M\in [20,\dots,80,100]$ and $\lambda \in [0.2,\dots,0.8,1.0]$ for RN18, and tuned by coarse tuning for WRN due to high computation cost.
   \subsection{CIFAR100}
    We used PreActResNet18 for CIFAR100.
    The learning rate of SGD was divided by 10 at the 100-th and 150-th epoch except for EWAT, and the initial learning rate was set to 0.1.
    Note that we found that the above setting is better than the initial learning rate of 0.05 for all methods. 
    For EWAT, we divided the learning rate of SGD at the 100-th and 105-th epoch following~\cite{EWAT} after we found that the division at the 100-th and 150-th epoch was worse than the division at the 100-th and 105-th epoch.
    We randomly initialized the perturbation and updated it for 10 steps with a step size of 2/255 for PGD.
    We used momentum of 0.9 and weight decay of 0.0005 and early stopping by evaluating test accuracies.
    We standardized datasets by using mean = [0.5070751592371323, 0.48654887331495095, 0.4409178433670343],
    and std = [0.2673342858792401, 0.2564384629170883, 0.27615047132568404] as the pre-process.
    $(M,\lambda)$ was set to (50, 0.6) based on the coarse hyperparameter tuning.
    \subsection{SVHN}
    We used PreActResNet18 for SVHN.
    The learning rate of SGD was divided by 10 at the 100-th and 150-th, and the initial learning rate was set to 0.05 for SOVR and 0.01 for others.
    We tested the initial learning rate of 0.05 for the other methods and found that the setting of 0.01 achieved better robust accuracies against Auto-Attack than the setting of 0.05.
    For EWAT, the learning rate of SGD was divided by 10 at the 100-th and 105-th epoch after we found that this setting was better than the division at 100-th and 105-th epoch. 
    The hyperparameters for PGD were based on \citep{AWP}:
    We randomly initialized the perturbation and updated it for 10 steps with a step size of 1/255.
    For the preprocessing,
    we standardized data by using the mean of 
    [0.5, 0.5, 0.5],
    and standard deviations of [0.5, 0.5, 0.5].
    $(M,\lambda)$ is set to (20,0.2) on the basis of the coarse hyperparameter tuning.
    \subsection{TRADES, AWP, SEAT, and Data augmentation by using synthetic data}
    For experimental settings of TRADES and AWP, we followed \cite{AWP} and only changed the training loss into SOVR in the training procedure and in the algorithm for computing the perturbation of AWP for SOVR+AWP, TSOVR, and TSOVR+AWP.
    We used the original codes of AWP~\citep{AWP}\footnote{\url{https://github.com/csdongxian/AWP}}.
    For AWP and AWP+SOVR, we found that models trained under the setup for TRADES+AWP in original codes, where the dataset is not standardized and AWP is applied after 10 epochs,
    achieves better robust accuracy than those trained under the setup for cross-entropy+TRADES in original codes.
    Thus, we used the code for TRADES+AWP by changing the loss functions. 
    $\beta_T$ of TRADES and TSOVR were set to 6, and $\gamma$ of AWP is set to 0.01 for AWP and AWP+SOVR.
    $\gamma$ of AWP was set to 0.005 for TRADES+AWP and TSOVR+AWP. AWP is applied after the 10-th epoch.
    We used WideResNet34-10 following \cite{AWP}. We used SGD with momentum of 0.9 and weight decay of 0.0005 for 200 epochs.
    The learning rate was set to 0.1 and was divided by 10 at the 100-th and 150-th epoch.
    For experimental settings of SEAT, we followed~\cite{SEAT} and only changed the loss into SOVR in the original code~\citep{SEAT}.\footnote{\url{https://github.com/whj363636/Self-Ensemble-Adversarial-Training}}
    We did not evaluate SEAT with CutMix in our experiments, but we fairly compare SEAT+SOVR with SEAT under the same condition.
    We used SGD with momentum of 0.9 and weight decay of $7\times 10^{-4}$ for 120 epochs.
    The initial learning rate was set to 0.1 till the 40-th epoch and then linearly reduced to 0.01 and 0.001 at the 60-th epoch and 120-th epoch, respectively.
    We used WideResNet32-10 following \cite{SEAT} for SEAT.
    $(M,\lambda)$ is tuned by grid search over $M\in [20,\dots,80,100]$ and $\lambda \in [0.2,\dots,0.8,1.0]$ for SOVR+AWP,
    and $(M,\lambda)$ is tuned by grid search over $M\in [20,\dots,80,100]$ and $\lambda \in [0.2,\dots,1.0,1.2]$ for TSOVR.
    $(M,\lambda)$ is set to (50,0.5) for SOVR+SEAT after coarse hyperparameter tuning.
    For experimental settings of data augmentation using 1M synthetic data by DDPM,
    we used the code of \cite{rade2021pytorch}.
    We only changed the maximum learning rate for SOVR and set to 0.2.
    Following \cite{rade2021pytorch}, label smoothing with 0.1 is applied to cross-entropy in SOVR and TSOVR.
    \subsection{Experimental Setups in Section~\ref{vulsec}}\label{ExTochuSec}
    For the experiments in Section~\ref{vulsec}, we used the models obtained under the above settings, which are the same as models used in Section 6.
    To obtain the histograms of logit margins, we used the models and computed logit margin loss on adversarial examples of training data set for each data point at 200 epochs.
    Thus, the number of data points of CIFAR10 and CIFAR100 is 50,000, and that of SVHN is 73,257.
    \section{Additional Results}\label{AddSec}
    \subsection{Histograms of Logit Margin Losses}\label{HisSub}
\begin{figure}[tbp]
      \centering
      \begin{subfigure}[t]{0.23\linewidth}
       \includegraphics[width=\linewidth]{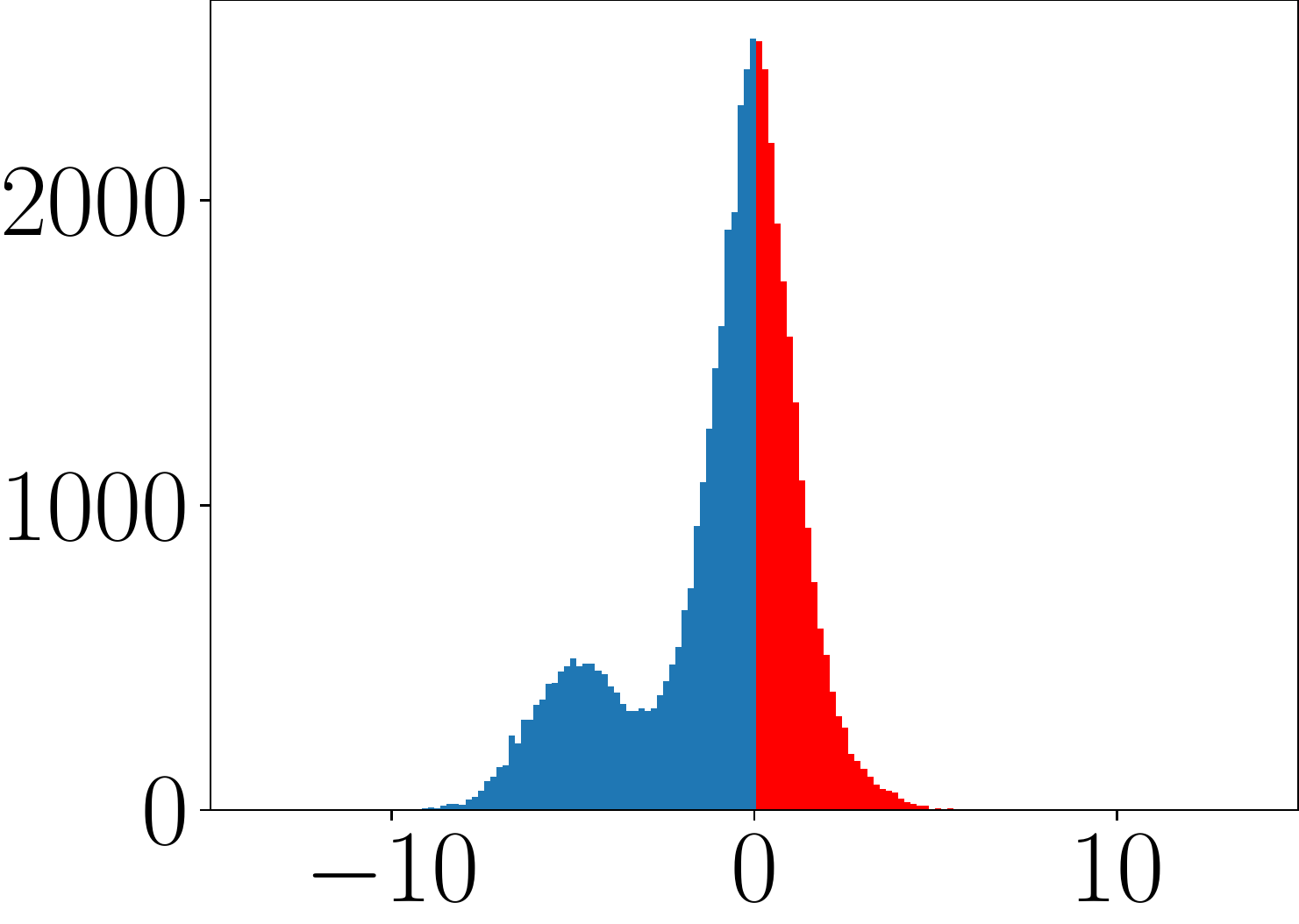}
       \caption{Best epoch}
     \end{subfigure}
      \begin{subfigure}[t]{0.23\linewidth}
       \includegraphics[width=\linewidth]{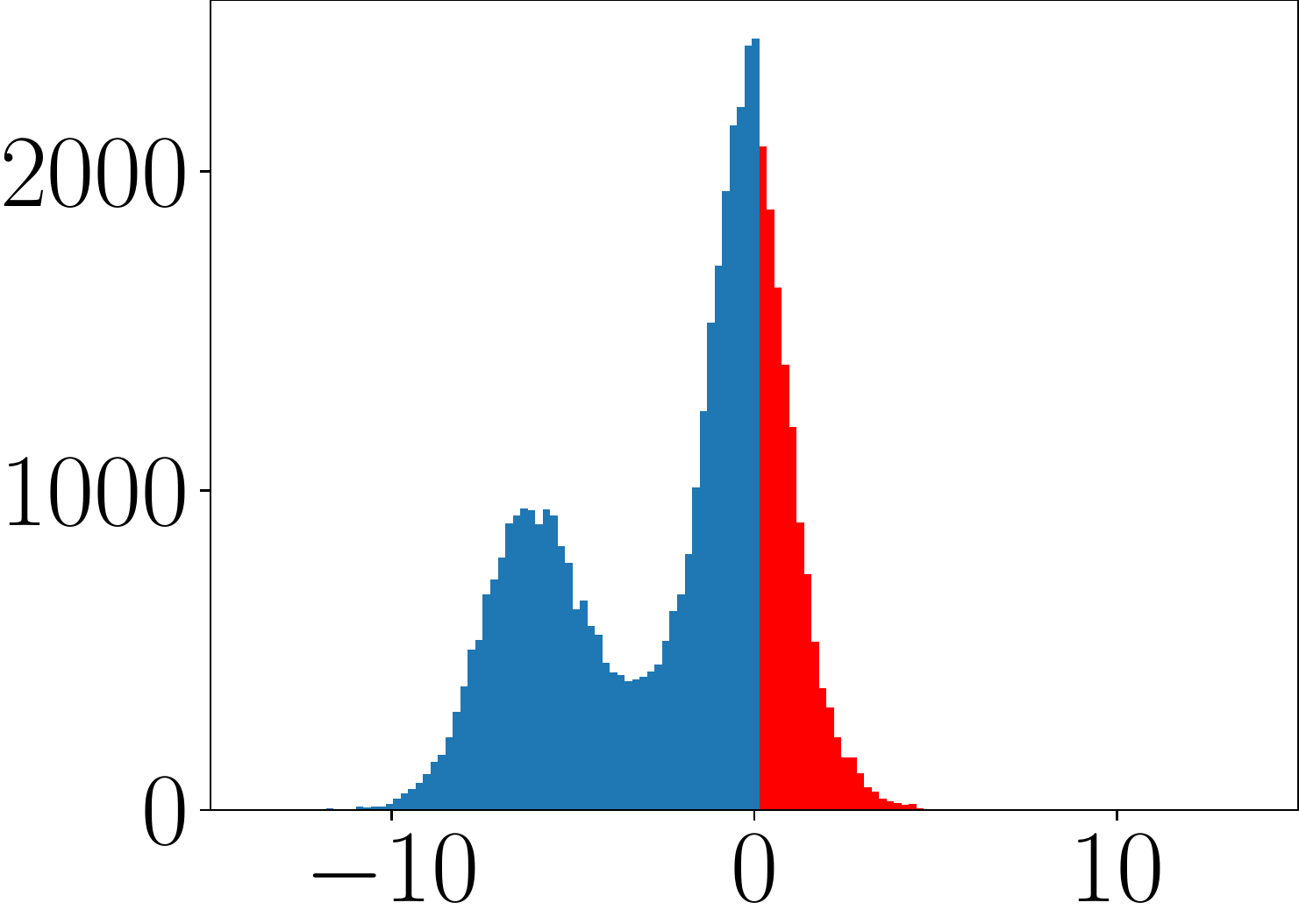}
       \caption{Last epoch}
     \end{subfigure}
      \caption{Histogram of logit margin losses of EWAT for training data on CIFAR10.}\label{LMHist-EWAT}
    \end{figure}

\begin{figure}[tbp]
      \begin{subfigure}[t]{0.23\linewidth}
        \includegraphics[width=\linewidth]{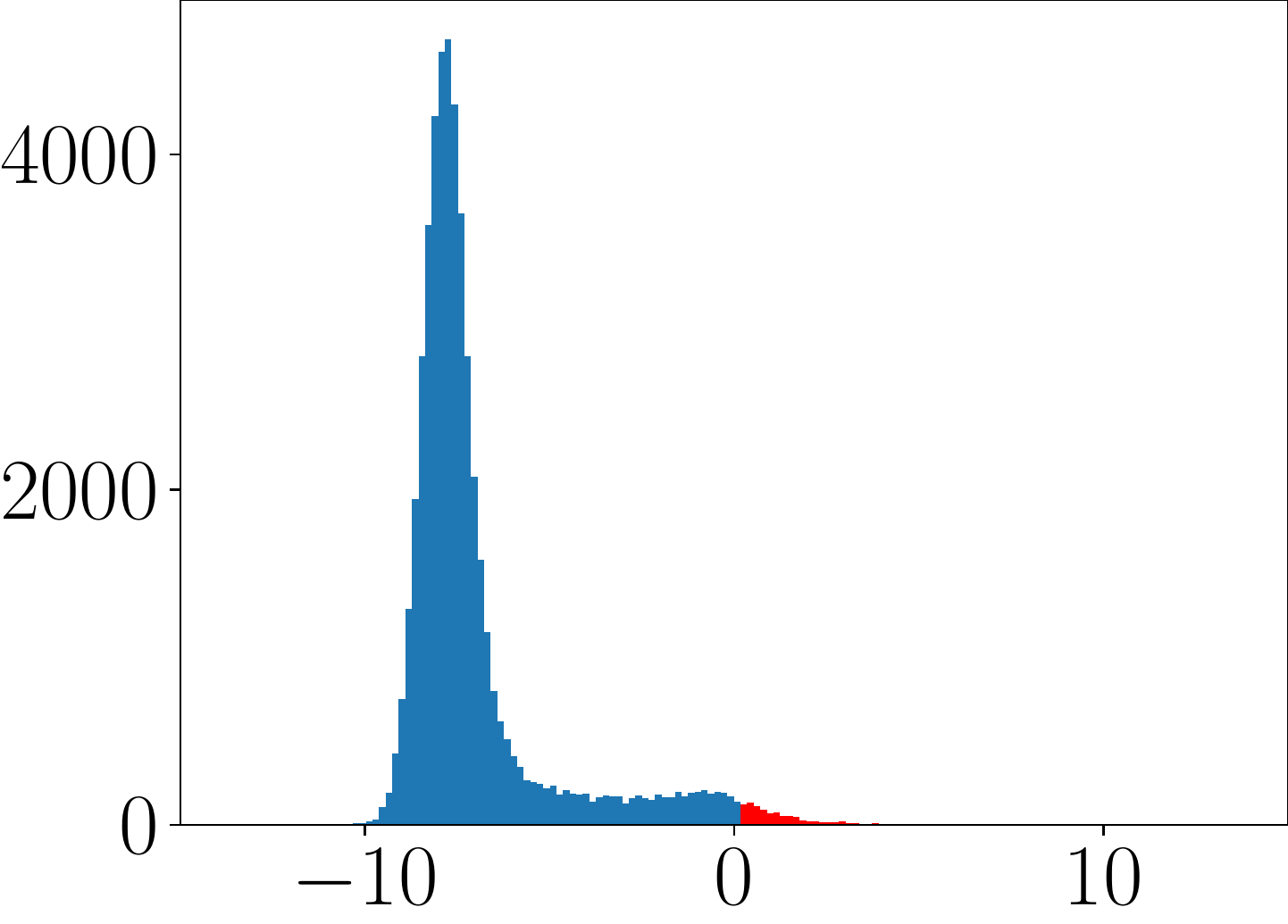}
        \caption{AT}\label{C10WRN-LMHist-AT}
    \end{subfigure}
    \begin{subfigure}[t]{0.22\linewidth}
      \includegraphics[width=\linewidth]{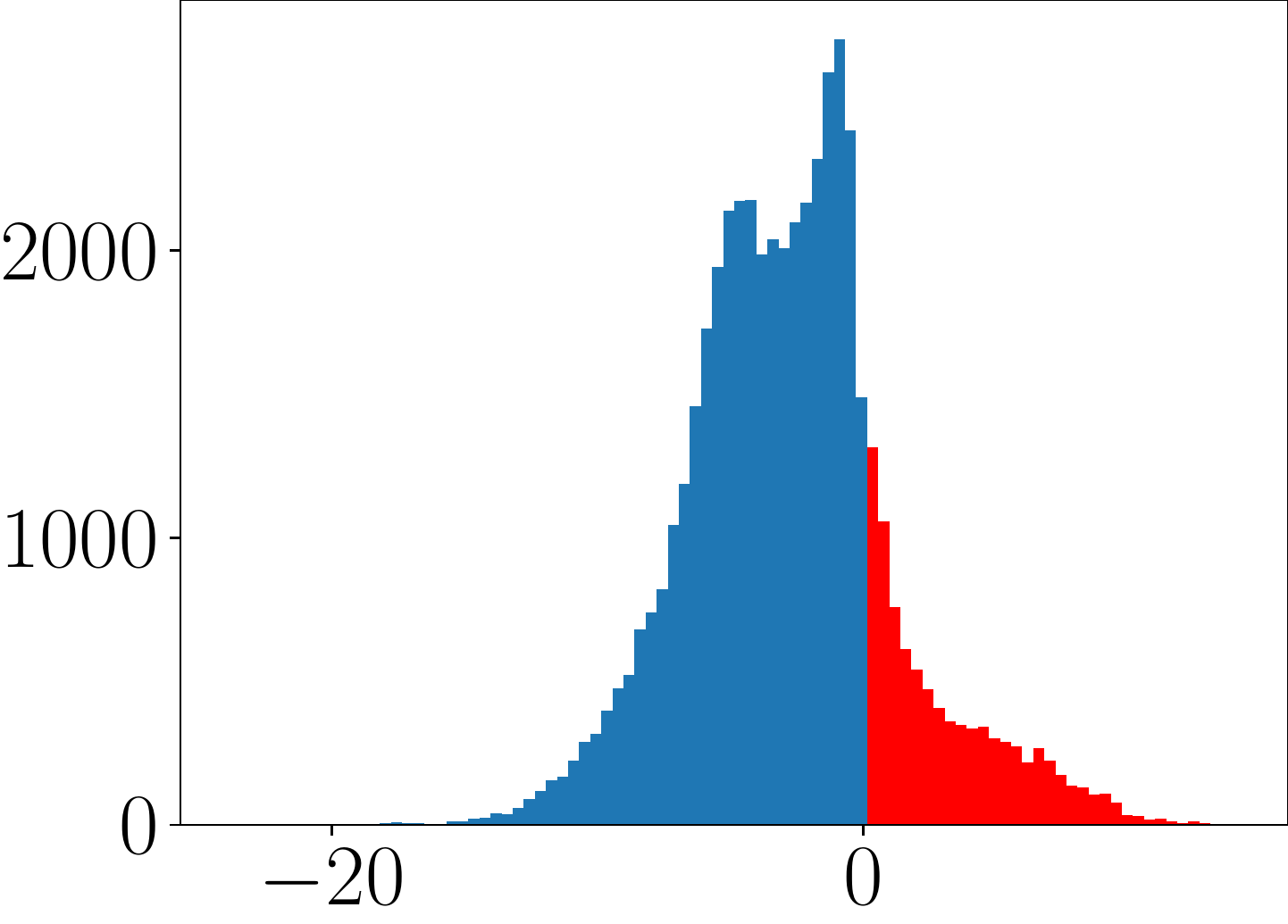}
      \caption{MMA}\label{C10WRN-LMHist-MMA}
    4\end{subfigure}
    \begin{subfigure}[t]{0.23\linewidth}
      \includegraphics[width=\linewidth]{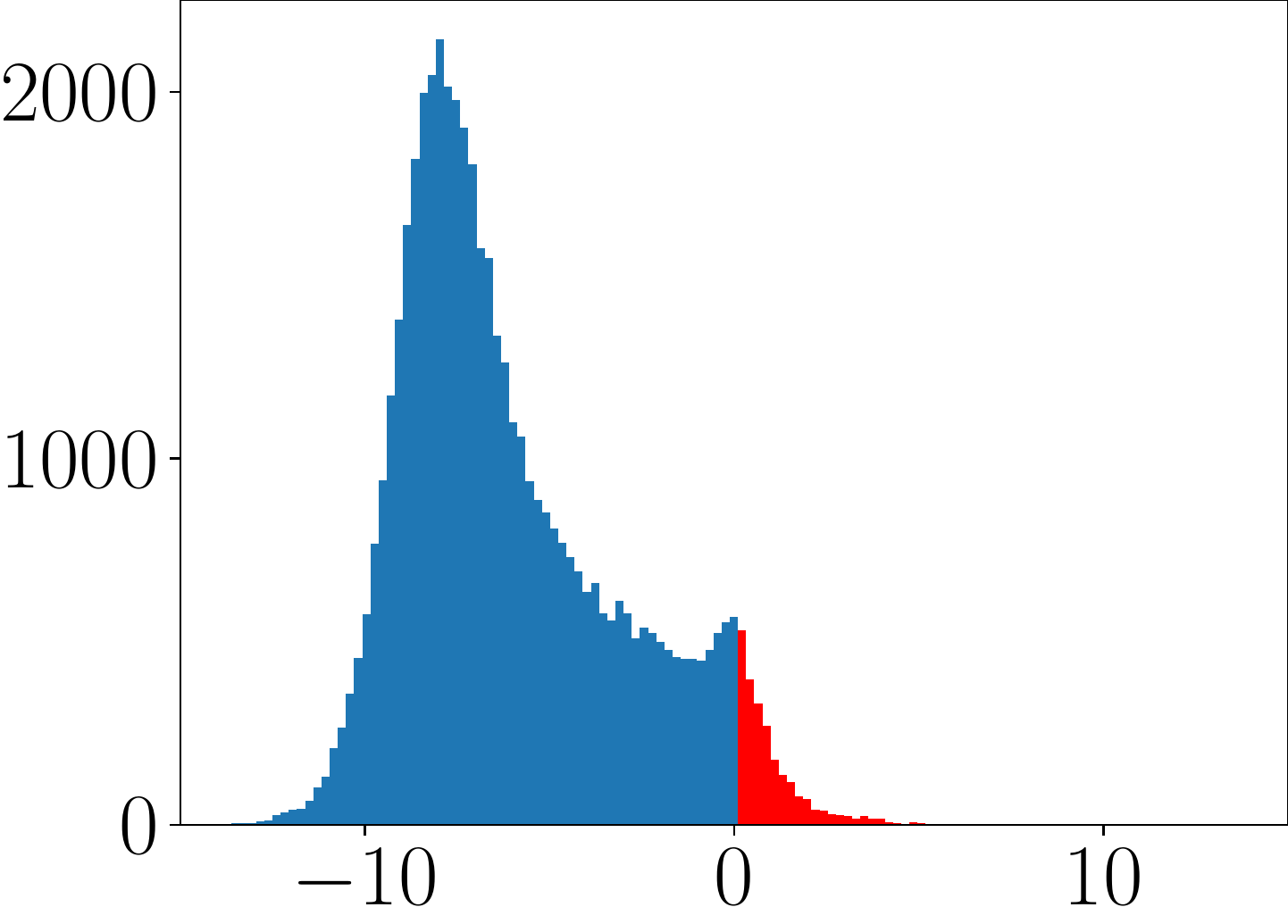}
      \caption{MART}\label{C10WRN-LMHist-MART}
    \end{subfigure}
    \begin{subfigure}[t]{0.23\linewidth}
      \includegraphics[width=\linewidth]{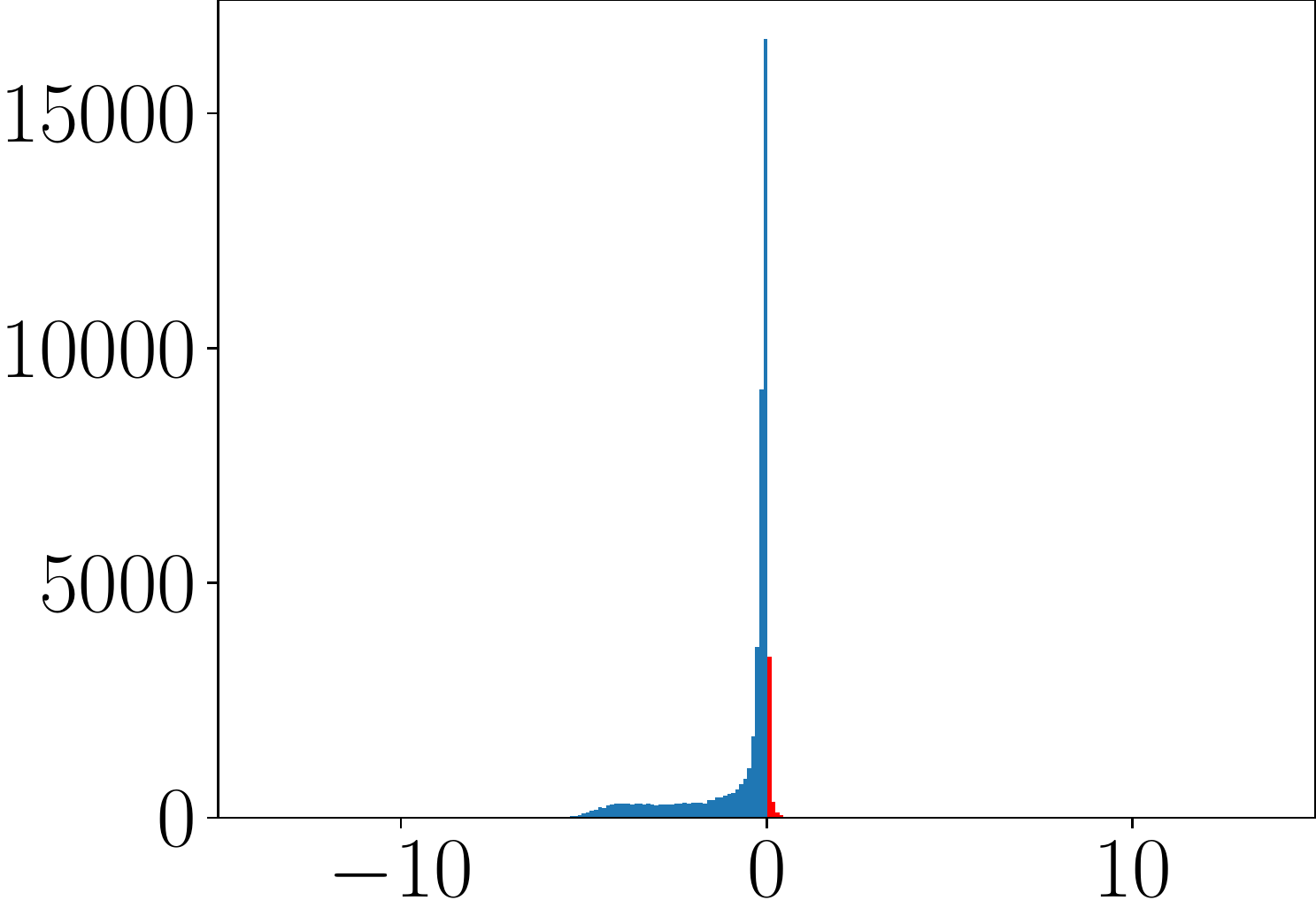}
      \caption{GAIRAT}\label{C10WRN-LMHist-GAIRAT}
    \end{subfigure}\\
    \centering
    \begin{subfigure}[t]{0.23\linewidth}
      \includegraphics[width=\linewidth]{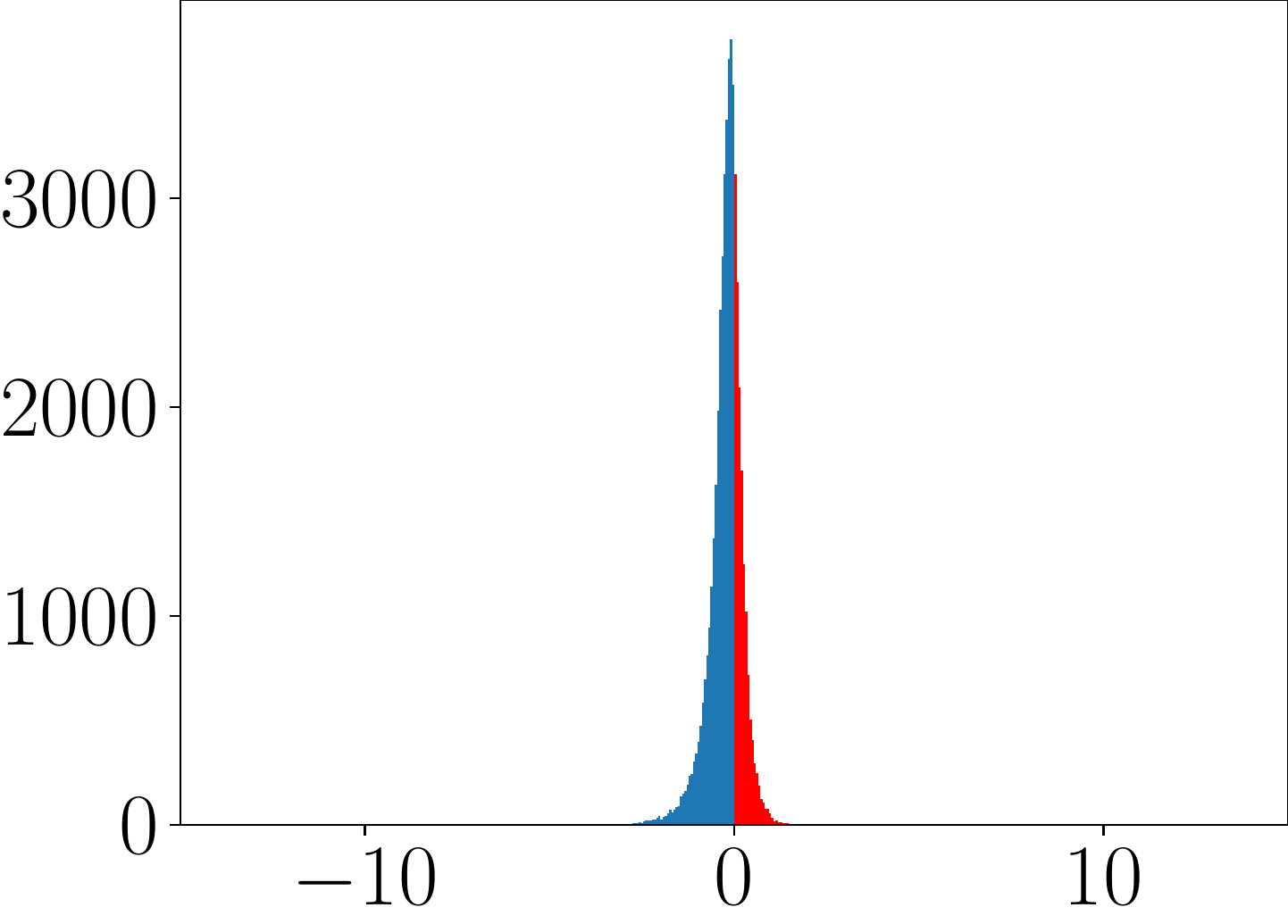}
      \caption{MAIL}\label{C10WRN-LMHist-MAIL}
    \end{subfigure}
    \begin{subfigure}[t]{0.23\linewidth}
      \includegraphics[width=\linewidth]{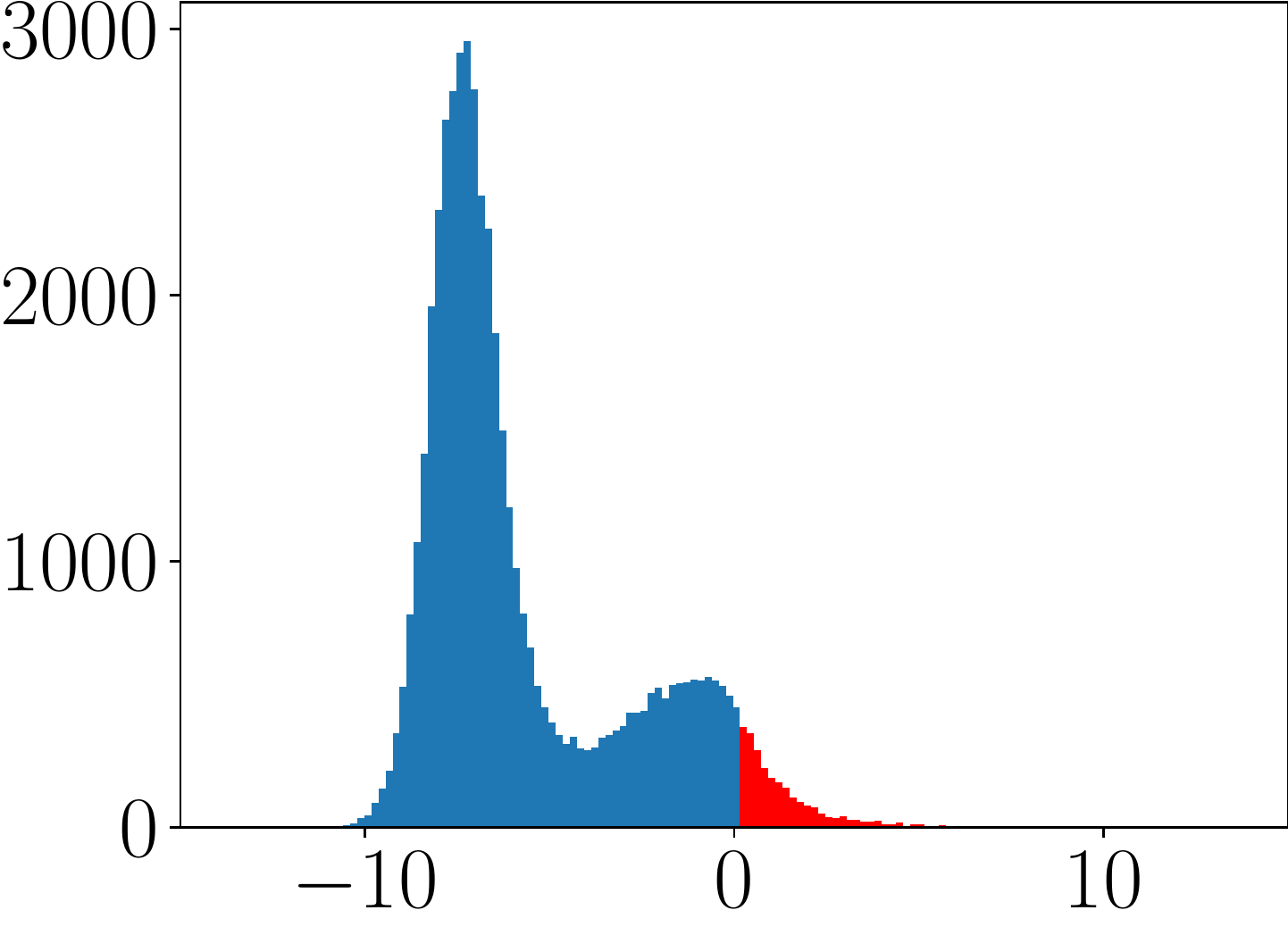}
      \caption{EWAT}\label{C10WRN-LMHist-EWL}
    \end{subfigure}
    \begin{subfigure}[t]{0.23\linewidth}
      \includegraphics[width=\linewidth]{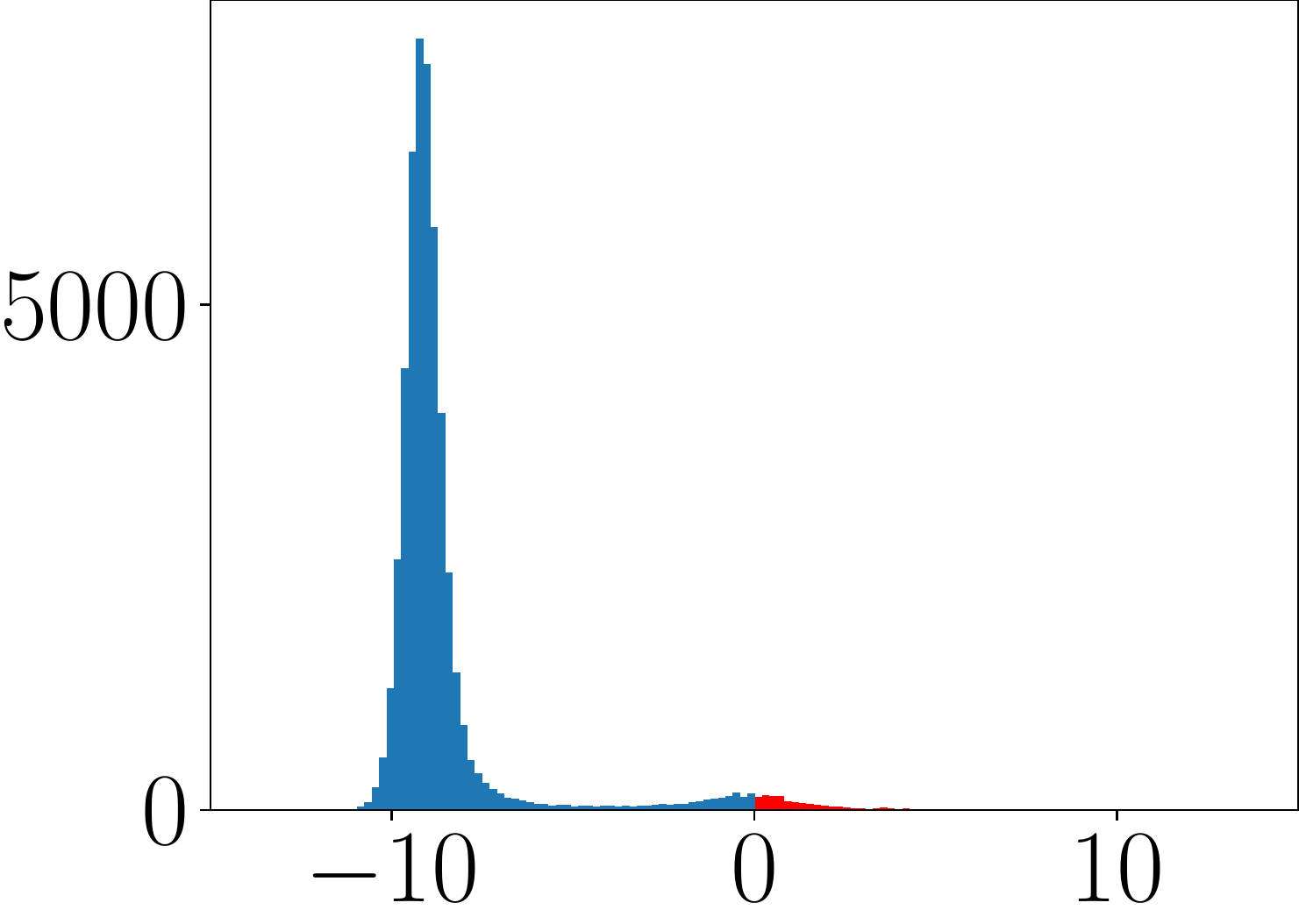}
      \caption{SOVR}\label{C10WRN-LMHist-SOVR}
    \end{subfigure}
    \caption{Histogram of logit margin losses for training samples of CIFAR10 with WideResNet34-10 at the last epoch. 
      We plot those on adversarial data $\bm{x}^\prime$ for the other methods. Blue bins are the data points that models correctly classify.}
      \label{C10WRN-LMHist}
    \end{figure}

\begin{figure}[tbp]
      \begin{subfigure}[t]{0.23\linewidth}
        \includegraphics[width=\linewidth]{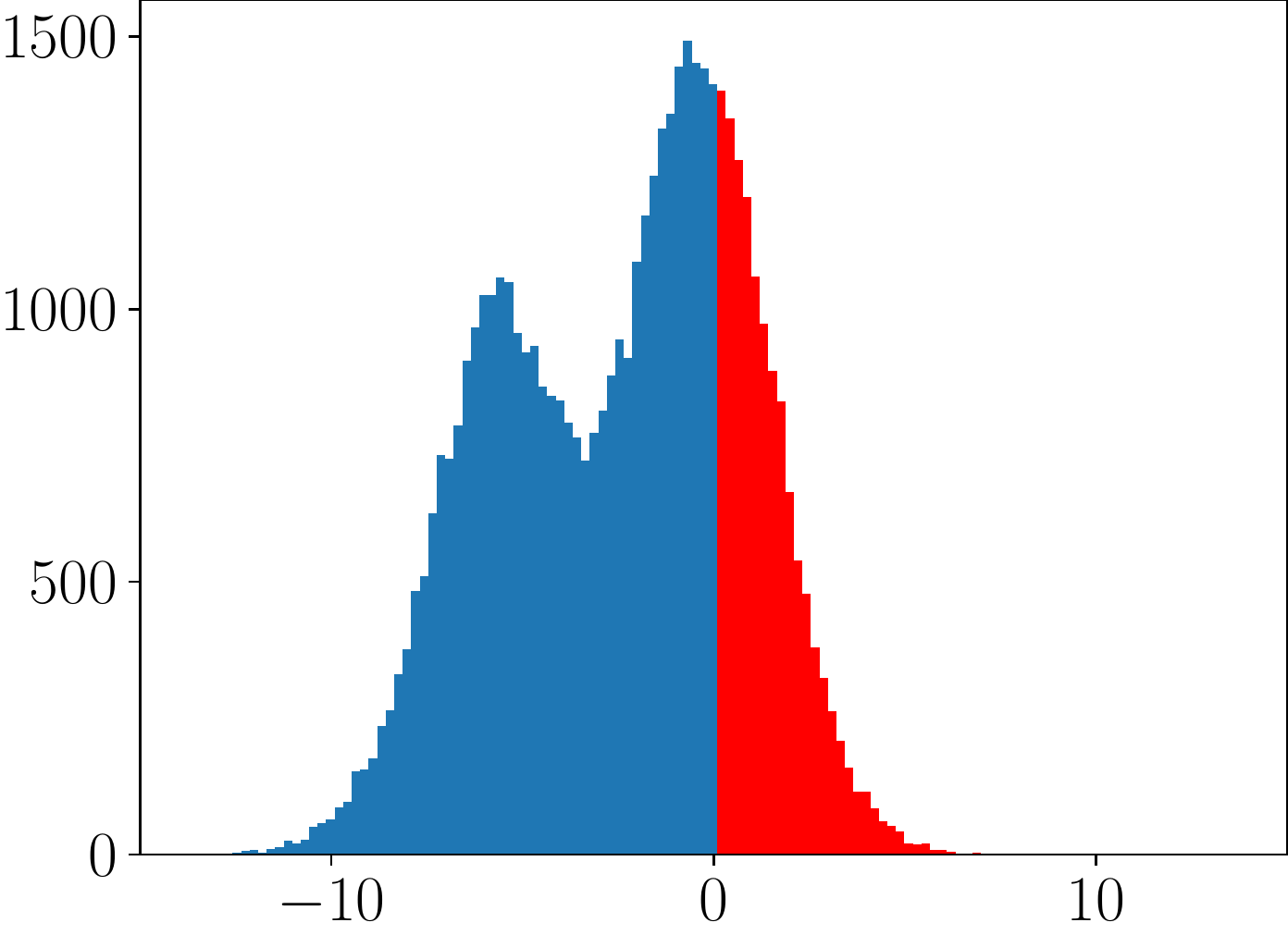}
        \caption{AT}\label{C100-LMHist-AT}
    \end{subfigure}
    \begin{subfigure}[t]{0.23\linewidth}
      \includegraphics[width=\linewidth]{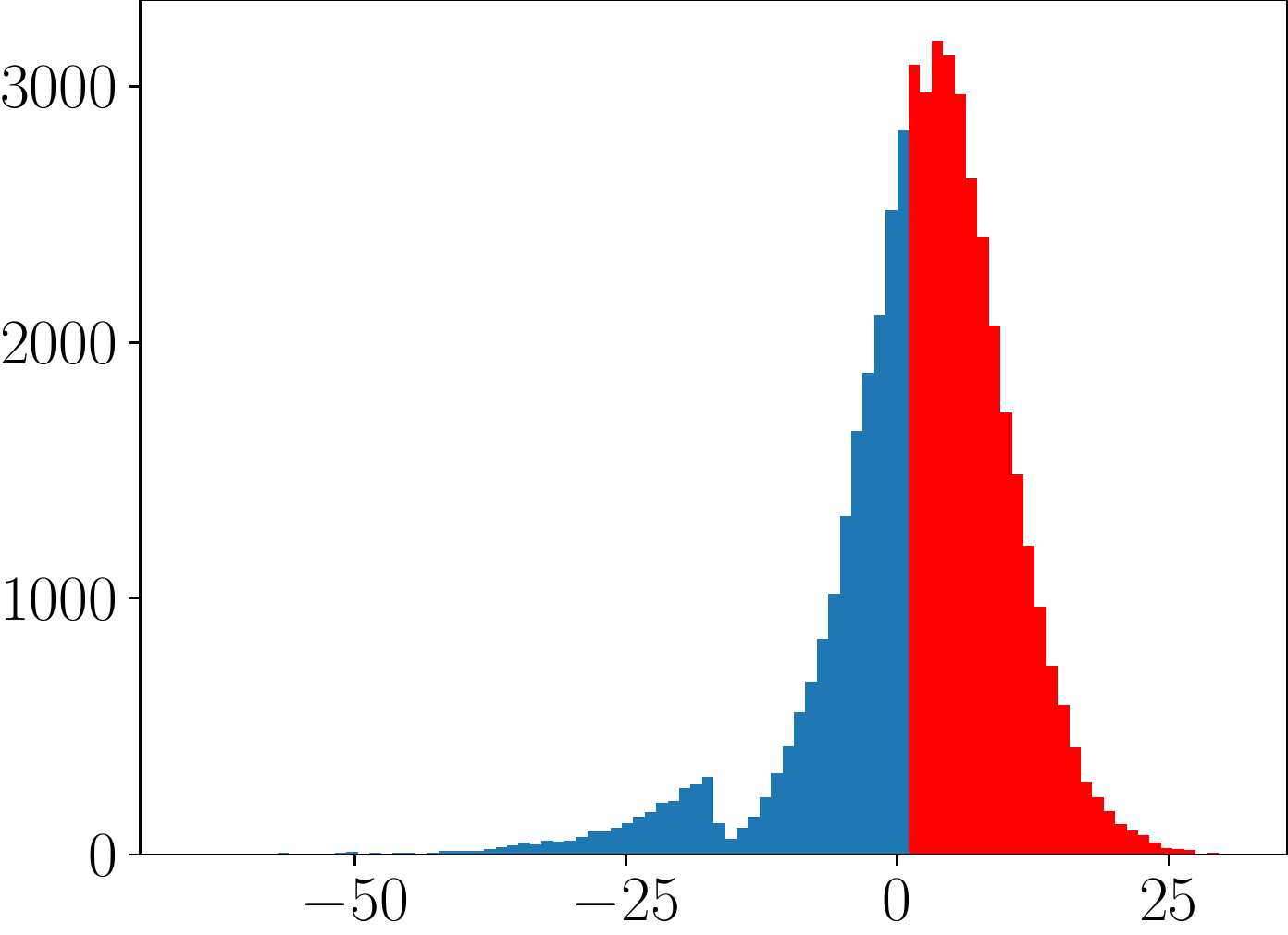}
      \caption{MMA}\label{C100-LMHist-MMA}
    4\end{subfigure}
    \begin{subfigure}[t]{0.23\linewidth}
      \includegraphics[width=\linewidth]{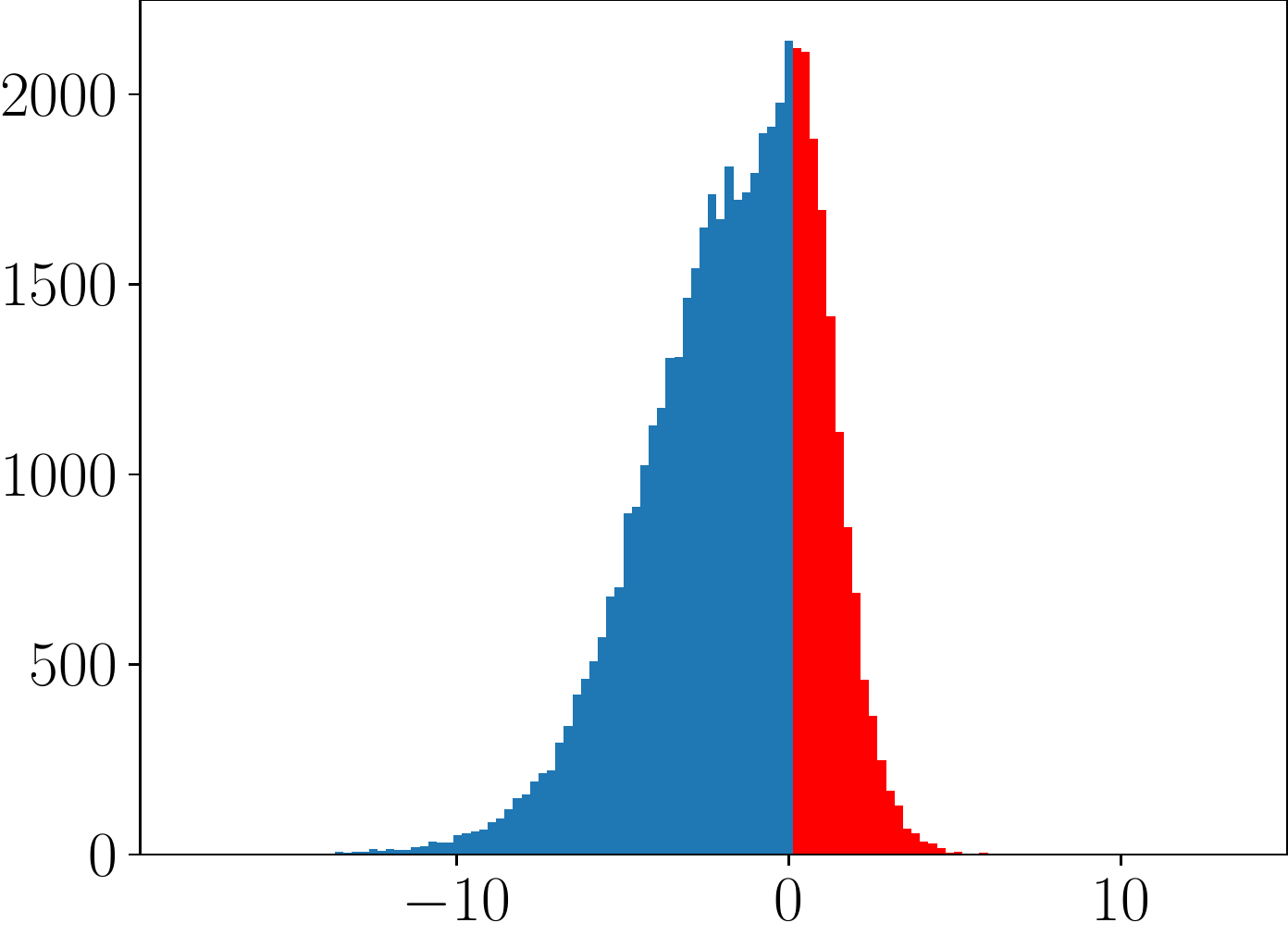}
      \caption{MART}\label{C100-LMHist-MART}
    \end{subfigure}
    \begin{subfigure}[t]{0.23\linewidth}
      \includegraphics[width=\linewidth]{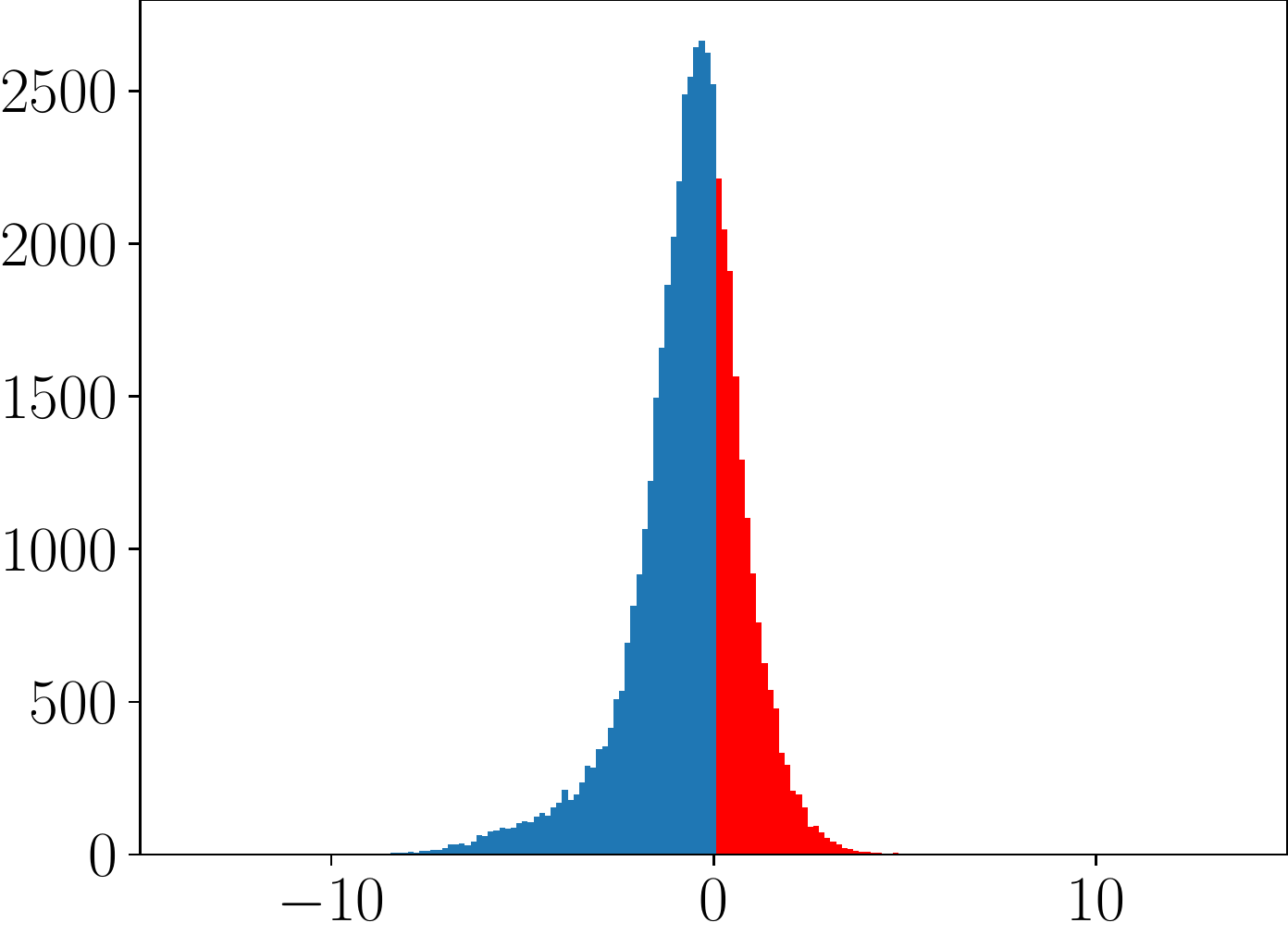}
      \caption{GAIRAT}\label{C100-LMHist-GAIRAT}
    \end{subfigure}\\
    \centering
    \begin{subfigure}[t]{0.23\linewidth}
      \includegraphics[width=\linewidth]{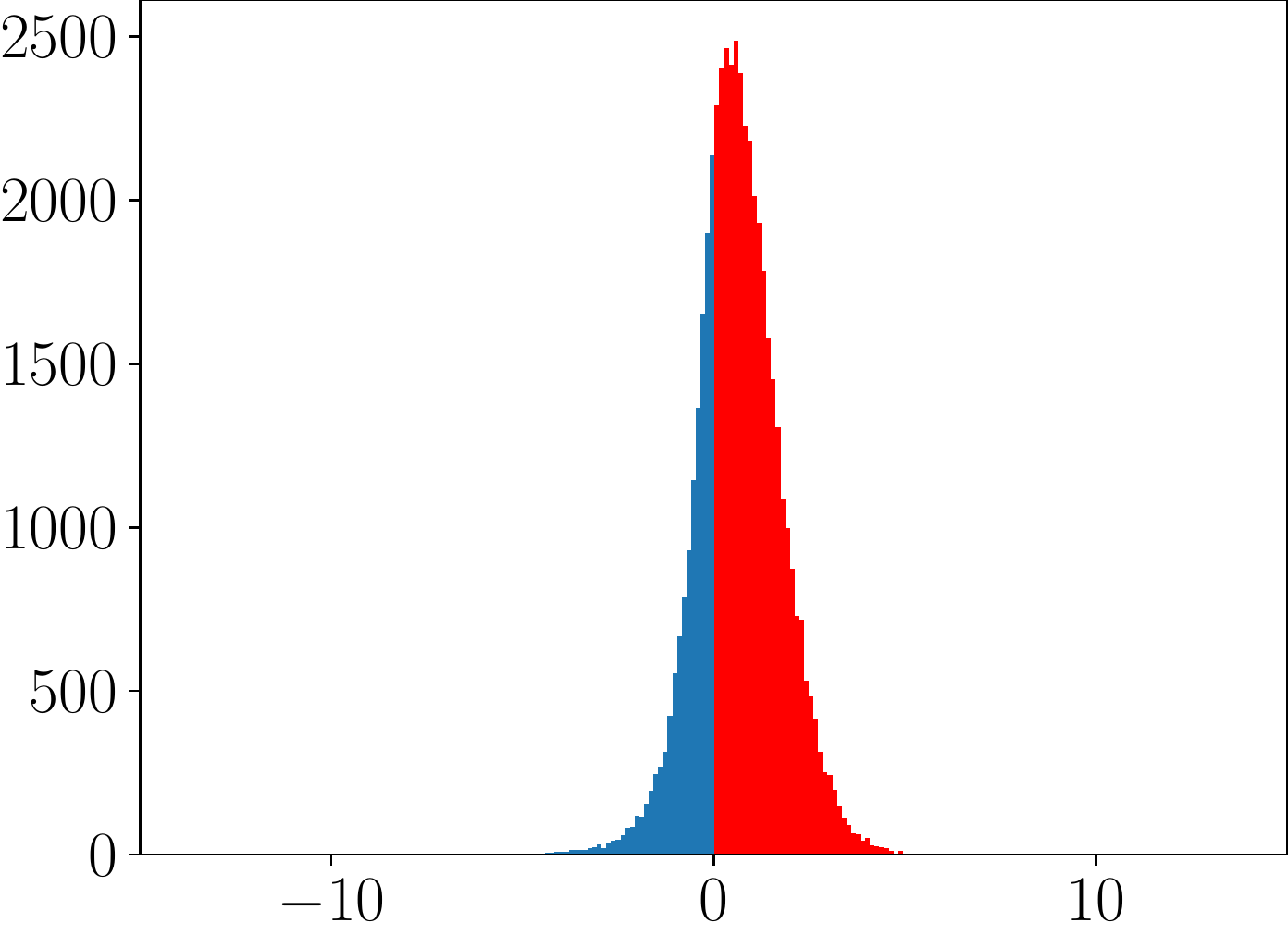}
      \caption{MAIL}\label{C100-LMHist-MAIL}
    \end{subfigure}
    \begin{subfigure}[t]{0.23\linewidth}
      \includegraphics[width=\linewidth]{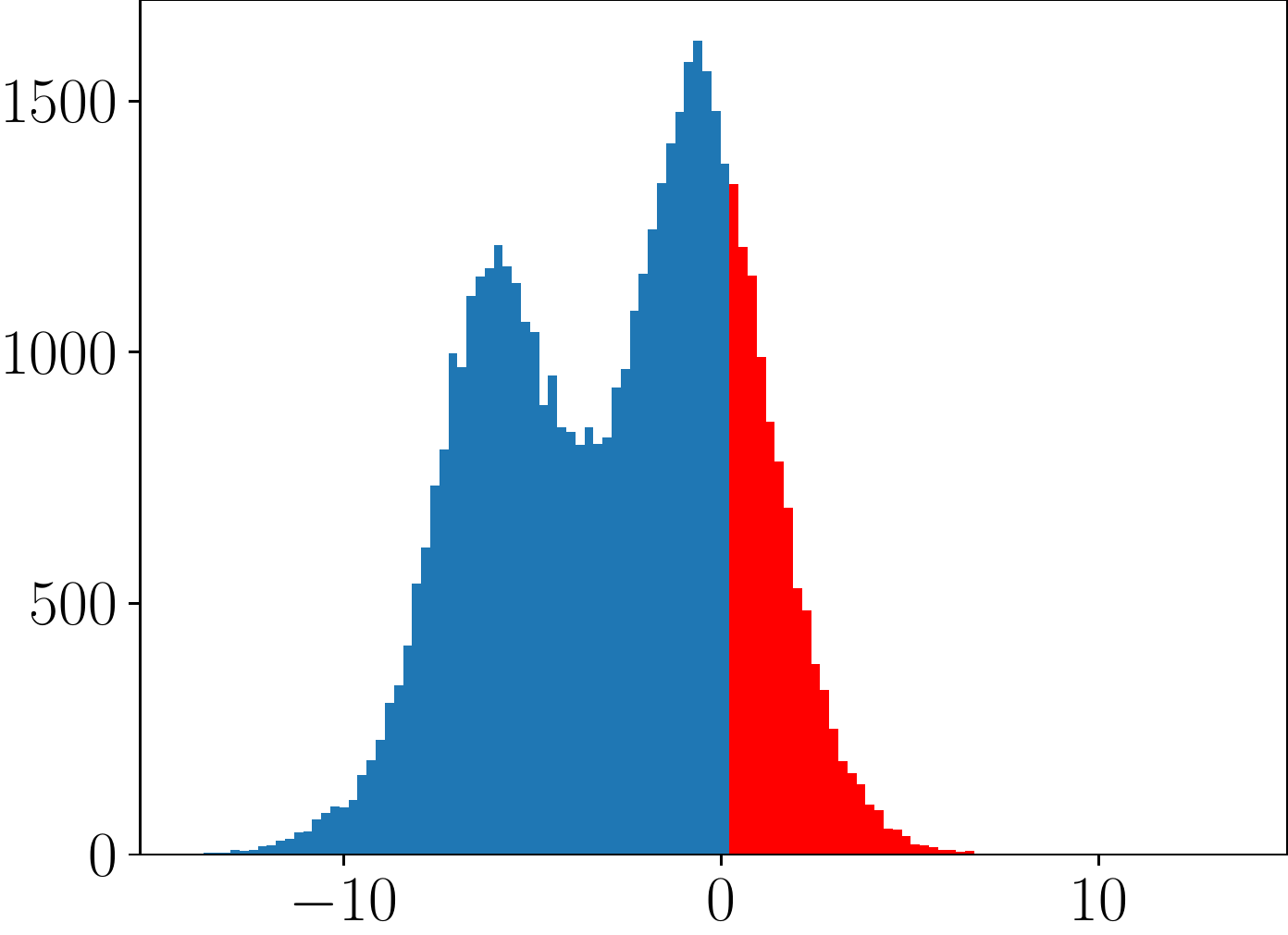}
      \caption{EWAT}\label{C100-LMHist-EWL}
    \end{subfigure}
    \begin{subfigure}[t]{0.23\linewidth}
      \includegraphics[width=\linewidth]{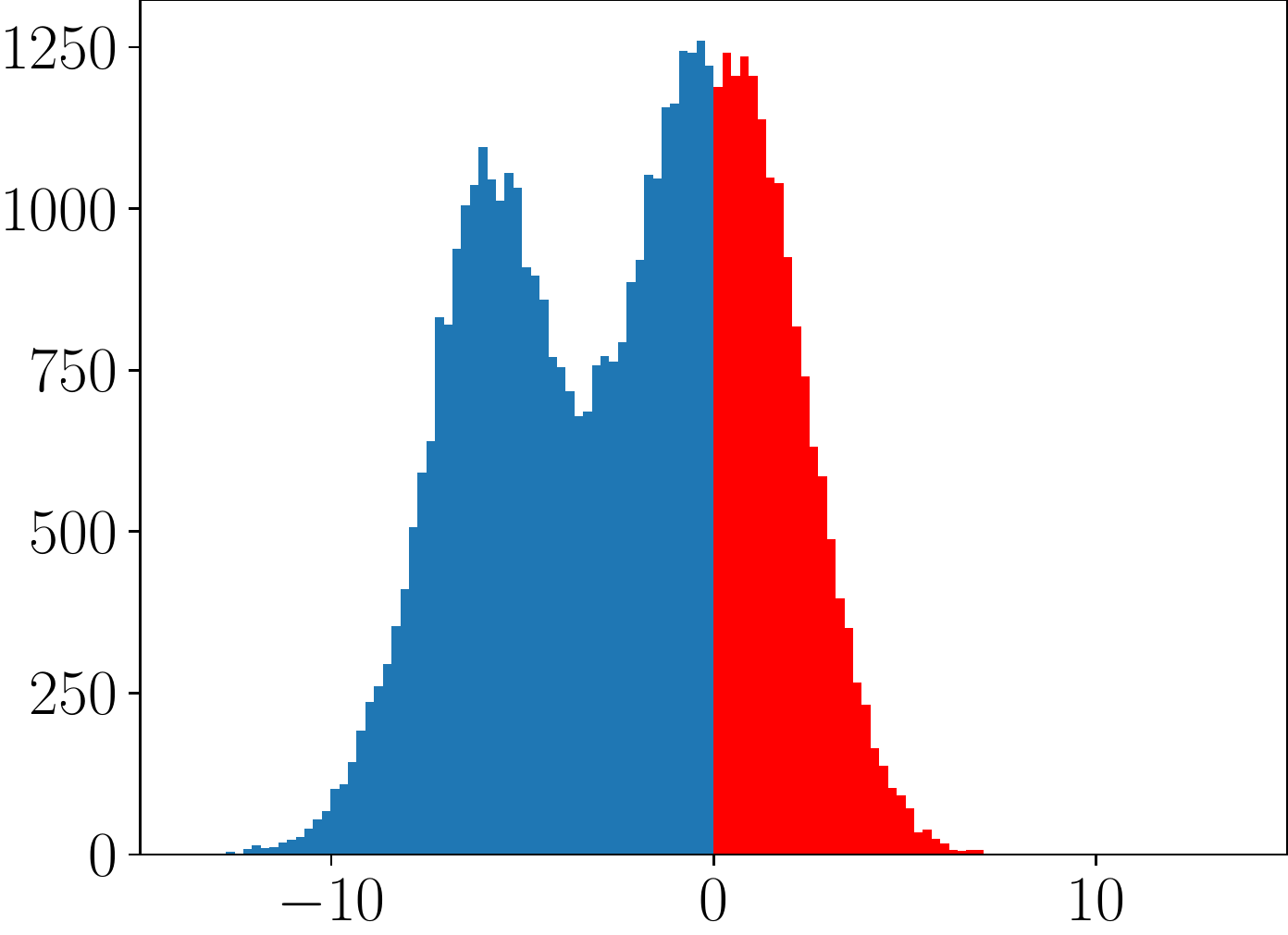}
      \caption{SOVR}\label{C100-LMHist-SOVR}
    \end{subfigure}
    \caption{Histogram of logit margin losses for training samples of CIFAR100 at the last epoch. 
      We plot those on adversarial data $\bm{x}^\prime$ for the other methods. Blue bins are the data points that models correctly classify.}
      \label{C100-LMHist}
    \end{figure}

\begin{figure}[tbp]
      \begin{subfigure}[t]{0.23\linewidth}
        \includegraphics[width=\linewidth]{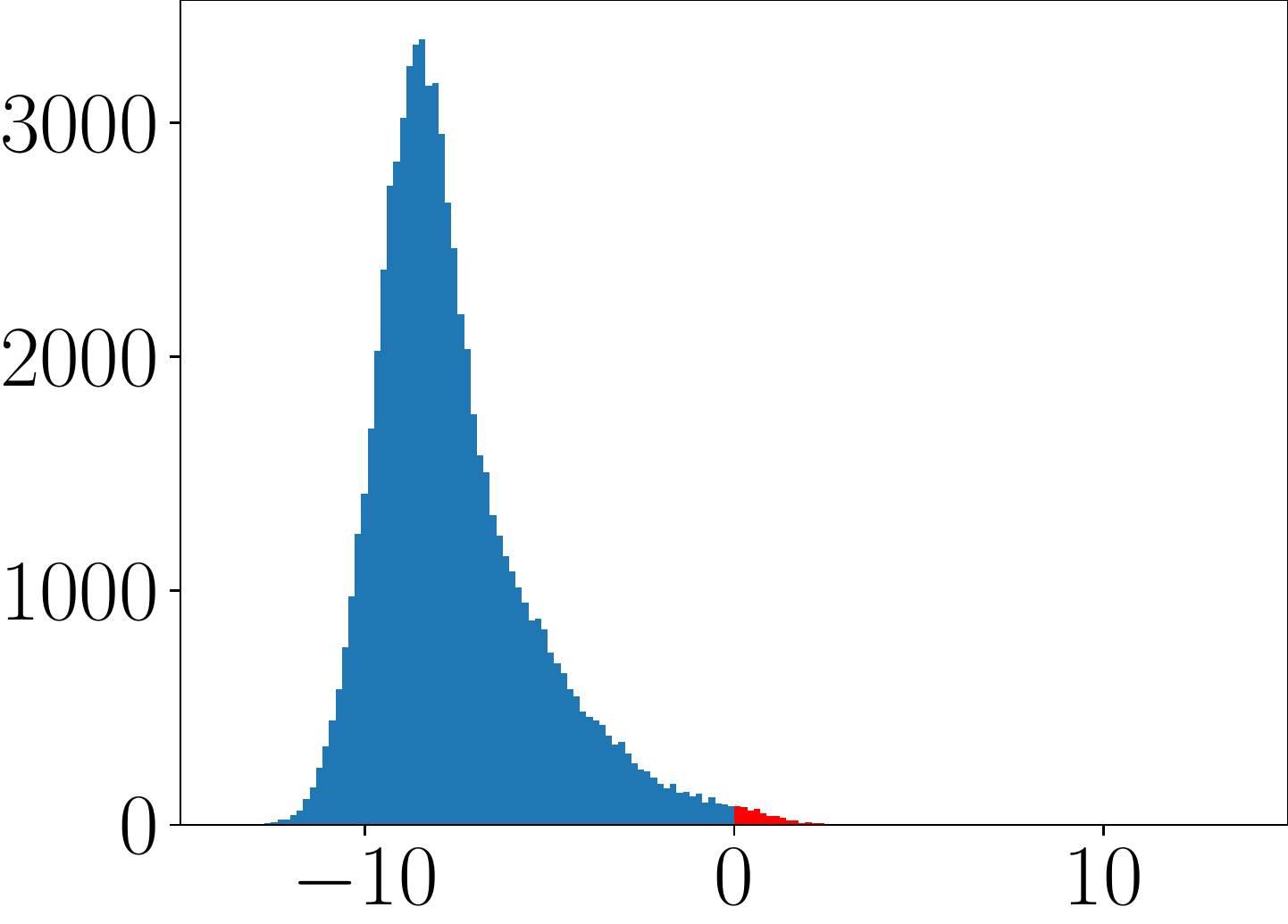}
        \caption{AT}\label{SVHN-LMHist-AT}
    \end{subfigure}
    \begin{subfigure}[t]{0.23\linewidth}
      \includegraphics[width=\linewidth]{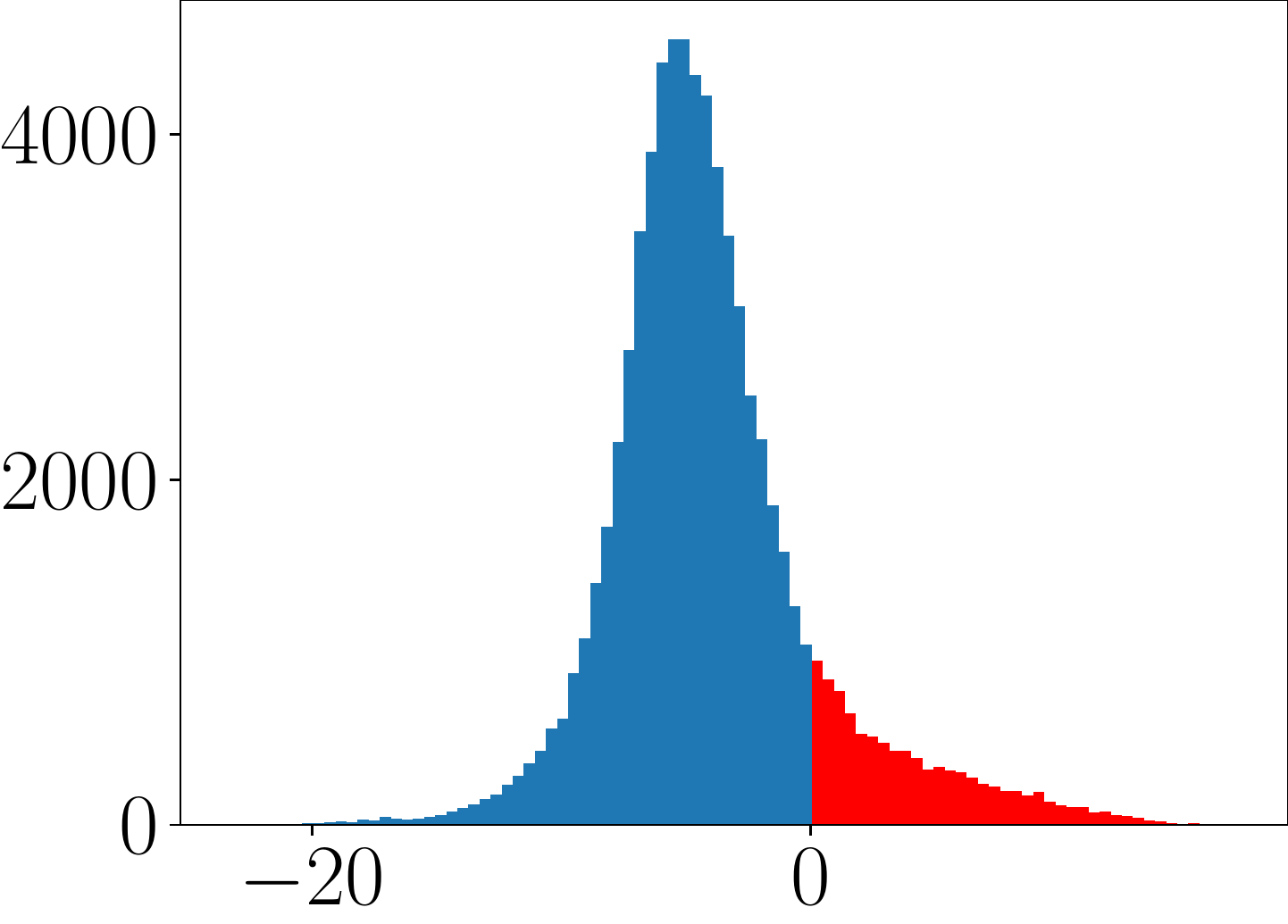}
      \caption{MMA}\label{SVHN-LMHist-MMA}
    \end{subfigure}
    \begin{subfigure}[t]{0.23\linewidth}
      \includegraphics[width=\linewidth]{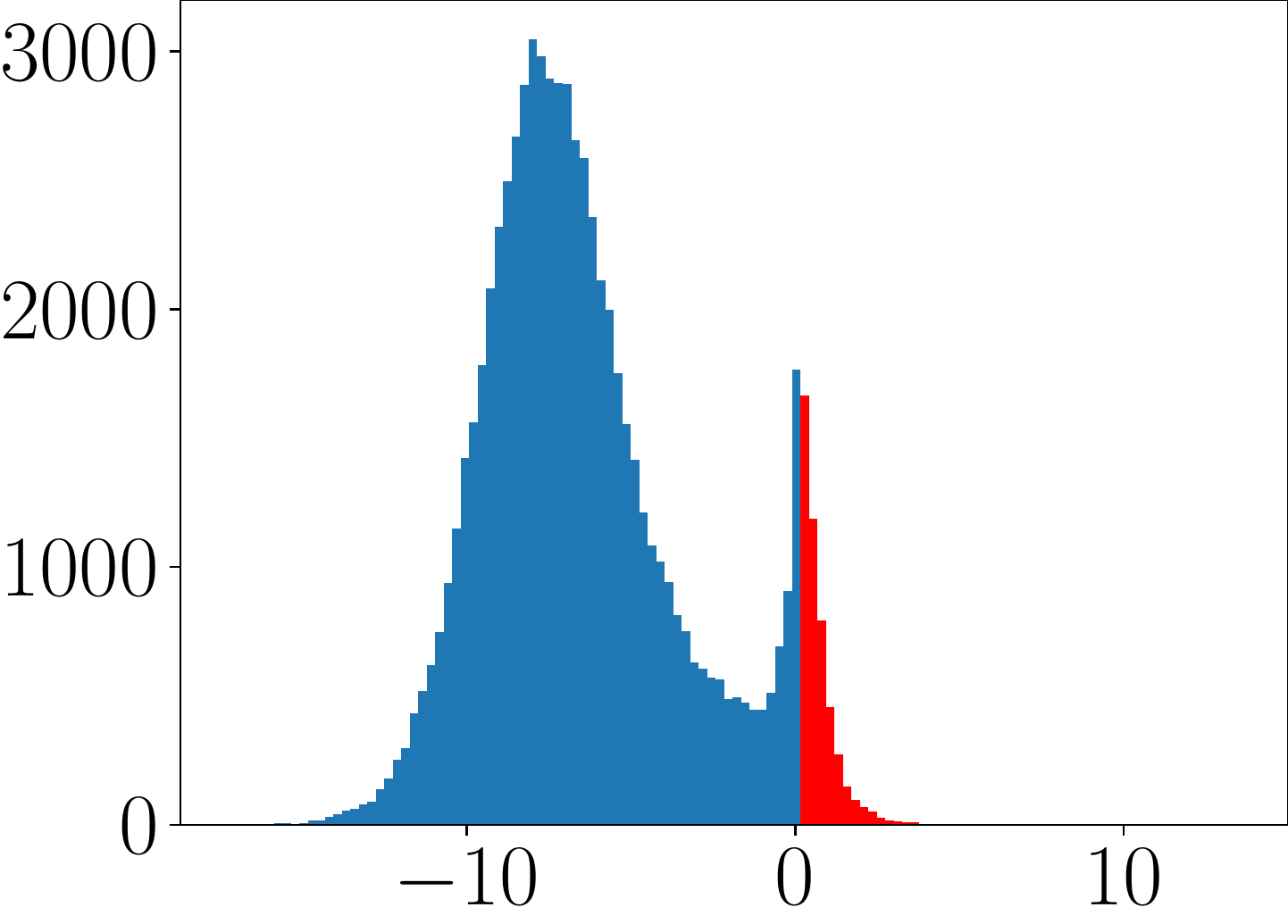}
      \caption{MART}\label{SVHN-LMHist-MART}
    \end{subfigure}
    \begin{subfigure}[t]{0.23\linewidth}
      \includegraphics[width=\linewidth]{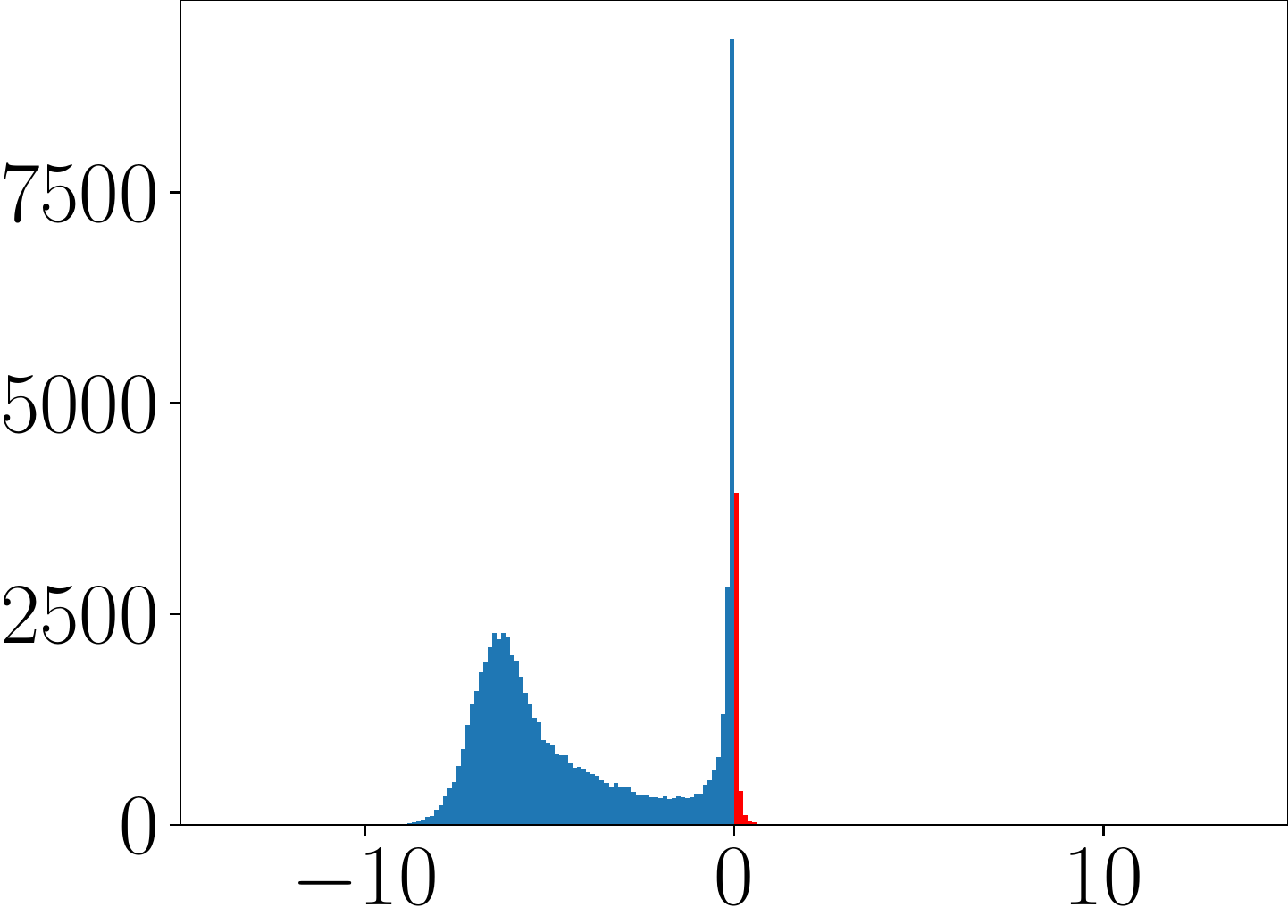}
      \caption{GAIRAT}\label{SVHN-LMHist-GAIRAT}
    \end{subfigure}\\
    \centering
    \begin{subfigure}[t]{0.23\linewidth}
      \includegraphics[width=\linewidth]{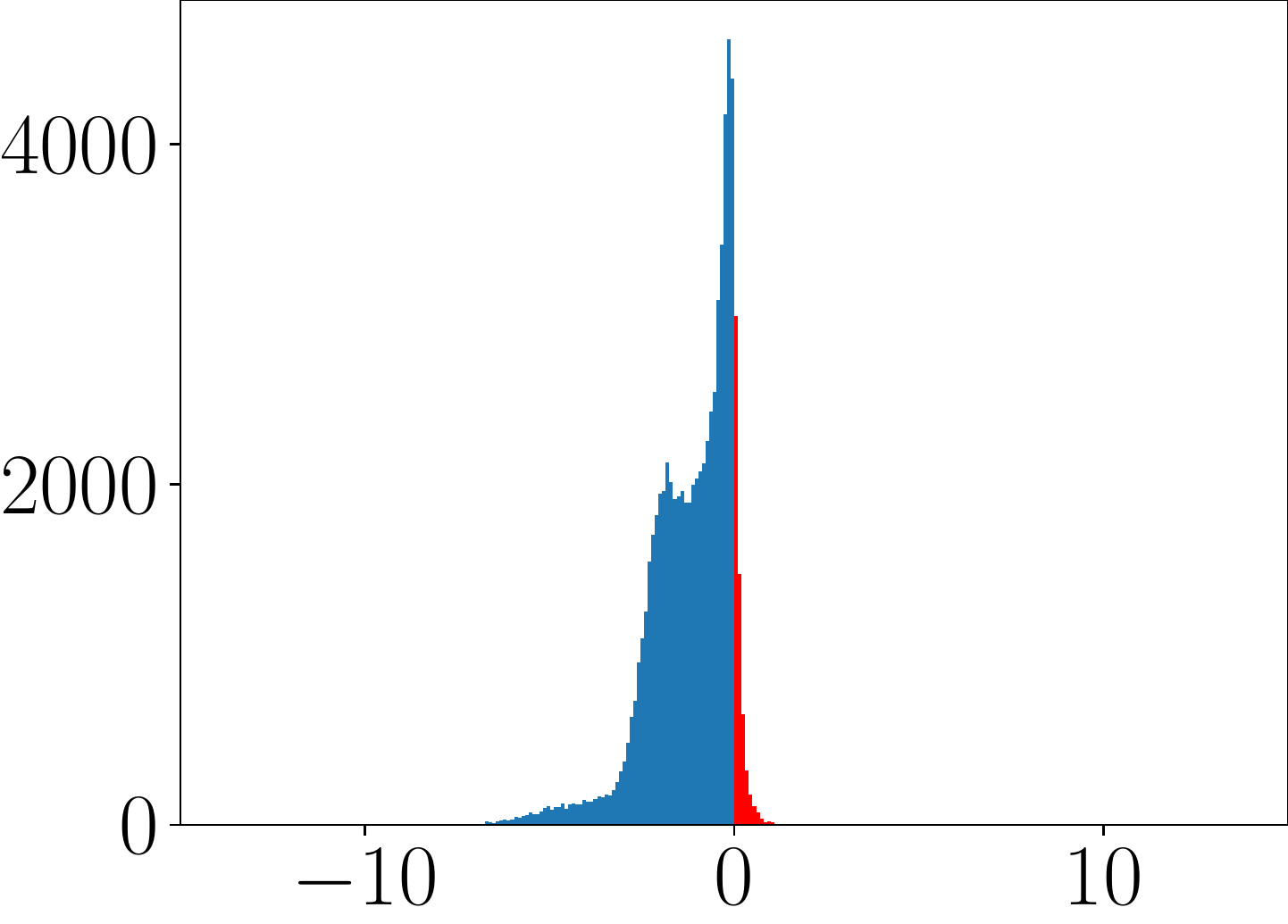}
      \caption{MAIL}\label{SVHN-LMHist-MAIL}
    \end{subfigure}
    \begin{subfigure}[t]{0.23\linewidth}
      \includegraphics[width=\linewidth]{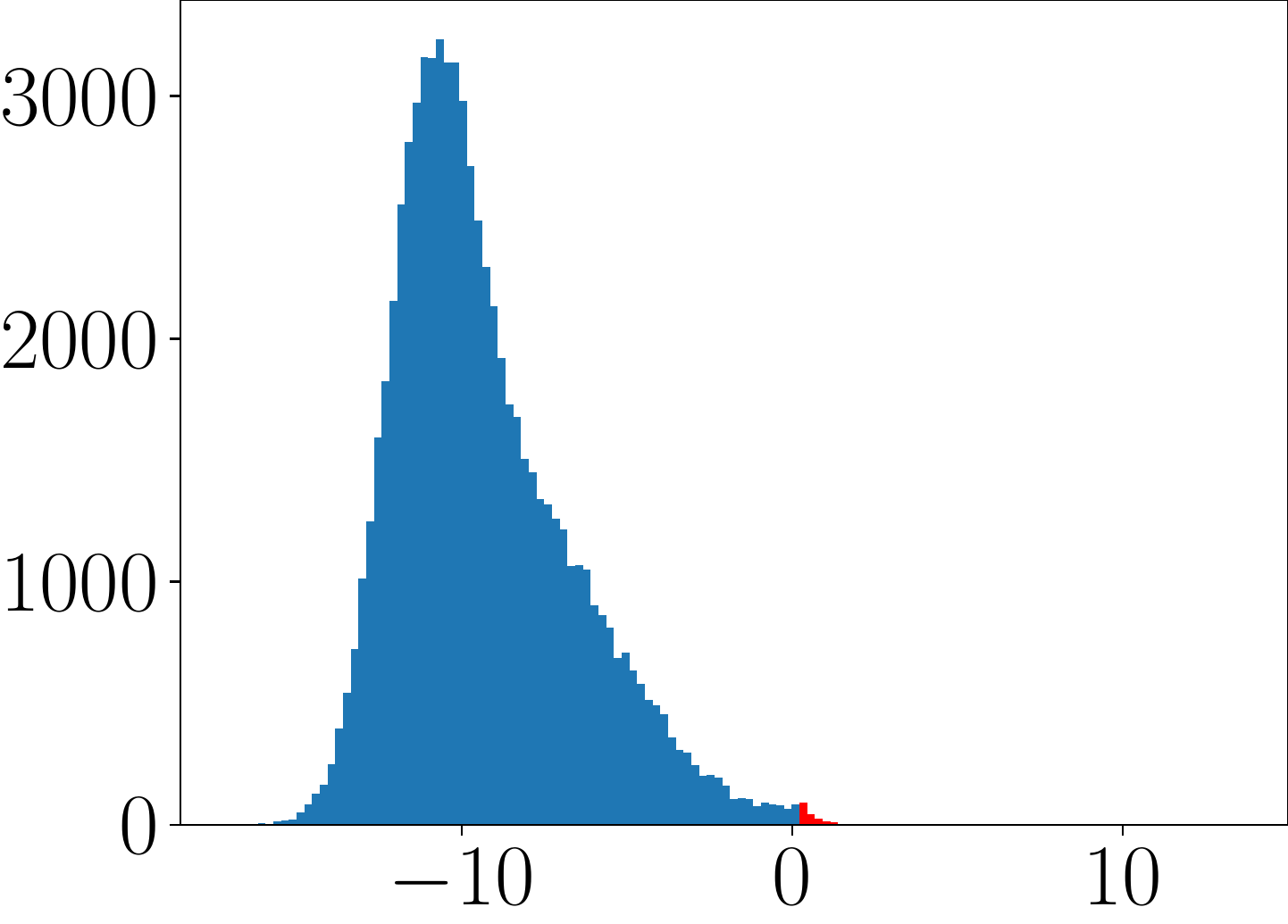}
      \caption{EWAT}\label{SVHN-LMHist-EWL}
    \end{subfigure}
    \begin{subfigure}[t]{0.23\linewidth}
      \includegraphics[width=\linewidth]{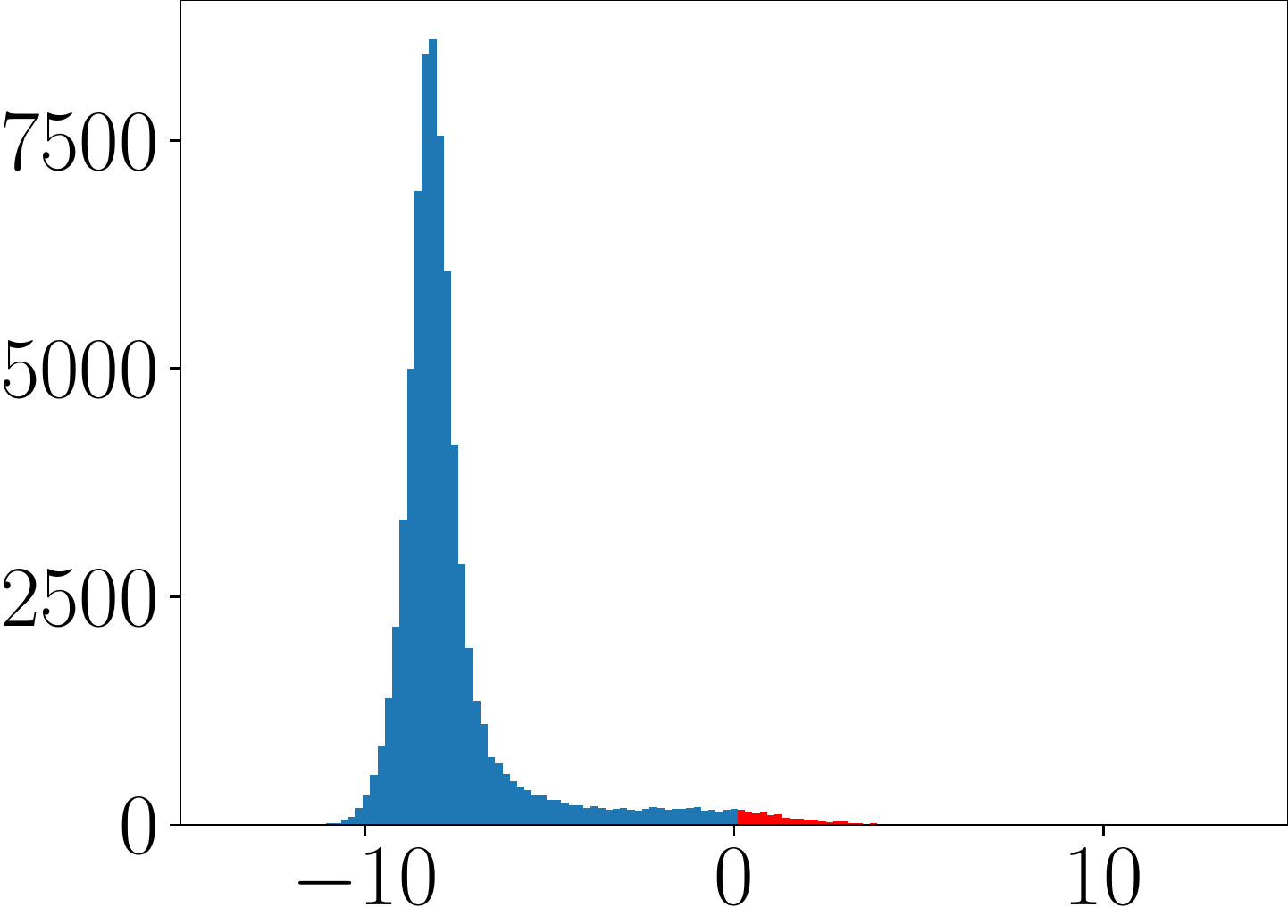}
      \caption{SOVR}\label{SVHN-LMHist-SOVR}
    \end{subfigure}
    \caption{Histogram of logit margin losses for training samples of SVHN at the last epoch. 
      We plot those on adversarial data $\bm{x}^\prime$ for the other methods. Blue bins are the data points that models correctly classify.}
      \label{SVHN-LMHist}
    \end{figure}

\begin{figure}[tbp]
      \begin{subfigure}[t]{0.23\linewidth}
        \includegraphics[width=\linewidth]{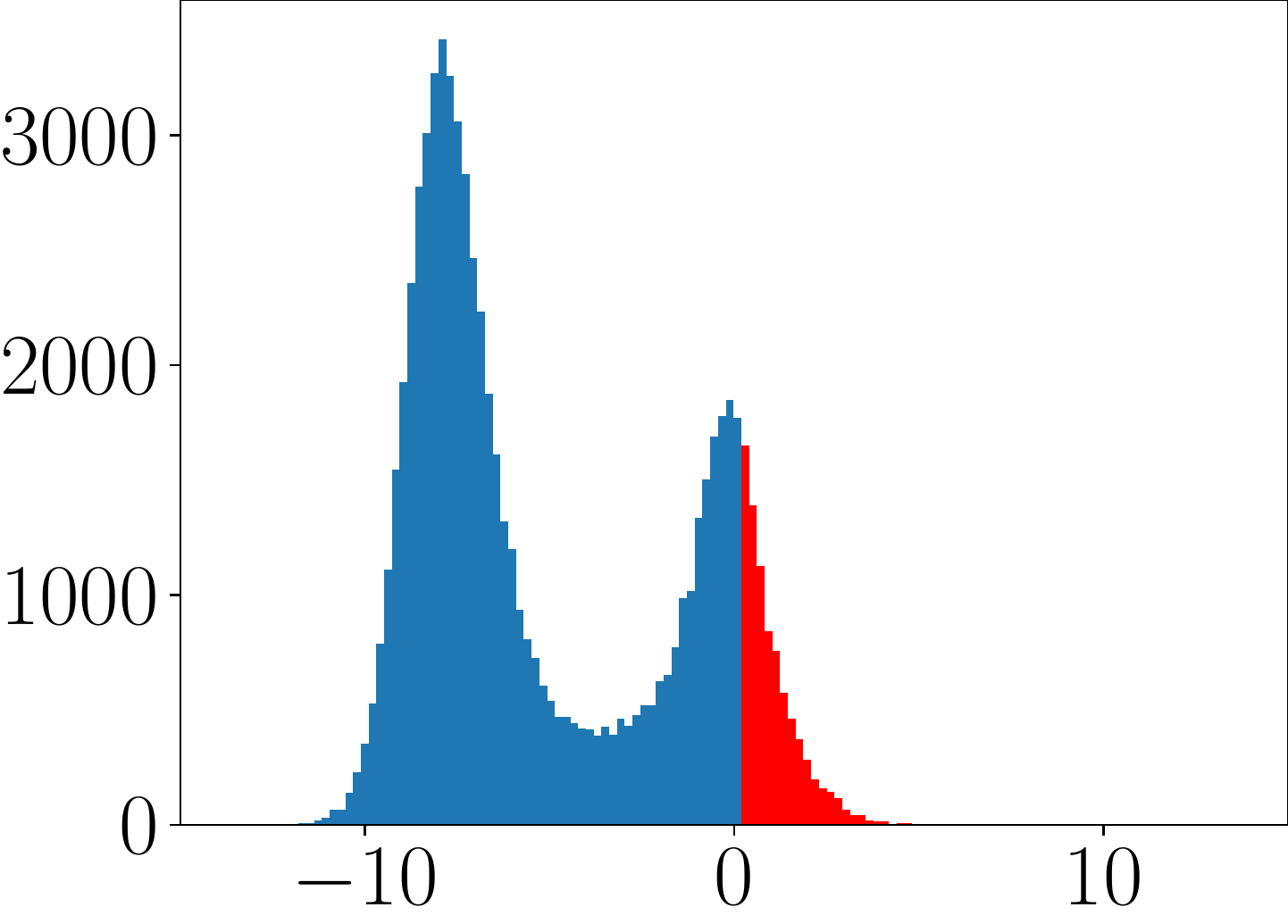}
        \caption{AT}\label{SVHN-LMHist-ATAt100}
    \end{subfigure}
    \begin{subfigure}[t]{0.23\linewidth}
      \includegraphics[width=\linewidth]{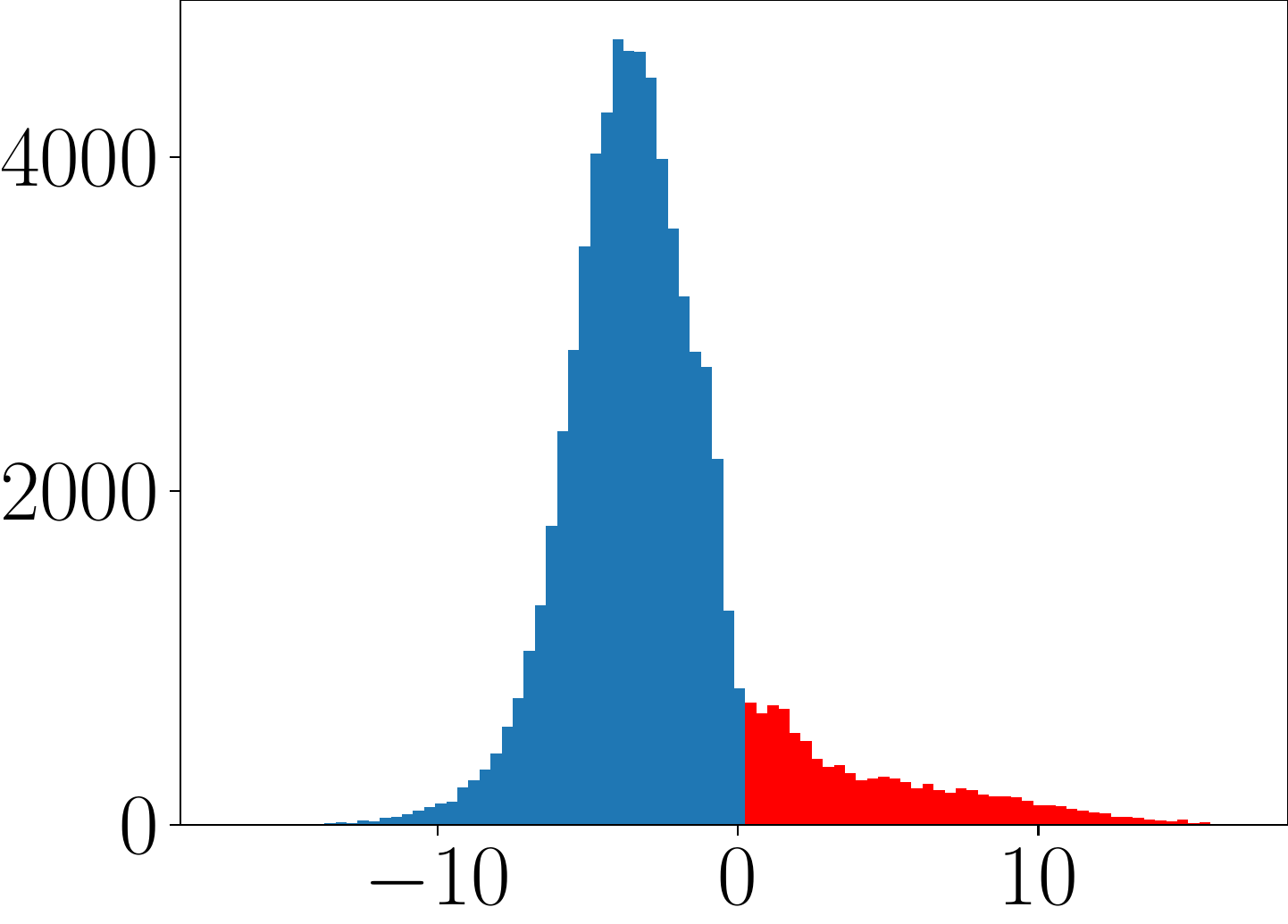}
      \caption{MMA}\label{SVHN-LMHist-MMAAt100}
    \end{subfigure}
    \begin{subfigure}[t]{0.23\linewidth}
      \includegraphics[width=\linewidth]{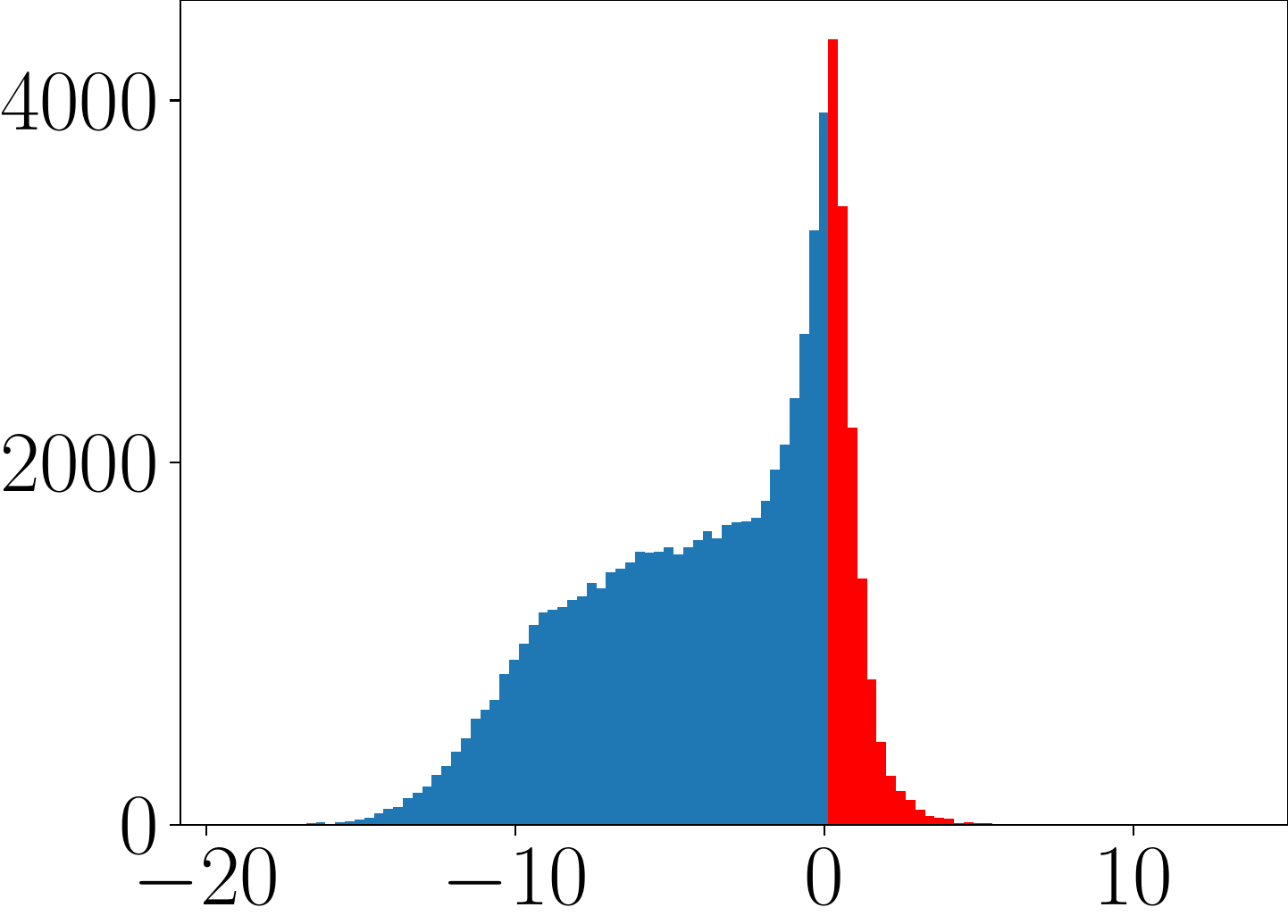}
      \caption{MART}\label{SVHN-LMHist-MARTAt100}
    \end{subfigure}
    \begin{subfigure}[t]{0.23\linewidth}
      \includegraphics[width=\linewidth]{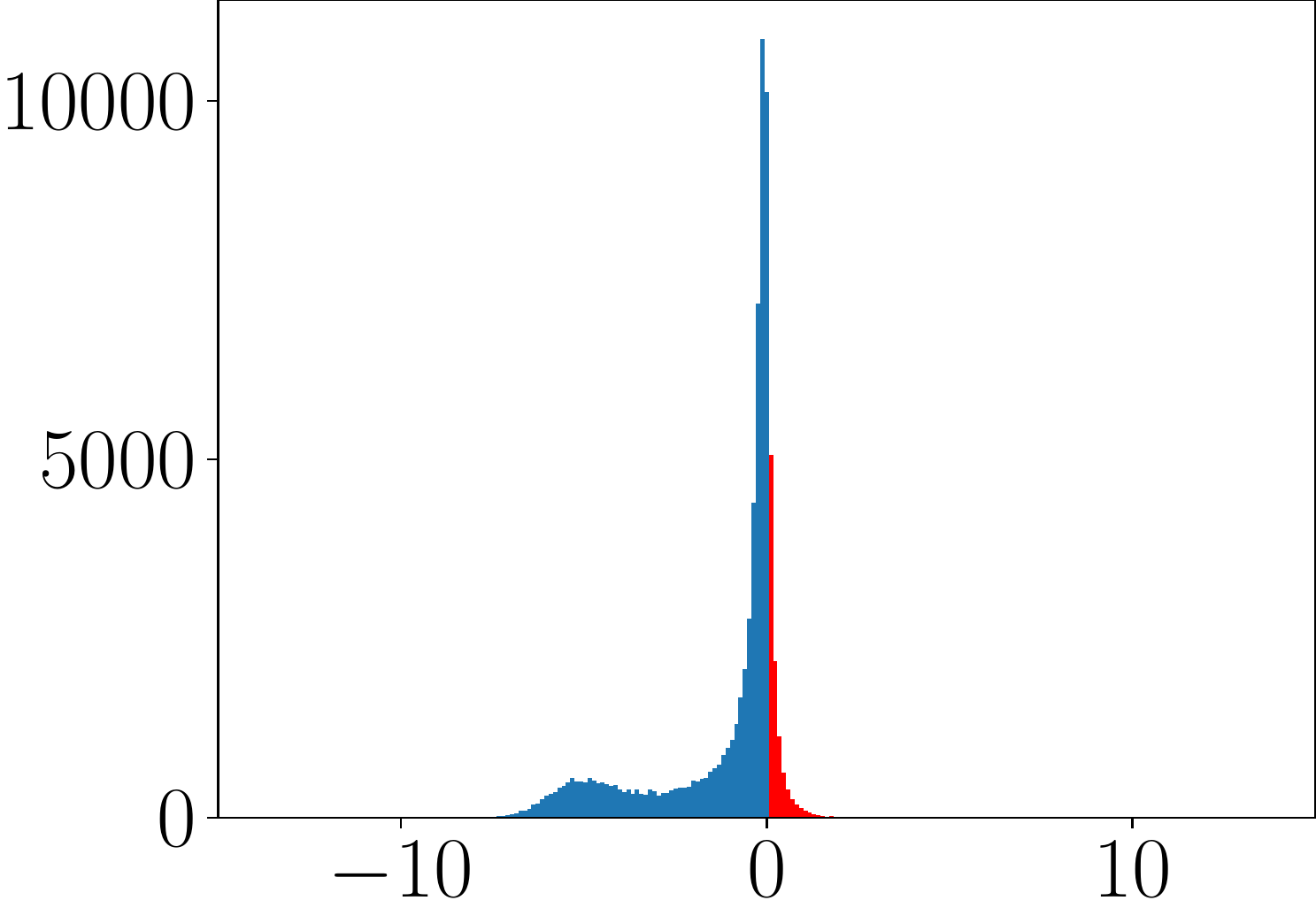}
      \caption{GAIRAT}\label{SVHN-LMHist-GAIRATAt100}
    \end{subfigure}\\
    \centering
    \begin{subfigure}[t]{0.23\linewidth}
      \includegraphics[width=\linewidth]{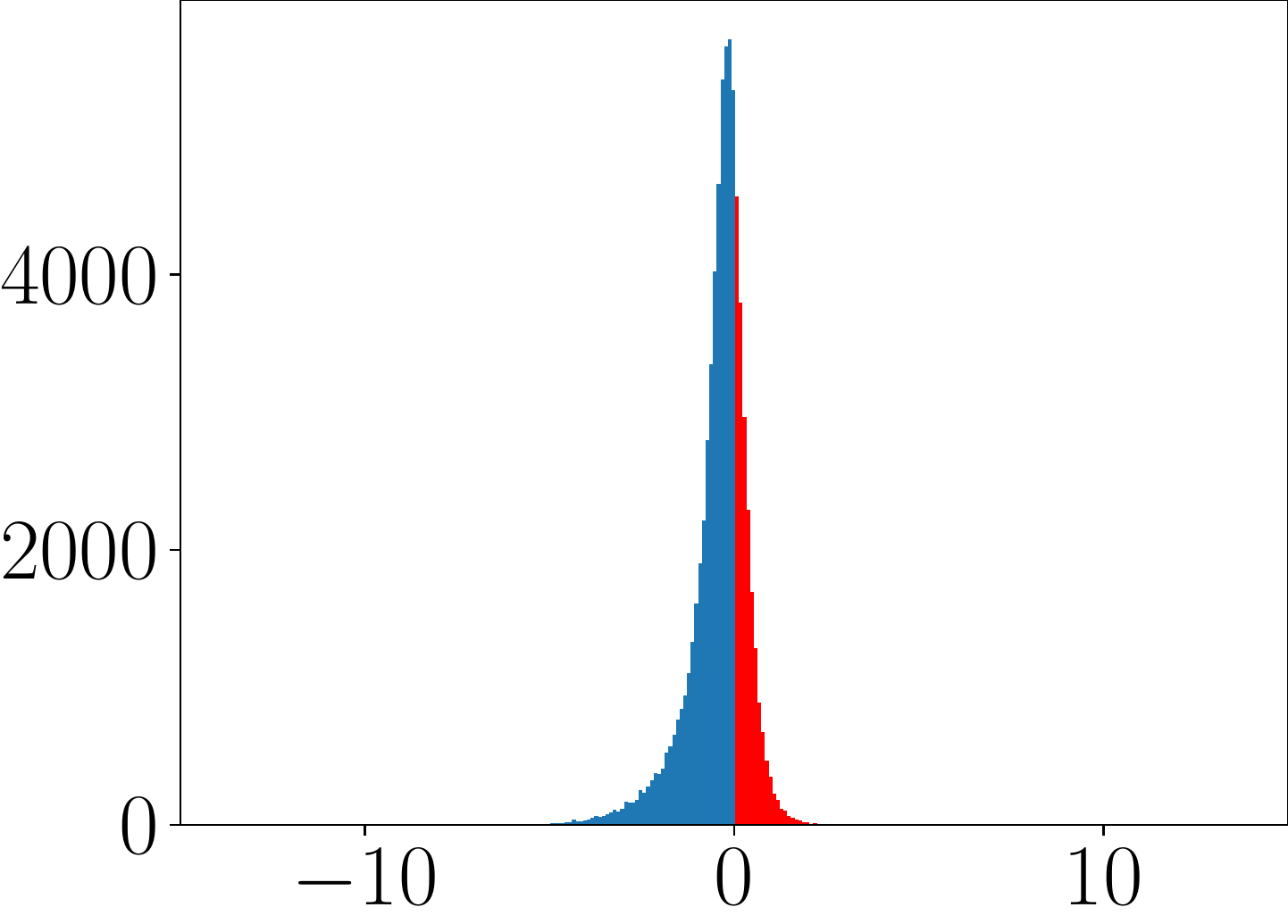}
      \caption{MAIL}\label{SVHN-LMHist-MAILAt100}
    \end{subfigure}
    \begin{subfigure}[t]{0.23\linewidth}
      \includegraphics[width=\linewidth]{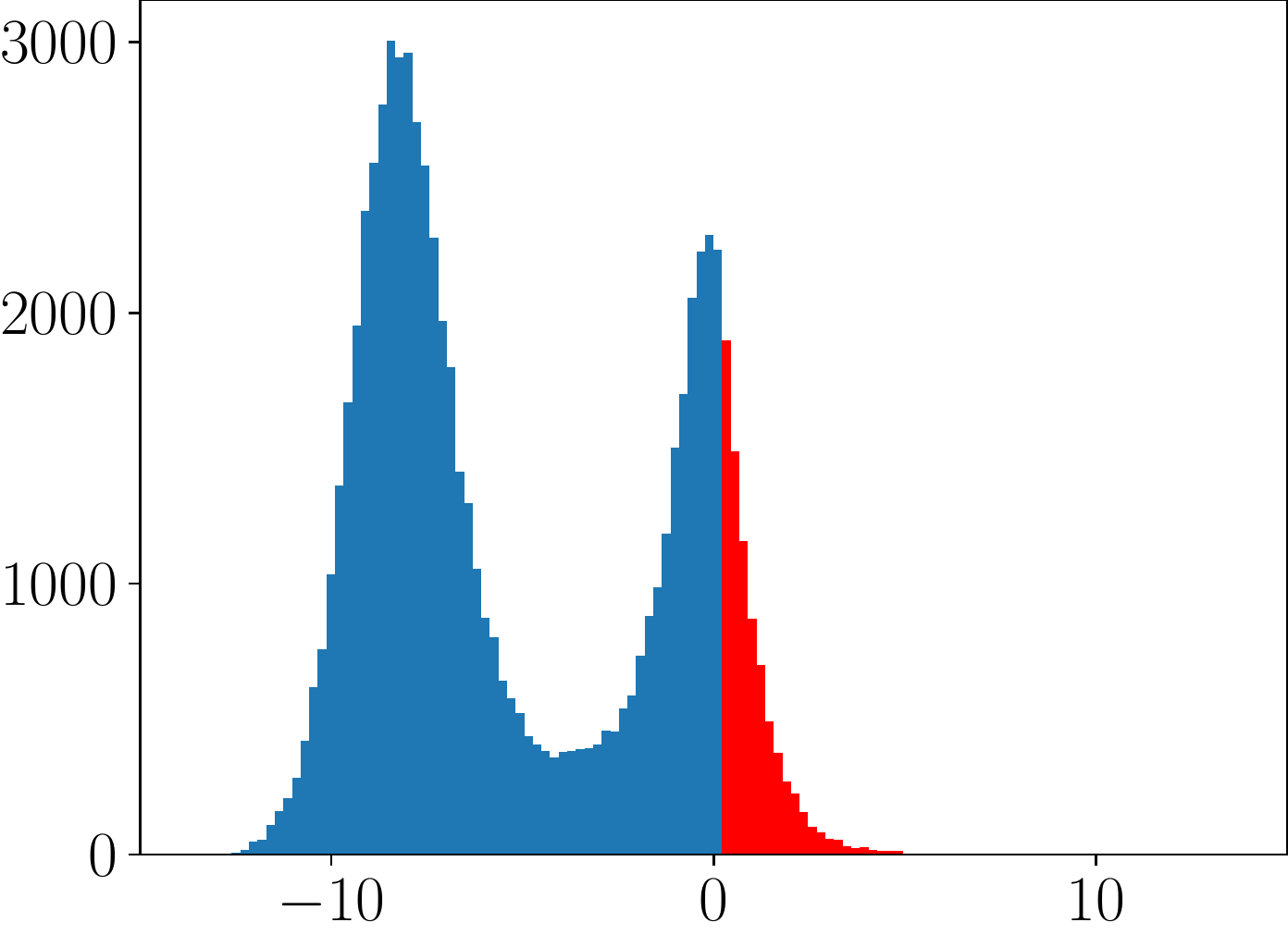}
      \caption{EWAT}\label{SVHN-LMHist-EWLAt100}
    \end{subfigure}
    \begin{subfigure}[t]{0.23\linewidth}
      \includegraphics[width=\linewidth]{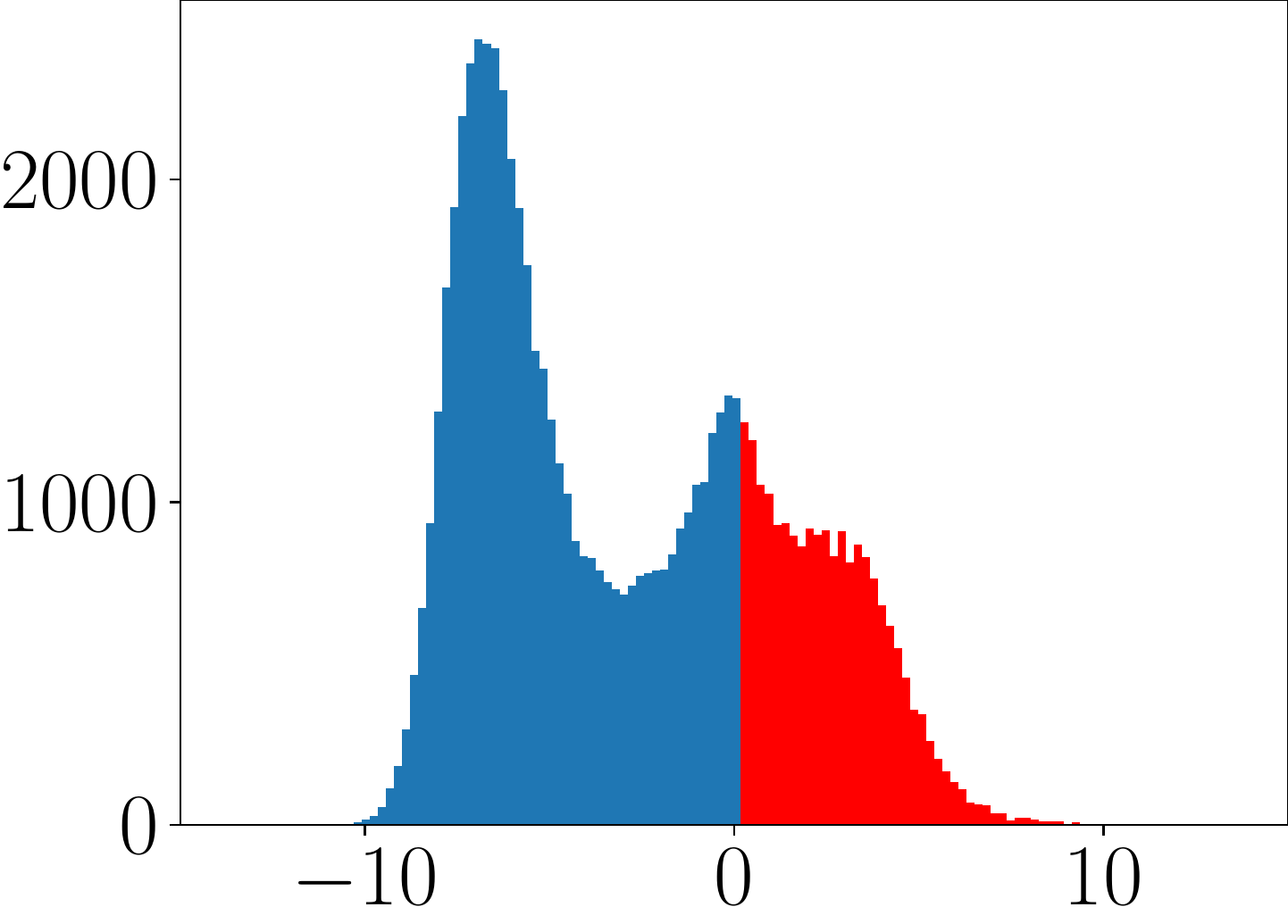}
      \caption{SOVR}\label{SVHN-LMHist-SOVRAt100}
    \end{subfigure}
    \caption{Histogram of logit margin losses for training samples of SVHN at the 100-th epoch. 
      We plot those on adversarial data $\bm{x}^\prime$ for the other methods. Blue bins are the data points that models correctly classify.}
      \label{SVHN-LMHistAt100}
    \end{figure}

\begin{figure}[tbp]
     \centering
     \begin{subfigure}[t]{0.23\linewidth}
       \includegraphics[width=\linewidth]{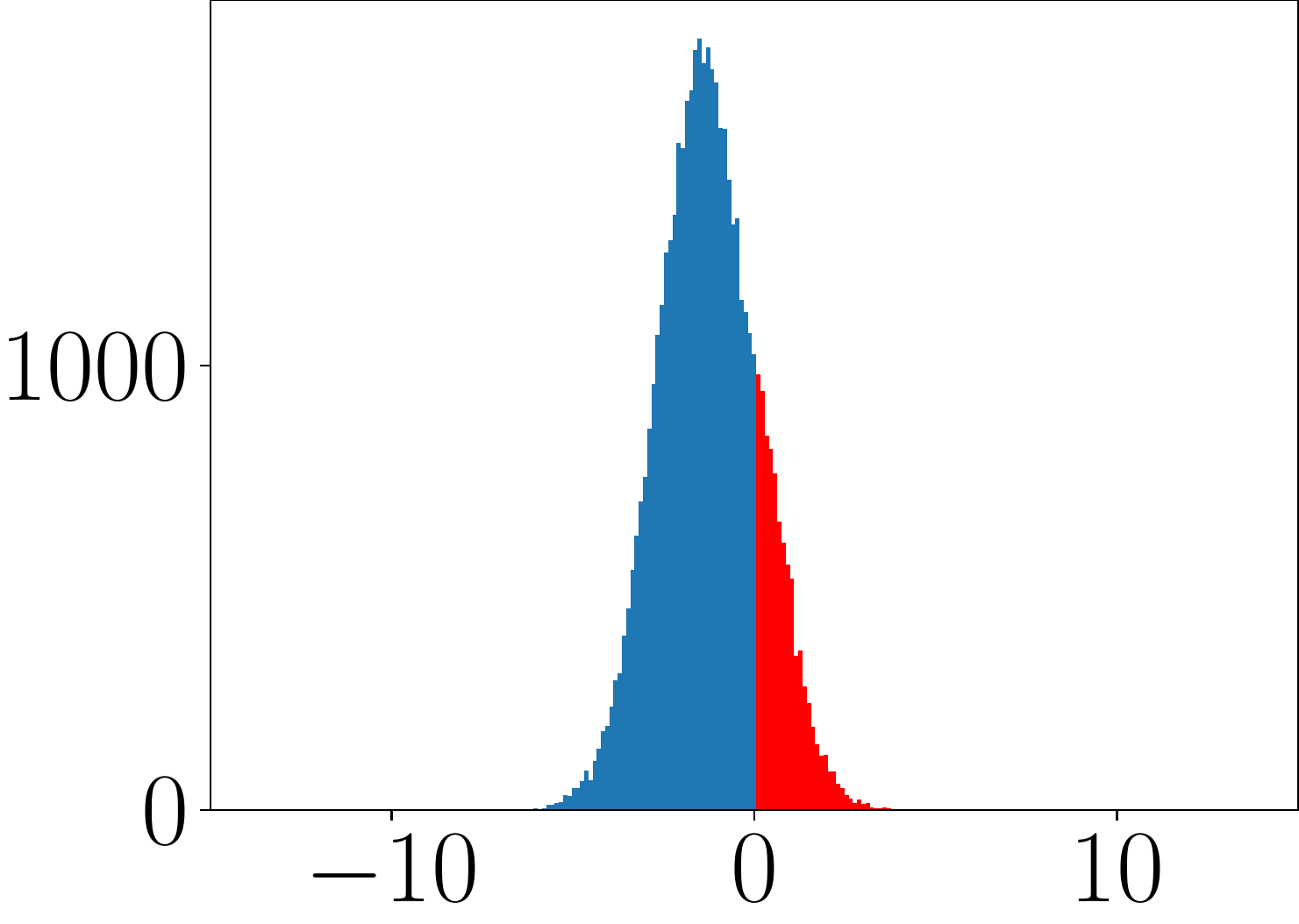}
       \caption{Best epoch}
     \end{subfigure}
     \begin{subfigure}[t]{0.23\linewidth}
     \includegraphics[width=\linewidth]{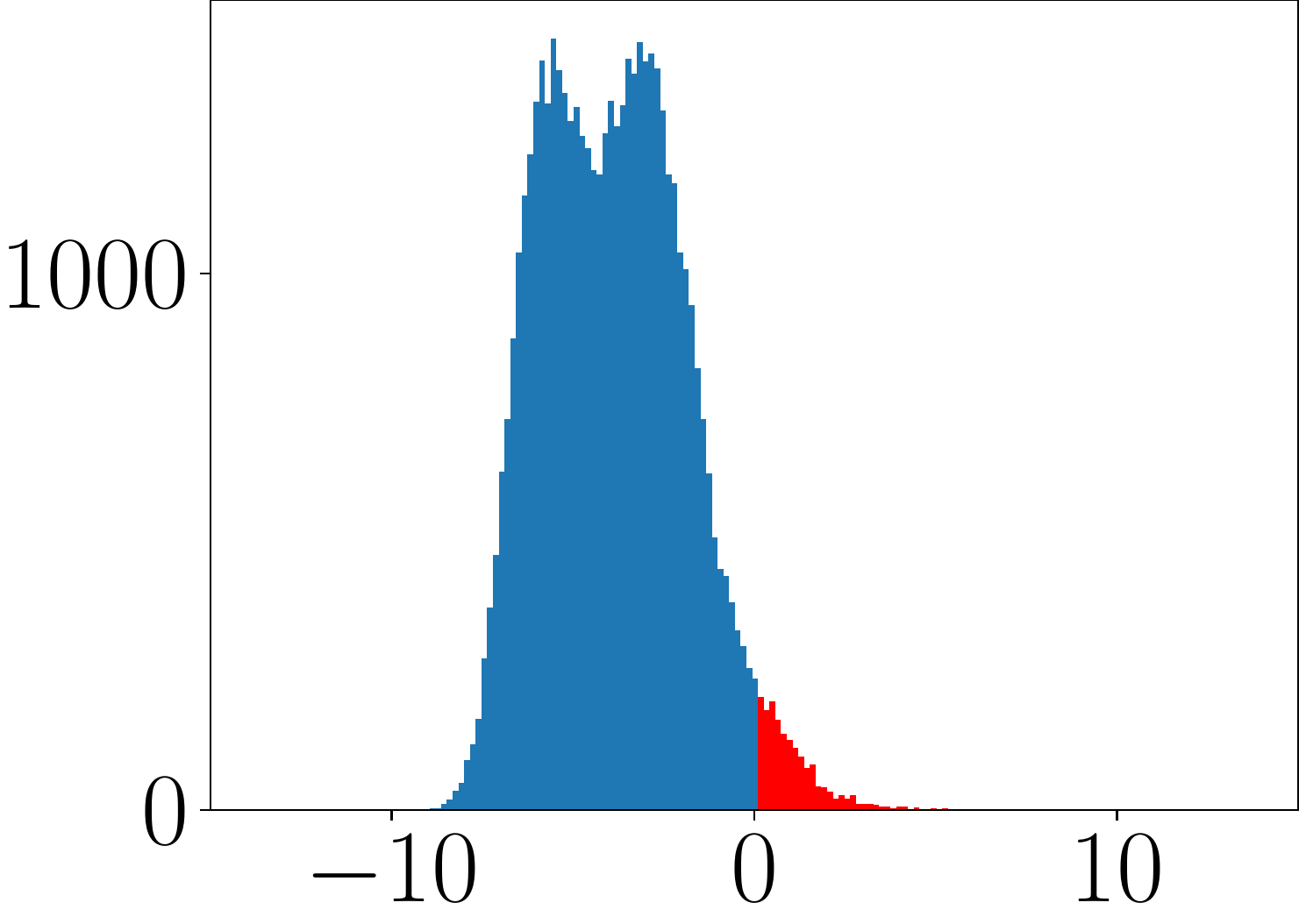}
     \caption{Last epoch}
   \end{subfigure}
   \caption{Histogram of logit margin losses on CIFAR10 with WRN for TRADES.}
     \label{TRADES-LMHist}
   \end{figure}

    We show the additional histograms of logit margin losses in this section.
    First, \rfig{LMHist-EWAT} plots the result of EWAT on training samples of CIFAR10 at the last epoch.
    Compared with SOVR, EWAT does not increase logit margins for important (difficult) samples (right peak).
    Figures~\ref{C10WRN-LMHist}-\ref{SVHN-LMHistAt100} plot the histograms when using WideResNet and other datasets.
    SOVR tends to increase the left peak under all conditions, and thus, it decreases logit margin losses $\ell_{\mathrm{LM}}$, and thus, it increases the logit margins $|\ell_{\mathrm{LM}}|$.
    Figure~\ref{SVHN-LMHist} shows that AT does not have two peaks on SVHN.
    To investigate histograms on SVHN in detail, we additionally evaluate logit margin losses at the 100-th epoch in \rfig{SVHN-LMHistAt100}.
    This figure shows that the histogram on SVHN has two peaks at the 100-th epoch, but they became one peak at the 200-th epoch (\rfig{SVHN-LMHist}).
    This might cause the optimal $(M, \lambda)$ for SOVR to be smaller than that for other datasets.
    Figure~\ref{TRADES-LMHist} plots the histograms of TRADES and shows that
    TRADES has two peaks but they are close to each other.
    This might be because the objective functions for adversarial examples and parameters are different.
    Table~\ref{LMTabProp} lists the average of logit margin losses.
    Since the distributions of logit margin losses are long-tailed as shown in histograms,
    the difference in average values of logit margin losses among methods is small.
    Even so, SOVR tends to have the lowest logit margin losses under almost all settings.
    
    
    \begin{table}[tbp]
      \centering
      \caption{Average logit margin losses for training dataset $\bm{x}^\prime$ at the last epoch. }
      \label{LMTabProp}
      \resizebox{\linewidth}{!}{
      \begin{tabular}{ccccccccc}\toprule
      Dataset &AT&MART&MMA&GAIRAT&MAIL&EWAT&SOVR \\ \midrule
      CIFAR10 (RN18)&-3.66$\pm$0.02&-3.46$\pm$0.02&0.558$\pm$0.1  &-0.258$\pm$ 0.02&-0.0293$\pm$ 0.02&-2.251$\pm$0.007&-4.34$\pm$ 0.02\\
      CIFAR10 (WRN)&-6.96$\pm$0.01&-5.65$\pm$0.7&-2.73$\pm$0.6  &-0.717$\pm$ 0.03&-0.203$\pm$ 0.01&-5.69$\pm$0.04&-8.56$\pm$ 0.04\\
      CIFAR100&-2.41$\pm$0.02&-1.76$\pm$0.01&2.07$\pm$0.06   &-0.68$\pm$0.02 &0.684$\pm$0.1&-1.36$\pm$0.02&-2.44$\pm$0.1\\
      SVHN&-7.55$\pm$0.01&-8.08$\pm$2  &-4.31$\pm$0.04&-3.91$\pm$0.01&-1.17$\pm$0.03 &-9.32$\pm$0.04&-7.59$\pm$0.1\\
      \bottomrule
      \end{tabular}
      }
    \end{table}
    \subsection{Evaluation of Gradient Norms}
    \begin{figure}[tbp]
     \centering
     \includegraphics[width=\linewidth]{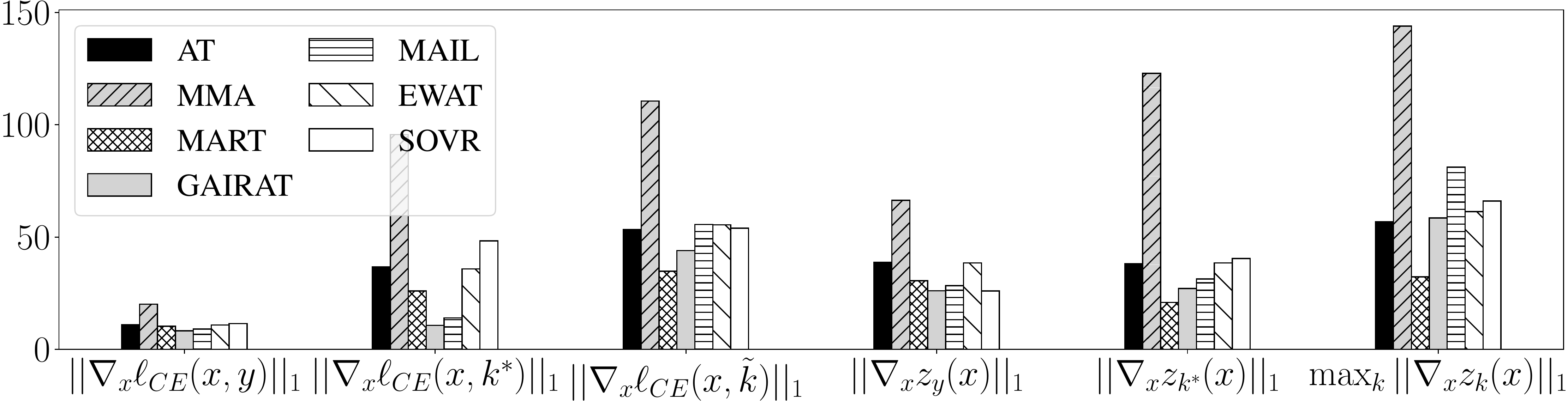}
     \caption{Average of gradient norms with respect to data points.
     $\tilde{k}$ is randomly selected labels, and $k^*\!=\!\argmax_{k\neq y}z_{k}(\bm{x})$.}
     \label{AllLip}
   \end{figure}
   Even though logit margins of importance-aware methods are very small, they are robust against PGD and some attacks (\rfig{ATDis}).
    To reveal the cause of this robustness, we additionally evaluate th  gradient norms for loss and logit functions (\rfig{AllLip}).
    In this figure, the gradient norms of cross-entropy $||\nabla_x\ell_{\mathrm{CE}}(\bm{x},y)||_{1}$ are relatively small values in all methods.
    This indicates that adversarial training essentially attempts to suppress the gradient norms
    for the cross-entropy.   
    MMA has the largest gradient norms, and this is the reason MMA is not robust against
    Auto-Attack except for SQUARE (\rfig{ATDis}), which does not use gradient.
    GAIRAT and MAIL have the smallest and second smallest $||\nabla_x\ell_{\mathrm{CE}}(\bm{x},y)||_1$,
    and this is the reason they are robust against PGD despite the small logit margins (\rfig{LMHist}).
    On the other hand, $\max_k ||\nabla_x z_{k}(\bm{x})||_1$ of importance-aware methods
    is larger than $||\nabla_x\ell_{\mathrm{CE}}(\bm{x},y)||_1$ of them and that of AT.
    As a result, they can have larger rate of potentially misclassified samples (\rfig{Poten}).
    Gradient norms of
    cross-entropy for the label that has the largest logit except for the true label $||\nabla_x\ell_{\mathrm{CE}}(\bm{x},k^*)||_1$
    are smaller than those for the randomly selected labels.
    This implies $k^*\neq \hat{k}$, and we need to use the loss that depends on the logits for all classes rather than logit margin loss, which only cares $z_k^*$ and $z_y$.
    Gradient norms of EWAT and SOVR are not significantly different from those of AT.
    Thus, SOVR can reduce the rate of potentially misclassified samples by large logit margins and not large gradient norm.
    \subsection{Trajectories of Logit Margin Losses in Adversarial Training with OVR and Cross-entropy}\label{ATLMTrajSec}
    \begin{figure}[tb]
     \centering
   \begin{subfigure}[t]{0.32\linewidth}
       \centering
       \includegraphics[width=\linewidth]{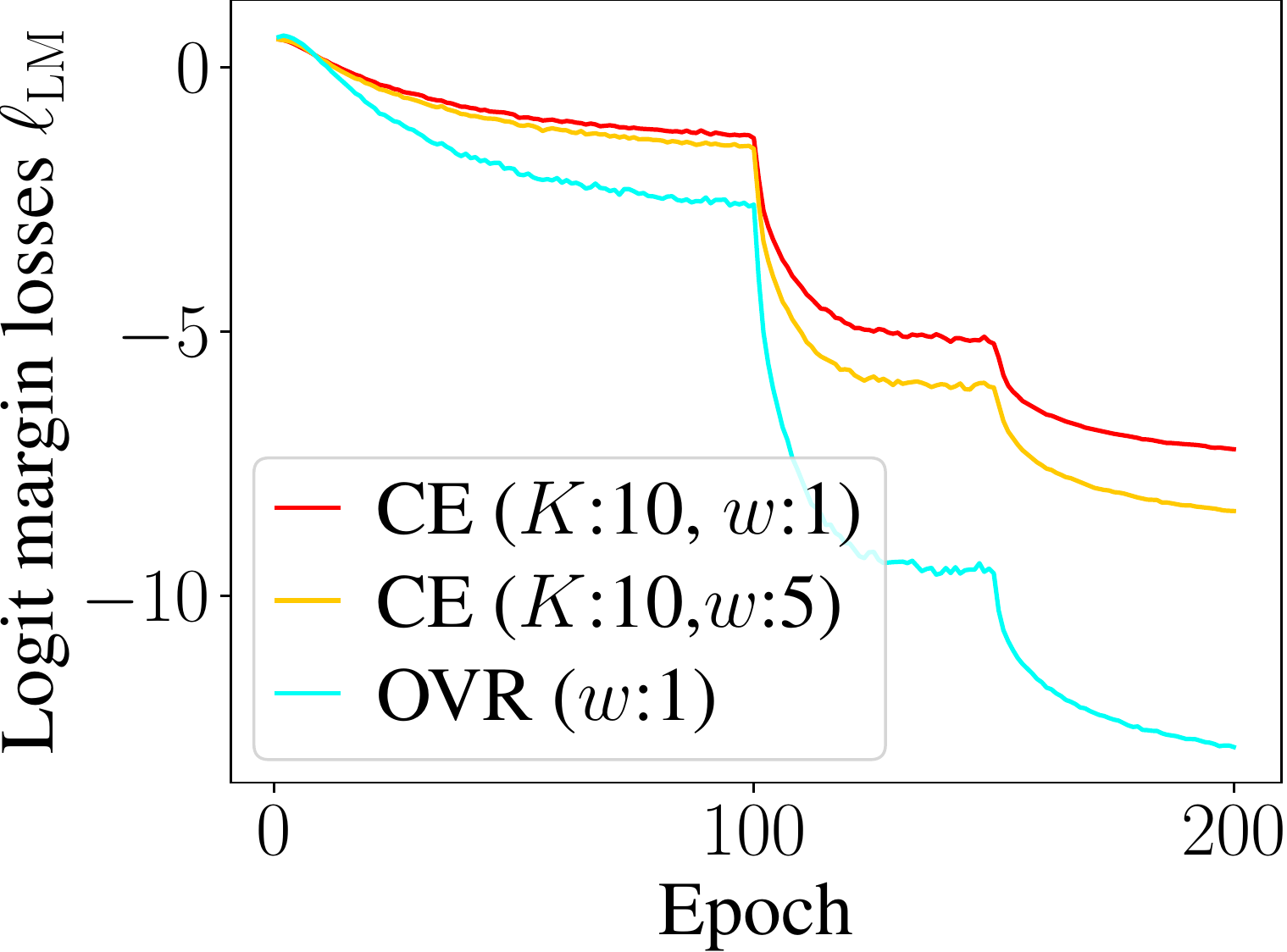}
       \caption{CIFAR 10 (WRN)}
       \label{TrajWRN}
     \end{subfigure}\hfill
       \begin{subfigure}[t]{0.32\linewidth}
       \centering
     \includegraphics[width=\linewidth]{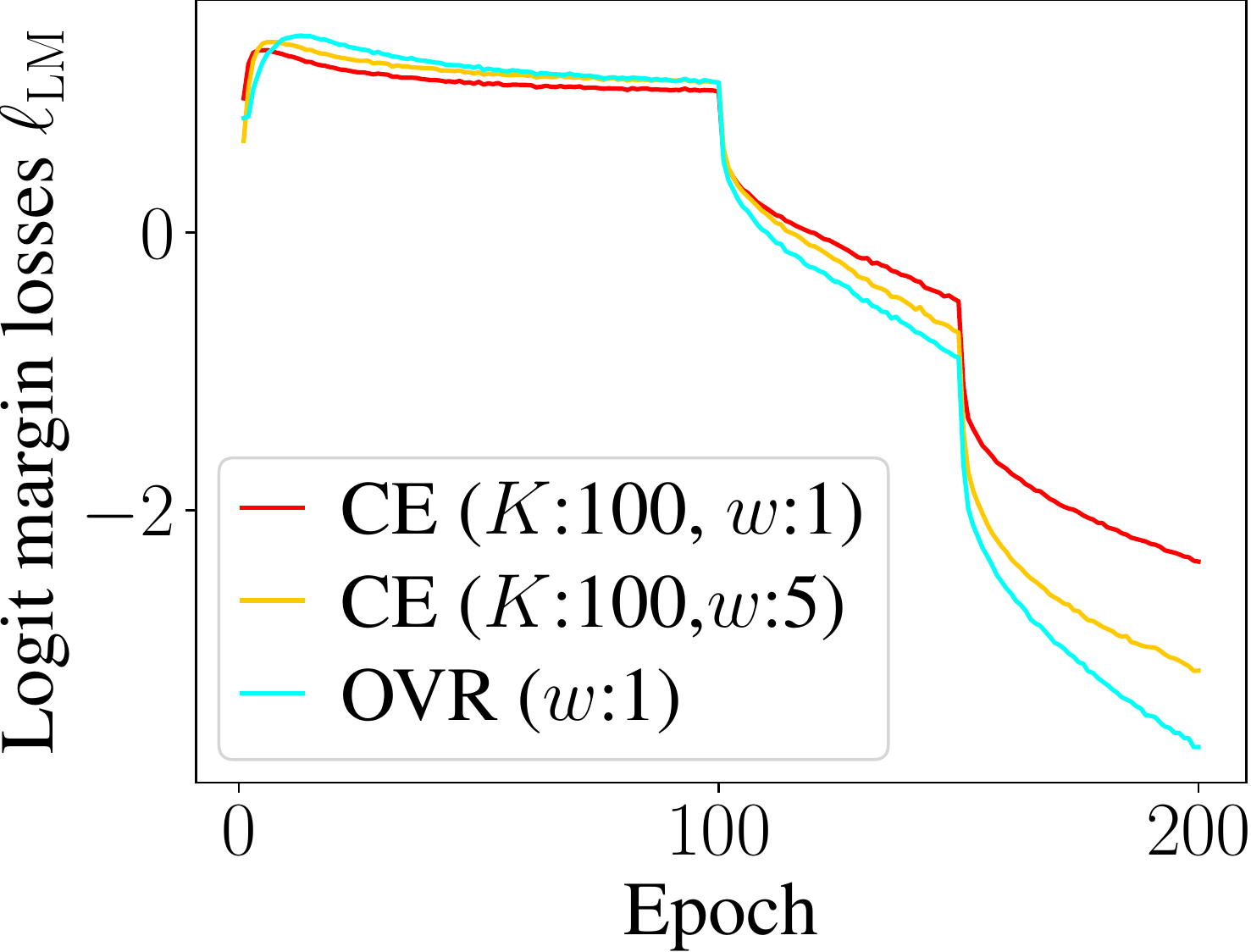}
     \caption{CIFAR 100}
     \label{TrajC100}
   \end{subfigure}\hfill
   \begin{subfigure}[t]{0.32\linewidth}
     \centering
     \includegraphics[width=\linewidth]{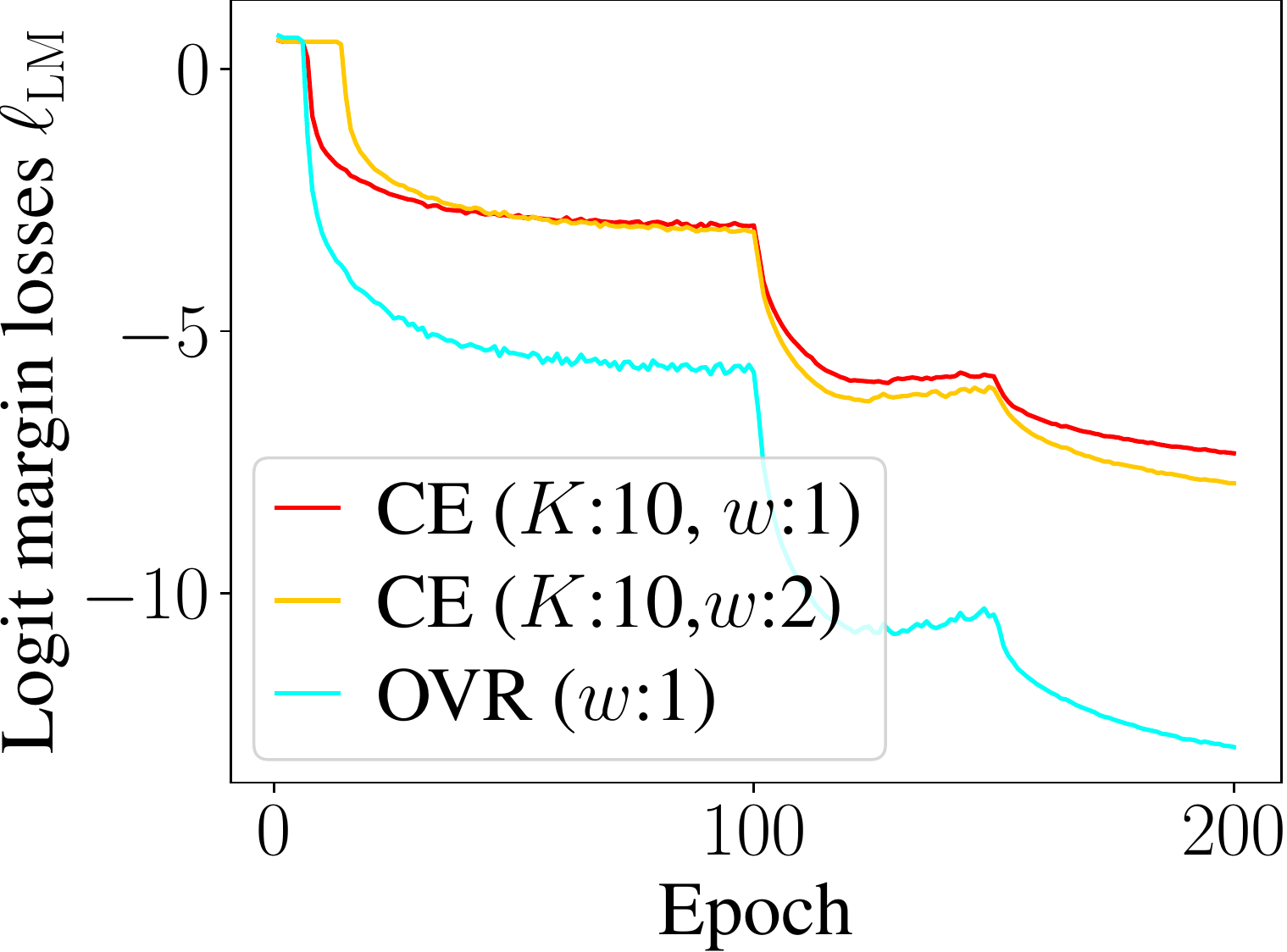}
     \caption{SVHN}
     \label{TrajSVHN}
   \end{subfigure}
   \caption{Trajectories of logit margin losses $\ell_{\mathrm{LM}}$ in adversarial training using cross-entropy and OVR.
   RN18 is used on CIFAR100 and SVHN.}\label{LMTrajATs}
   \end{figure}
   In this section, we evaluate the trajectories of logit margin losses in adversarial training on real data.
   Experimental setup is the same with that of Section~\ref{ExSec} except for the learning rate on CIFAR10, and thus, we minimize the OVR and cross-entropy averaged over the dataset, unlike \req{ToyProb}.
   While the learning rates are set to 0.1 for cross-entropy and 0.05 for SOVR in Section~\ref{ExSec} on CIFAR10,
   learning rate is set to 0.05 on CIFAR10 for both cross-entropy and OVR to fairly compare their logit margin losses in this experiment.
   Since we could not obtain results on SVHN with the weight of $w\!=\!5$ due to unstability, we used $w\!=\!2$ on SVHN.
   \rfig{LMTrajATs} plots the logit margin losses averaged over the dataset against epochs in adversarial training with OVR and cross-entropy on CIFAR10, CIFAR100, and SVHN.
 In \rfig{LMTrajATs}, OVR decreases the logit margin losses more than cross-entropy on all dataset.
   We also evaluate $\ell_{\mathrm{LM}}(\bm{z}^{OVR})/\ell_{\mathrm{LM}}(\bm{z}^{CE})$ at the last epoch, which is expected to be about two from Theorem~\ref{lmthm}.
   Table~\ref{DivLm} list $\ell_{\mathrm{LM}}(\bm{z}^{OVR})/\ell_{\mathrm{LM}}(\bm{z}^{CE})$ at the last epoch, and it is about two,
   and thus, logit margin losses in adversarial training follow Theorem~\ref{lmthm} well even though we assume a simple problem that only considers one data point and 
   assumes that logits are directly moved by the gradient for Theorem~\ref{lmthm}.
   Since the number of classes $K$ of CIFAR100 is 100 and larger than other datasets,
   the logit margins of cross-entropy is larger than OVR at the beginning of training. 
   This result corresponds to the case of CE ($K=100$) in \rfig{LMSim}, and this phenomenon is also able to be explained by the simple problem~\req{ToyProb}.
   To the best of our knowledge, this is the first study that explicitly reveals the logit margin of minimization of cross-entropy depends on the number of classes.
   Though the logit margin loss of OVR in Theorem~\ref{lmthm} does not depend on $K$,
   the logit margins of OVR on CIFAR100 is smaller than those on CIFAR10 and SVHN.
   This is because CIFAR100 is a more difficult dataset than CIFAR10 and SVHN: robust accuracies of CIFAR100 is about 25~\% whereas those of CIFAR10 and SVHN are about 50~\% in \rtab{PropAA}.
   \begin{table}[tb]
     \centering
     \begin{tabular}{ccccc}\toprule
           CIFAR10 (RN18)&CIFAR10 (WRN)&CIFAR100&SVHN\\\midrule
           1.87&1.78&1.56&1.76\\
           \bottomrule
     \end{tabular}
     \caption{$\ell_{\mathrm{LM}}(\bm{z}^{OVR})/\ell_{\mathrm{LM}}(\bm{z}^{CE})$ with $w=1.0$ at the last epoch}\label{DivLm}
   \end{table}
   \subsection{Effects of Hyperparameters $\lambda$}\label{hypSub}
\begin{figure}
     \centering
     \begin{subfigure}[t]{0.32\linewidth}
       \includegraphics[width=\linewidth]{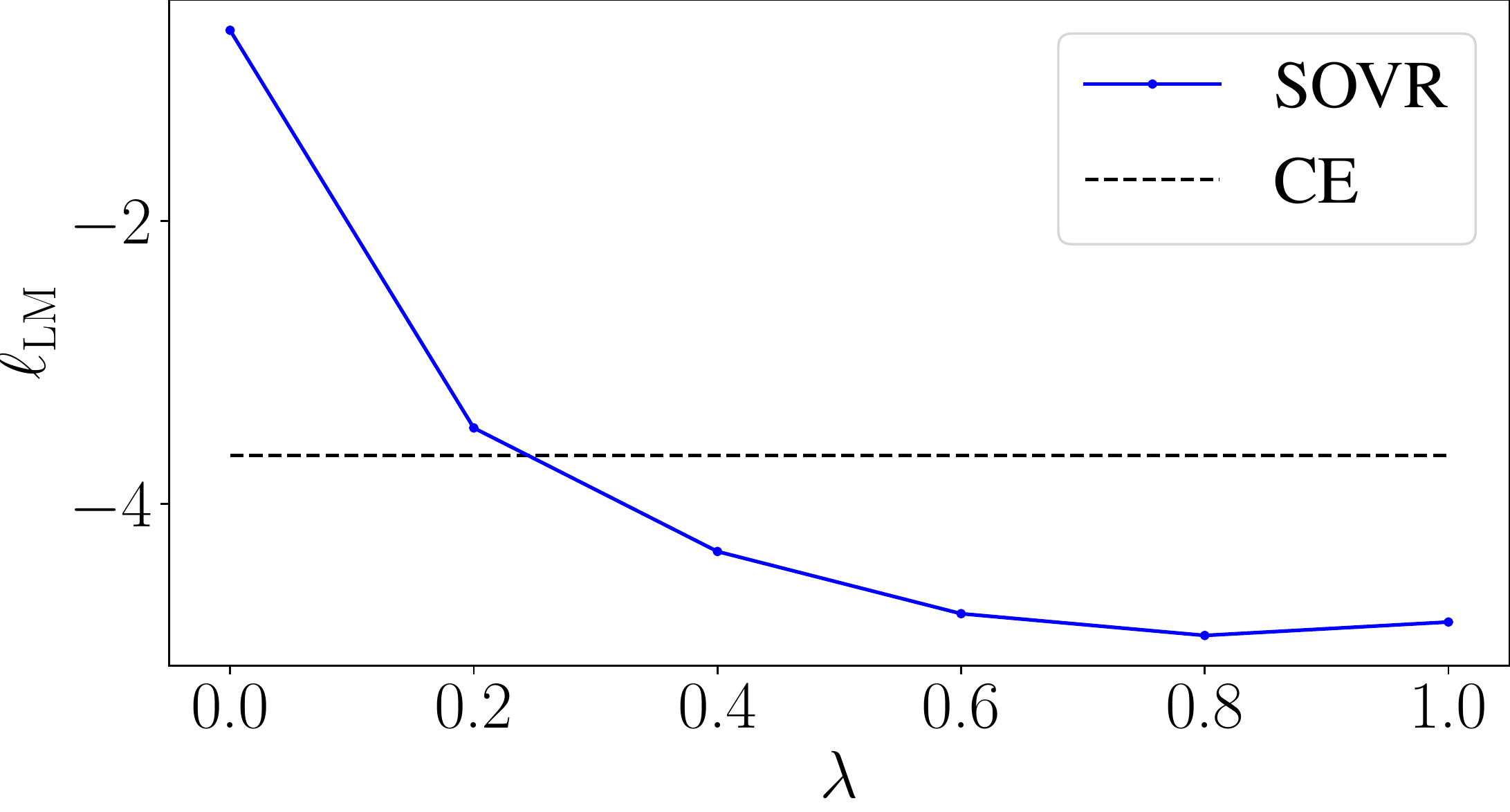}
       \caption{$\ell_{\mathrm{LM}}(\bm{x}^\prime)$ Vs $\lambda$}\label{lamVsLM}
     \end{subfigure}  
     \hfill
       \begin{subfigure}[t]{.32\linewidth}\centering
         \includegraphics[width=\linewidth]{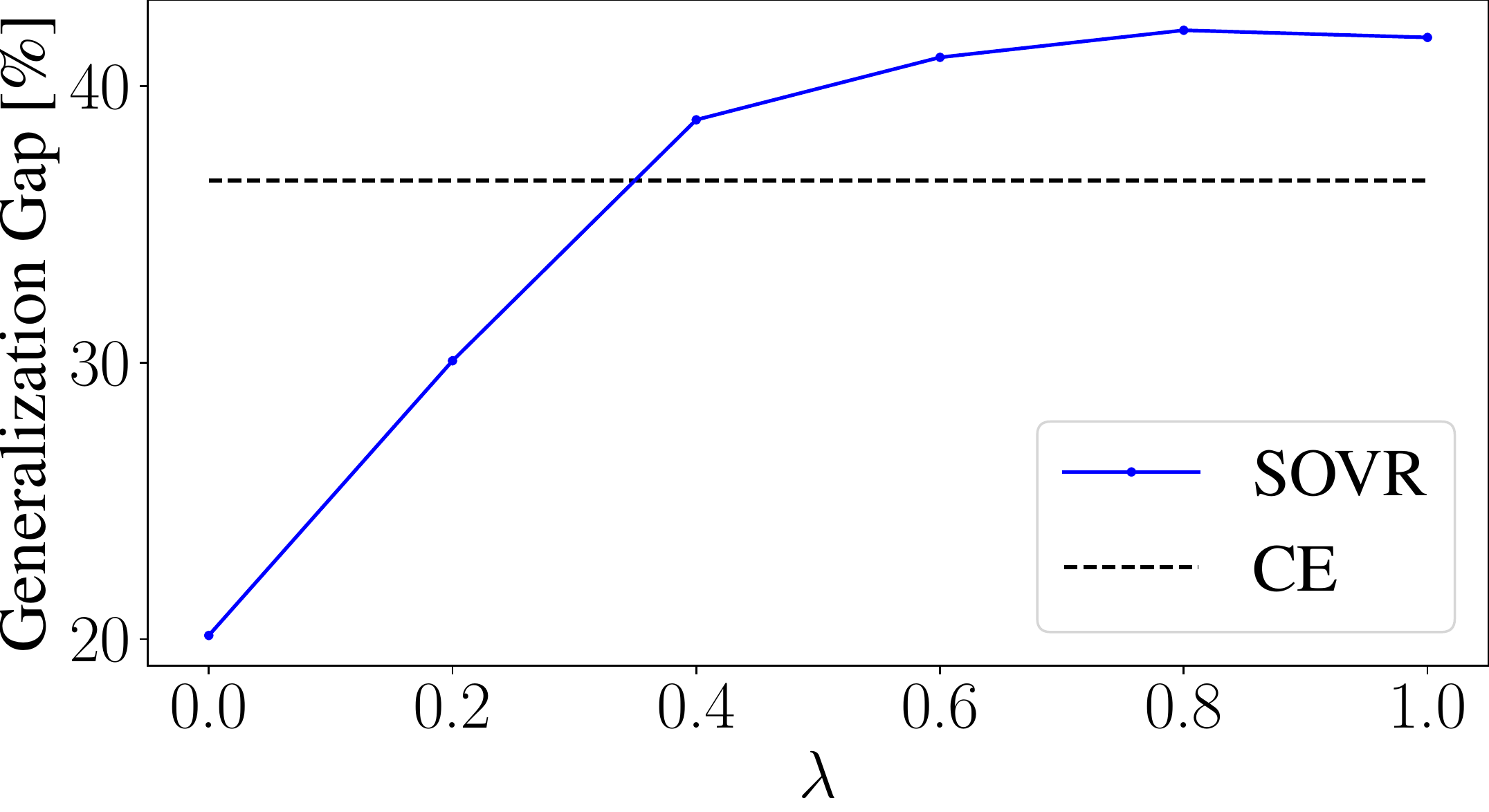}
         \caption{Generalization Gap Vs $\lambda$}\label{lamVsGap}
       \end{subfigure}
       \hfill
             \begin{subfigure}[t]{0.32\linewidth}
       \includegraphics[width=\linewidth]{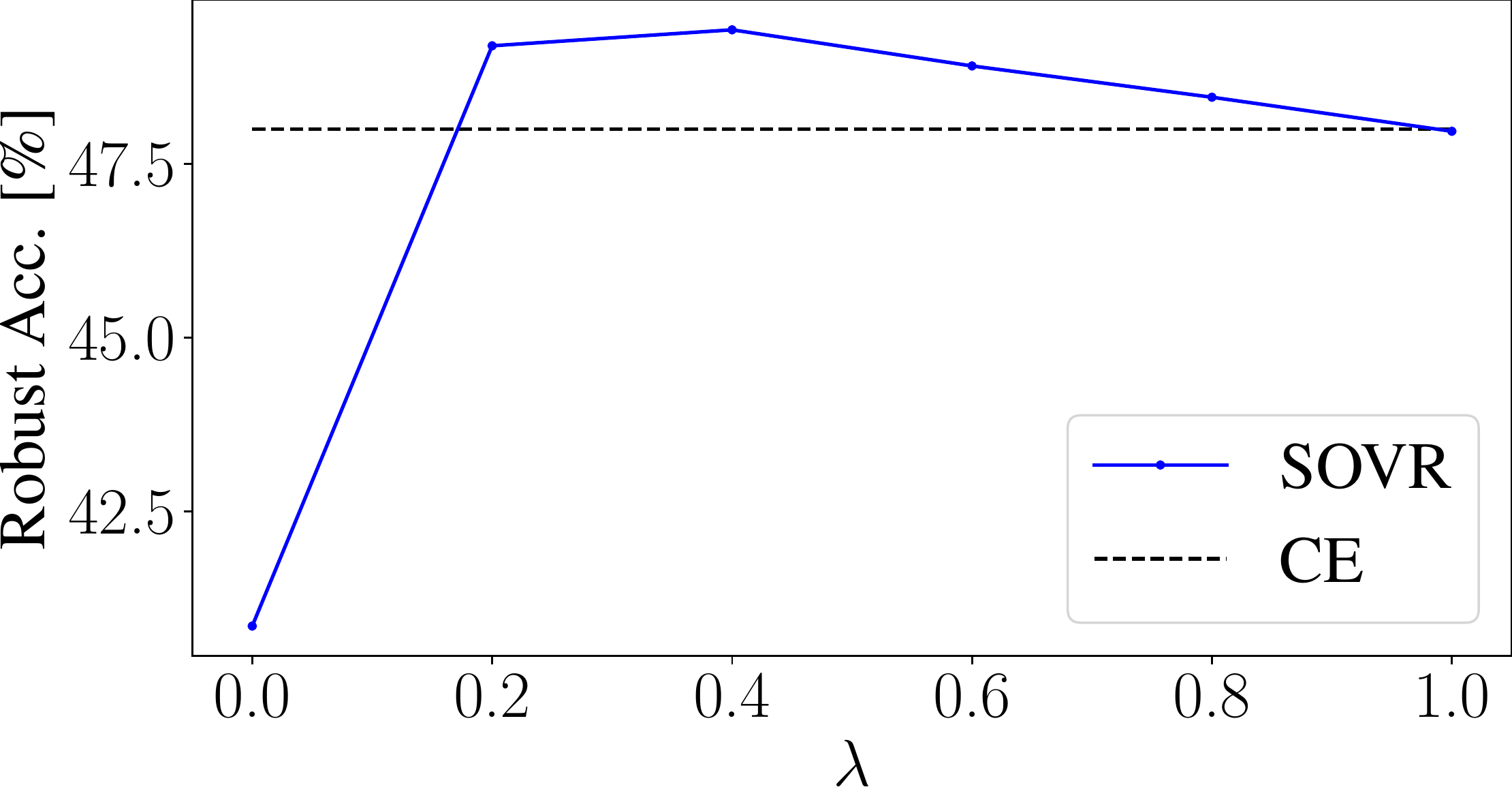}
       \caption{Robust Acc. Vs $\lambda$}\label{lamVsAcc}
     \end{subfigure}
     \caption{The effect of rate of applying OVR $\lambda$. $M$ is set to 40. 
     Generalization gap is a gap between training robust accuracy and test robust accuracy against PGD ($\mathcal{K}$=20) at the last epoch.
     Robust Acc. is robust accuracy against Auto-Attack.  Dashed gray line corresponds to the results of AT using cross-entropy loss.}
     \label{lamPara}
   \end{figure}

    SOVR has hyperparameters ($M$, $\lambda$).
    In this section, we evaluate the effects of $\lambda$.
    Figure~\ref{lamPara} plots $\ell_{LM}$ on CIFAR10 with RN18, generalization gap, and robust accuracy against Auto-Attack.
    We set $M$ to 40.
    Note that $\lambda=0$ corresponds to that models are trained on only a set of $\mathbb{S}$, i.e., AT only using the 60 percent of the data points in minibatch when $M=40$.
    First, $\ell_{LM}(\bm{x}^\prime)$ is monotonically decreasing due to increases in $\lambda$ (\rfig{lamVsLM}).
    However, robust accuracies against Auto-Attack are not monotonically increasing against $\lambda$ (\rfig{lamVsAcc}).
   This is because generalization gap increases (\rfig{lamVsGap}).
    Thus, too high weights on important samples causes overfitting.
    SOVR is always superior or comparable to AT in terms of robustness against Auto-Attack under all tested values of ($0<M\leq 100$, $0<\lambda\leq 1$). 
 
    \subsection{Individually Test of Auto-Attack}\label{AASub}
    \begin{figure}
      \centering
      \includegraphics[width=\linewidth]{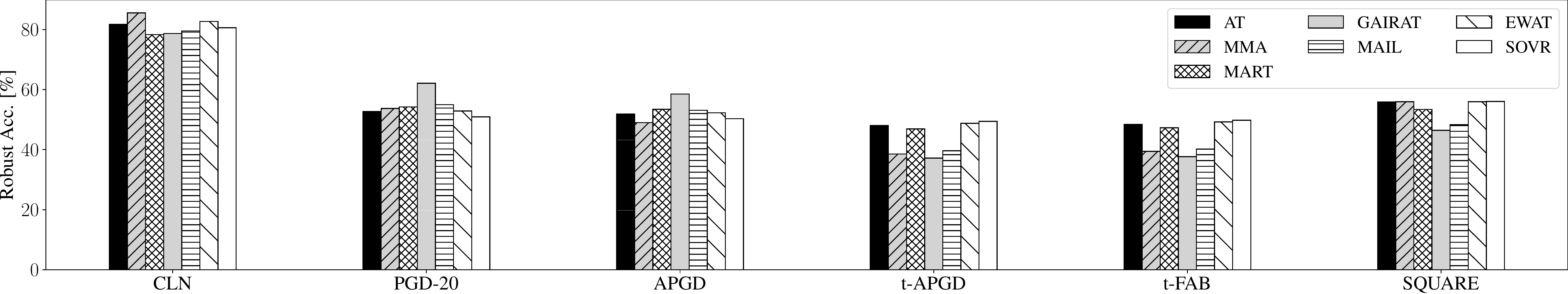}
      \caption{The robustness against PGD-20 and Auto-Attack on the test set of CIFAR10. We decompose the robust accuracy for Auto-Attack into
      robust accuracy in each phase.}
      \label{AllATDis}
    \end{figure}
    For importance-aware methods,
    we evaluate the robust accuracies against all components of Auto-Attack in Section~\ref{AASec}.
    In this section, we additionally evaluate EWAT by individually using Auto-Attack and discuss the results of SOVR.
    Figure~\ref{AllATDis} plots the results, and SOVR is the most robust against t-APG and t-FABand.
    In addition, it is more robust against SQUARE than AT and EWAT.
    Although the robust accuracy of SOVR against PGD-20 is lower than those of AT and EWAT,
    SOVR outperforms other methods in terms of the robustness against the worst-case attacks,
    which is the goal of this study.
 \subsection{Evaluation Using Various Attacks}\label{VarSub}
    We list robust accuracies against various attacks;
    FGSM~\citep{fgsm}, 100-step PGD~\citep{pgd2}, 100-step PGD with CW loss~\citep{pgd2,candw}, SPSA~\citep{SPSA} in \rtab{VarAttack}.
    Hyperparameters of SPSA are as follows: the number of steps is set to 100, the perturbation
    size is set to 0.001, learning rate is set to 0.01, and the number of samples for each gradient estimation is set to 256.
    In this table, we repeat the clean accuracies and robust accuracies against Auto-Attack from the table in the main paper.
    In addition, we list the worst robust accuracies, which are 
    the least robust accuracy among attacks in the table for each method.
    In this table, importance-aware methods tend to fail to improve the robustness against SPSA.
    Since SPSA does not directly use gradients, this result indicates that importance-aware methods improve the robustness by obfuscating gradients~\citep{best}.
    Against some attacks, MMA achieves the highest robust accuracy on several datasets.
    However, our goal is improving the true robustness, i.e., robust accuracies against the worst-case attacks in $\bm{\delta}\in \{||\bm{\delta}||_{\infty}\leq 8/255\}$.
    MMA does not improve the robustness against the worst-case attacks (the columns of Worst). 
    We can see that Auto-Attack always achieves the least robust accuracies, and SOVR improves them:
    Robust accuracies against Auto-Attack of SOVR are 5.9-12.2 percent points greater than those of MMA.
    \begin{table}[tbp]
      \centering
      \caption{Robust Accuracy against various attacks ($L_\infty$, $\varepsilon=8/255$). CLN denotes accuracy on clean data, and AA denotes Auto-Attack. 
      Worst represents the least robust accuracy among attacks in the table for each method.  }
      \label{VarAttack}
      \resizebox{\linewidth}{!}{
      \begin{tabular}{ccccccccc}\toprule
      &Method&CLN&FGSM&PGD&CW&SPSA&AA&Worst \\ \midrule
      &AT&81.6$\pm$0.5&57.6$\pm$0.1&52.5$\pm$0.4&50.0$\pm$0.4&56.8$\pm$0.2&48.0$\pm$0.2&48.0$\pm$0.2 \\\cmidrule(r){2-8}\cmidrule(r){9-9}
      &MART&78.3$\pm$1&58.0$\pm$0.3&54.0$\pm$0.1&48.7$\pm$0.2&54.2$\pm$0.1&46.9$\pm$0.3&46.9$\pm$0.3 \\ \cmidrule(r){2-8}\cmidrule(r){9-9}
      &MMA&$\bm{85.5\pm0.7}$&$\bm{65.5\pm2}$&51.6$\pm$0.2&51.0$\pm$0.6&56.3$\pm$1&37.2$\pm$0.9&37.2$\pm$0.9 \\ \cmidrule(r){2-8}\cmidrule(r){9-9}
      C10&GAIRAT&78.7$\pm$0.7&63.1$\pm$0.7&$\bm{62.0\pm0.4}$&40.01$\pm$1&47.4$\pm$1&37.7$\pm$1&37.7$\pm$1 \\ \cmidrule(r){2-8}\cmidrule(r){9-9}
      (RN18)&MAIL&79.5$\pm$0.4&57.8$\pm$0.1&54.97$\pm$0.08&42.1$\pm$0.2&49.1$\pm$0.4&39.6$\pm$0.4 &39.6$\pm$0.4\\ \cmidrule(r){2-8}\cmidrule(r){9-9}
      &EWAT&82.8$\pm$0.4&57.7$\pm$0.4&52.3$\pm$0.4&50.4$\pm$0.7&56.8$\pm$0.2&48.2$\pm$0.7 &48.2$\pm$0.7\\ \cmidrule(r){2-8}\cmidrule(r){9-9}
      &SOVR&81.9$\pm$0.2&57.0$\pm$0.2&50.9$\pm$0.5&$\bm{51.5\pm0.2}$&$\bm{57.7\pm0.2}$&$\bm{49.4\pm0.3}$ &$\bm{49.4\pm0.3}$\\ 
      \midrule
      &AT&85.6$\pm$0.5&60.9$\pm$0.4&55.1$\pm$0.4&54.0$\pm$0.6&60.8$\pm$0.5&51.9$\pm$0.5&51.9$\pm$0.5\\ \cmidrule(r){2-8}\cmidrule(r){9-9}
      &MART&81.5$\pm$1&61.3$\pm$0.6&57.2$\pm$0.2&52.1$\pm$0.3&57.8$\pm$0.6&50.44$\pm$0.09&50.44$\pm$0.09\\ \cmidrule(r){2-8}\cmidrule(r){9-9}
      &MMA&$\bm{87.8\pm1}$&$\bm{68.6\pm1}$&55.7$\pm$1&$\bm{55.4\pm0.7}$&59.6$\pm$2&43.1$\pm$0.6&43.1$\pm$0.6 \\ \cmidrule(r){2-8}\cmidrule(r){9-9}
      C10&GAIRAT&83.0$\pm$0.7&64.1$\pm$0.5&$\bm{62.9\pm0.4}$&44.4$\pm$0.7&52.1$\pm$0.5&41.8$\pm$0.6&41.8$\pm$0.6 \\ \cmidrule(r){2-8}\cmidrule(r){9-9}
      (WRN)&MAIL&82.2$\pm$0.4&59.3$\pm$0.5&56.0$\pm$0.5&45.7$\pm$0.2&53.0$\pm$0.2&43.3$\pm$0.1 &43.3$\pm$0.1\\ \cmidrule(r){2-8}\cmidrule(r){9-9}
      &EWAT&86.0$\pm$0.5&60.6$\pm$0.4&54.5$\pm$0.1&53.8$\pm$0.3&60.7$\pm$0.4&51.6$\pm$0.3&51.6$\pm$0.3\\\cmidrule(r){2-8}\cmidrule(r){9-9}
      &SOVR&85.0$\pm$1&60.8$\pm$0.1&54.5$\pm$0.2&55.2$\pm$0.2&$\bm{61.6\pm0.1}$&$\bm{53.1\pm0.2}$ &$\bm{53.1\pm0.2}$\\ 
      \midrule
      &AT&89.8$\pm$0.6&30.1$\pm$0.4&27.7$\pm$0.2&25.6$\pm$0.3&29.3$\pm$0.3&23.7$\pm$0.3&23.7$\pm$0.3\\ \cmidrule(r){2-8}\cmidrule(r){9-9}
      &MART&86.9$\pm$0.6&$\bm{31.0\pm0.2}$&$\bm{29.36\pm0.06$}&25.4$\pm$0.3&28.5$\pm$0.1&23.9$\pm$0.3&23.9$\pm$0.3 \\\cmidrule(r){2-8}\cmidrule(r){9-9}
      &MMA&$\bm{93.9\pm0.4}$&25.7$\pm$0.3&19.4$\pm$0.2&20.5$\pm$0.1&24.3$\pm$0.3&18.4$\pm$0.2&18.4$\pm$0.2\\ \cmidrule(r){2-8}\cmidrule(r){9-9}
      C100&GAIRAT&89.9$\pm$0.4&27.9$\pm$0.3&26.0$\pm$0.2&21.9$\pm$0.4&25.9$\pm$0.1&19.8$\pm$0.5&19.8$\pm$0.5 \\ \cmidrule(r){2-8}\cmidrule(r){9-9}
      &MAIL&89.4$\pm$0.4&24.81$\pm$0.08&23.29$\pm$0.06&18.3$\pm$0.5&21.9$\pm$0.5& 16.7$\pm$0.3&16.7$\pm$0.3\\ \cmidrule(r){2-8}\cmidrule(r){9-9}
      &EWAT&90.2$\pm$0.6&30.07$\pm$0.08&27.4$\pm$0.3&25.3$\pm$0.2&29.3$\pm$0.1&23.52$\pm$0.06&23.52$\pm$0.06\\\cmidrule(r){2-8}\cmidrule(r){9-9}
      &SOVR&90.0$\pm$1&30.2$\pm$0.2&27.4$\pm$0.2&$\bm{26.1\pm0.1}$&$\bm{29.9\pm0.1}$&$\bm{24.3\pm0.2}$&$\bm{24.3\pm0.2}$\\ 
      \midrule
      &AT&53.0$\pm$0.7&61.1$\pm$0.5&50.6$\pm$0.4&47.7$\pm$0.8&55.7$\pm$0.9&45.6$\pm$0.4&45.6$\pm$0.4 \\ \cmidrule(r){2-8}\cmidrule(r){9-9}
      &MART&49.2$\pm$0.1&64.4$\pm$0.5&56.5$\pm$0.2&49.0$\pm$0.4&56.69$\pm$0.08&46.9$\pm$0.3&46.9$\pm$0.3 \\ \cmidrule(r){2-8}\cmidrule(r){9-9}
      &MMA&$\bm{60.6\pm0.6}$&$\bm{79.6\pm0.8$}&$\bm{65.0\pm1}$&$\bm{59.1\pm 2}$&$\bm{63.2\pm2}$&41.0$\pm$0.3&41.0$\pm$0.3 \\ \cmidrule(r){2-8}\cmidrule(r){9-9}
    SVHN&GAIRAT&52.0$\pm$0.5&65.8$\pm$0.4&60.4$\pm$0.6&40.6$\pm$0.7&48.8$\pm$0.6&37.6$\pm$0.6&37.6$\pm$0.6 \\ \cmidrule(r){2-8}\cmidrule(r){9-9}
      &MAIL&46.5$\pm$0.5&64.6$\pm$0.4&58.2$\pm$0.3&44.1$\pm$0.7&52.3$\pm$0.5&41.2$\pm$0.3&41.2$\pm$0.3\\ \cmidrule(r){2-8}\cmidrule(r){9-9}
      &EWAT&54.2$\pm$1&61.8$\pm$0.5&51.7$\pm$0.2&50.2$\pm$0.4&57.4$\pm$0.4&47.6$\pm$0.4&47.6$\pm$0.4\\\cmidrule(r){2-8}\cmidrule(r){9-9}
      &SOVR&52.1$\pm$0.8&65.3$\pm$2&50.7$\pm$0.2&52.5$\pm$0.2&60.7$\pm$0.4& $\bm{48.5\pm0.4}$&$\bm{48.5\pm0.4}$\\ 
      \bottomrule
      \end{tabular}
      }
    \end{table}
\begin{figure}
     \centering
     \begin{subfigure}[t]{0.45\linewidth}
       \includegraphics[width=\linewidth]{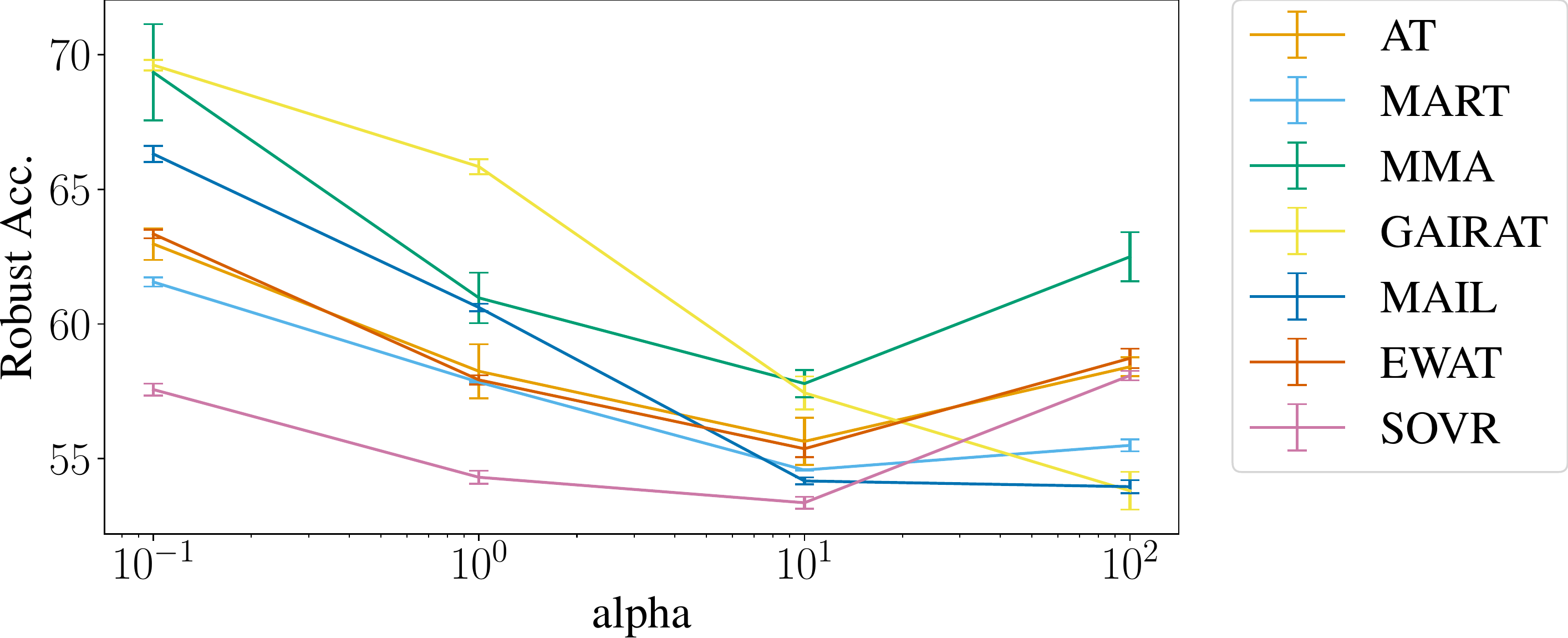}
       \caption{CIFAR10 (RN18)}\label{lgsc10}
     \end{subfigure}  
     \hfill
       \begin{subfigure}[t]{.45\linewidth}\centering
         \includegraphics[width=\linewidth]{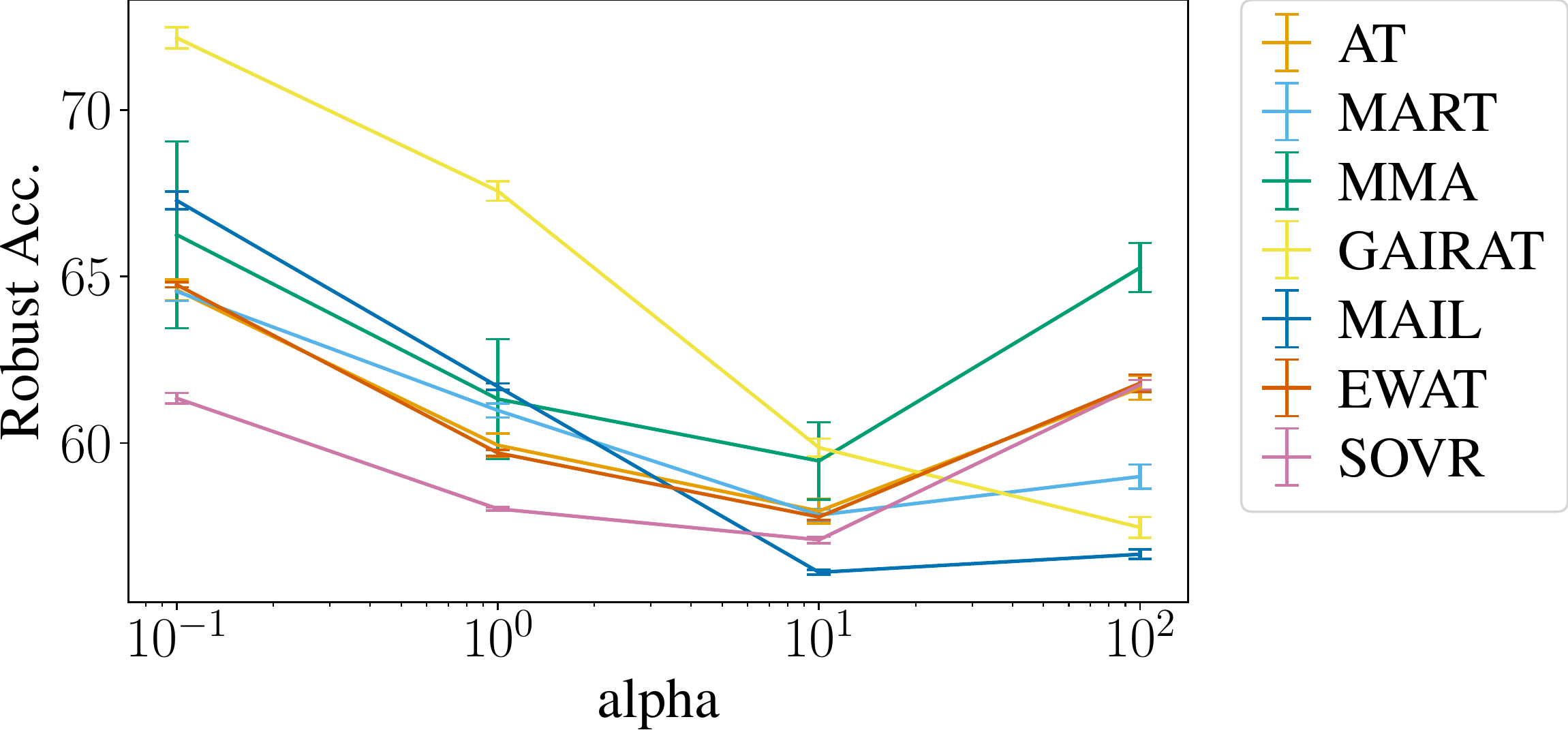}
         \caption{CIFAR10 (WRN)}\label{lgsc10wrn}
       \end{subfigure}
       \hfill
             \begin{subfigure}[t]{0.45\linewidth}
       \includegraphics[width=\linewidth]{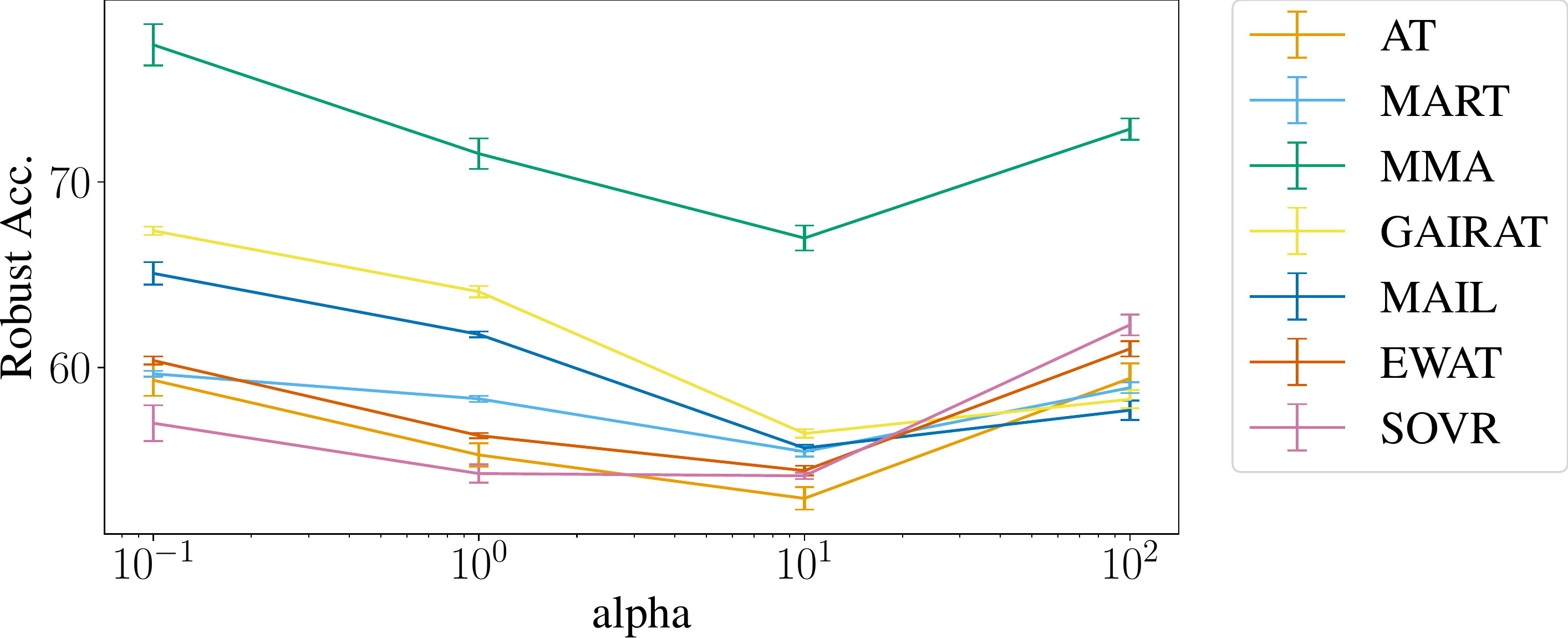}
       \caption{SVHN}\label{lgsSVHN}
     \end{subfigure}      \hfill
     \begin{subfigure}[t]{0.45\linewidth}
       \includegraphics[width=\linewidth]{Figure/LGS/SVHNLogitScaleAttack.pdf}
       \caption{CIFAR100}\label{lgsc100}
     \end{subfigure}
     \caption{Robust Accuracy against logit scaling attack.}
     \label{LGS}
   \end{figure}  
   \subsection{Evaluation using Logit Scaling Attack}
   In this section, we evaluate the robustness against the logit scaling attack \citep{hitaj2021evaluating}. 
   \cite{hitaj2021evaluating} reveals that GAIRAT tends to be vulnerable to logit scaling attacks.
   The logit scaling attack multiplies logits by $\alpha$ before applying softmax when generating PGD attacks.
   We set $\alpha=[0.1, 1.0, 10, 100]$.
 \rfig{LGS} plots robust accuracy against $\alpha$.
 This figure shows that the robust accuracies of GAIRAT and MAIL tend to decrease when increasing $\alpha$. 
 Though the robust accuracy of SOVR is the lowest on CIFAR 10 (RN18), it is higher than the robust accuracy against Auto-Attack.
 Thus, the logit scaling attack is not the worst-case attack.
 Since the robust accuracy of SOVR does not necessarily decrease when increasing $\alpha$,
 the results seem to be caused by high robustness against PGD (\rtab{VarAttack}) of other methods rather than vulnerability to logit scaling of SOVR.
 Previous methods tend to be designed to increase the robustness against PGD since Auto-Attack is a relatively recent attack.
 On the other hand, SOVR is designed to increase the robustness against the worst-case attack, which is Auto-Attack for now.
 \subsection{Evaluation using Auto-Attack with larger magnitudes}
 In this section, we evaluate the robustness against Auto-Attack with $\varepsilon=12/255$ and $16/255$. 
 These magnitudes are larger than the magnitude used in adversarial training, 
 and thus, this experiment evaluates the existence of overfitting to the magnitude used in training.
 \rtab{StrongAA} lists robust accuracies against Auto-Attack with $\varepsilon=12/255$ and $16/255$.
 This table shows that SOVR outperforms AT, and TSOVR achieves the largest robust accuracies.
 Thus, our methods do not overfit to the magnitude of attacks used in training.
 Interestingly, this table also shows that MART outperforms AT while the robust accuracies against $\varepsilon\!=\!8/255$ of MART are not always larger than those of AT (\rtab{PropAA}).
 MART might have better generalization performance for the magnitudes of attacks.
 \begin{table*}[tb]
  \centering
  \caption{Robust accuracy against Auto-Attack using $\varepsilon=12/255$ and $16/255$.}
  \resizebox{\linewidth}{!}{
  \begin{tabular}{lcccccccccc}\toprule
     \multicolumn{10}{c}{Robust Accuracy against Auto-Attack ($L_\infty$, adversarialy trained by using PGD with $\varepsilon\!=\!8/255$)}\\ \cmidrule(r){2-2}\cmidrule(r){3-8}\cmidrule(r){9-10}
    &AT&MART&MMA&GAIRAT&MAIL&EWAT&SOVR&TRADES &TSOVR\\ \midrule
  CIFAR10 (RN18, $\varepsilon\!=\!12/255$)&28.6$\pm$0.6&29.6$\pm$0.8&19.7$\pm$0.9&19$\pm$1&20.2$\pm$0.5&28.4$\pm$0.8&30.3$\pm$0.3&30.6$\pm$0.5&$\bm{31.9\pm0.4}$ \\
  CIFAR10 (RN18, $\varepsilon\!=\!16/255$)&13.6$\pm$0.6&15$\pm$1&8.8$\pm$0.5&7.8$\pm$0.5&7.9$\pm$0.5&13.2$\pm$0.2&14.3$\pm$0.1&16.0$\pm$0.8&$\bm{17.0\pm0.5}$\\
  \midrule
  CIFAR10 (WRN, $\varepsilon\!=\!12/255$)&31.2$\pm$0.2&32$\pm$1&25$\pm$1&22$\pm$1&23.0$\pm$0.2&31.1$\pm$0.3&32.7$\pm$0.3&34.8$\pm$0.2&$\bm{35.5\pm0.3}$\\
  CIFAR10 (WRN, $\varepsilon\!=\!16/255$)&15.3$\pm$0.3&17$\pm$1&13$\pm$1&9.1$\pm$0.7&9.43$\pm$0.4&14.8$\pm$0.1&16.1$\pm$0.4&19.6$\pm$0.1&$\bm{20.4\pm0.6}$\\
  \midrule
  SVHN (RN18, $\varepsilon\!=\!12/255$)&23.0$\pm$0.5&24$\pm$1&23$\pm$1&18.2$\pm$0.7&19$\pm$1&25.9$\pm$0.7&25$\pm$1&29.0$\pm$0.6&$\bm{29.4\pm0.5}$ \\
  SVHN (RN18, $\varepsilon\!=\!16/255$)&10.6$\pm$0.5&11.4$\pm$0.7&14$\pm$1&7.9$\pm$0.3&8.1$\pm$0.2&12.0$\pm$0.4&10.5$\pm$0.6&15.2$\pm$0.7&$\bm{15.4\pm0.7}$\\
  \midrule
  CIFAR100 (RN18, $\varepsilon\!=\!12/255$)&13.8$\pm$0.1&15.2$\pm$0.3&10.0$\pm$0.3&10.45$\pm$0.5&8.5$\pm$0.4&13.4$\pm$0.1&14.5$\pm$0.3&13.9$\pm$0.3&$\bm{16.0\pm0.2}$ \\
  CIFAR100 (RN18, $\varepsilon\!=\!16/255$)&7.53$\pm$0.03&9.4$\pm$0.1&5.9$\pm$0.3&5.1$\pm$0.4&4.1$\pm$0.4&7.1$\pm$0.1&8.1$\pm$0.1&8.2$\pm$0.1&$\bm{10.0\pm0.1}$\\
  \bottomrule
  \end{tabular}
  }
  \label{StrongAA}
\end{table*}\vspace{-1pt}
 \subsection{Dependence of the Number of Classes}\label{C100Sec}
 Figure~\ref{Poten} shows that the rate of AT gets close to SOVR on CIFAR100.
 This is because the number of classes of CIFAR100 ($K=100$) is ten times larger than other datasets ($K=10$),
 and logit margins of cross-entropy depend on the number of classes $K$ (\req{CEOVRApp}).
 Thus, this result is a piece of evidence that Theorem~\ref{lmthm} explains the difference of logit margins between OVR and cross-entropy.
 In certain finite time step $t$ (not the limit), Eqs. (\ref{LMOVRApp}) and (\ref{CEOVRApp}) show that the difference between OVR and cross-entropy depends on the number of classes $K$.
 Even so, \rfig{TrajC100} shows that the increase rate of logit margins of OVR is larger than that of cross-entropy against epochs.
 To achieve better performance, we can tune the hyper-parameter $\lambda$, which corresponds to $w_1$ in \req{LMOVRApp} of Theorem~\ref{lmthm}.
 When using $\lambda$= 0.6, SOVR achieves better robustness than $\lambda=0.5$ on CIFAR100 (\rtab{C100TuneTab}). 
 Cross-entropy also has the weight $w_2$ in \req{CEOVRApp}, and it is automatically tuned in GAIRAT, MAIL, and EWAT. 
 However, this tuning does not achieve comparable performance to SOVR.
 \begin{table}[tb]
   \centering
   \caption{Robust accuracy against Auto-Attack on CIFAR100 tuning $\lambda$.}\label{C100TuneTab}  
   \begin{tabular}{cccc}\toprule
      \multicolumn{4}{c}{Robust Accuracy against Auto-Attack ($L_\infty$, $\varepsilon\!=\!8/255$)}\\\midrule
     &AT&SOVR ($\lambda$=0.5)&SOVR ($\lambda$=0.6) \\ \midrule
   CIFAR100 (RN18)&23.7$\pm$0.3&24.3$\pm$0.2&$\bm{24.6\pm 0.1}$\\
    \midrule\midrule \multicolumn{4}{c}{Clean Accuracy}\\\midrule
 CIFAR100 (RN18)&$\bm{53.0\pm0.7}$&52.1$\pm$0.8&51.9 $\pm$ 0.6\\
   \bottomrule
   \end{tabular}
 \end{table}
 
  \subsection{Histogram of Probabilistic Margin Losses}\label{ProbMargin}
\begin{figure}[tbp]
     \begin{subfigure}[t]{0.16\linewidth}
         \includegraphics[width=\linewidth]{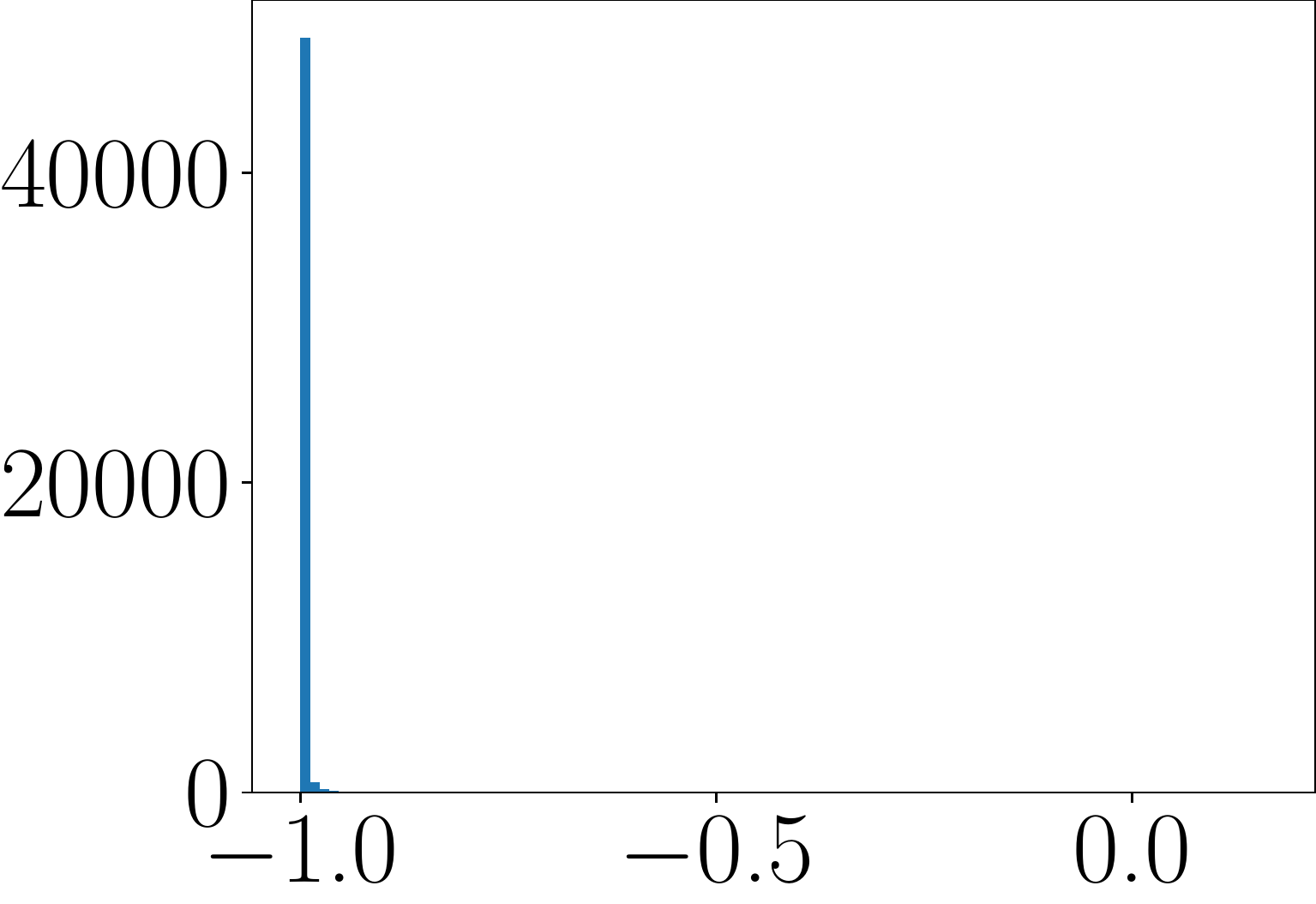}
         \caption{ST}\label{PMHist-ST}
     \end{subfigure}
     \begin{subfigure}[t]{0.16\linewidth}
       \includegraphics[width=\linewidth]{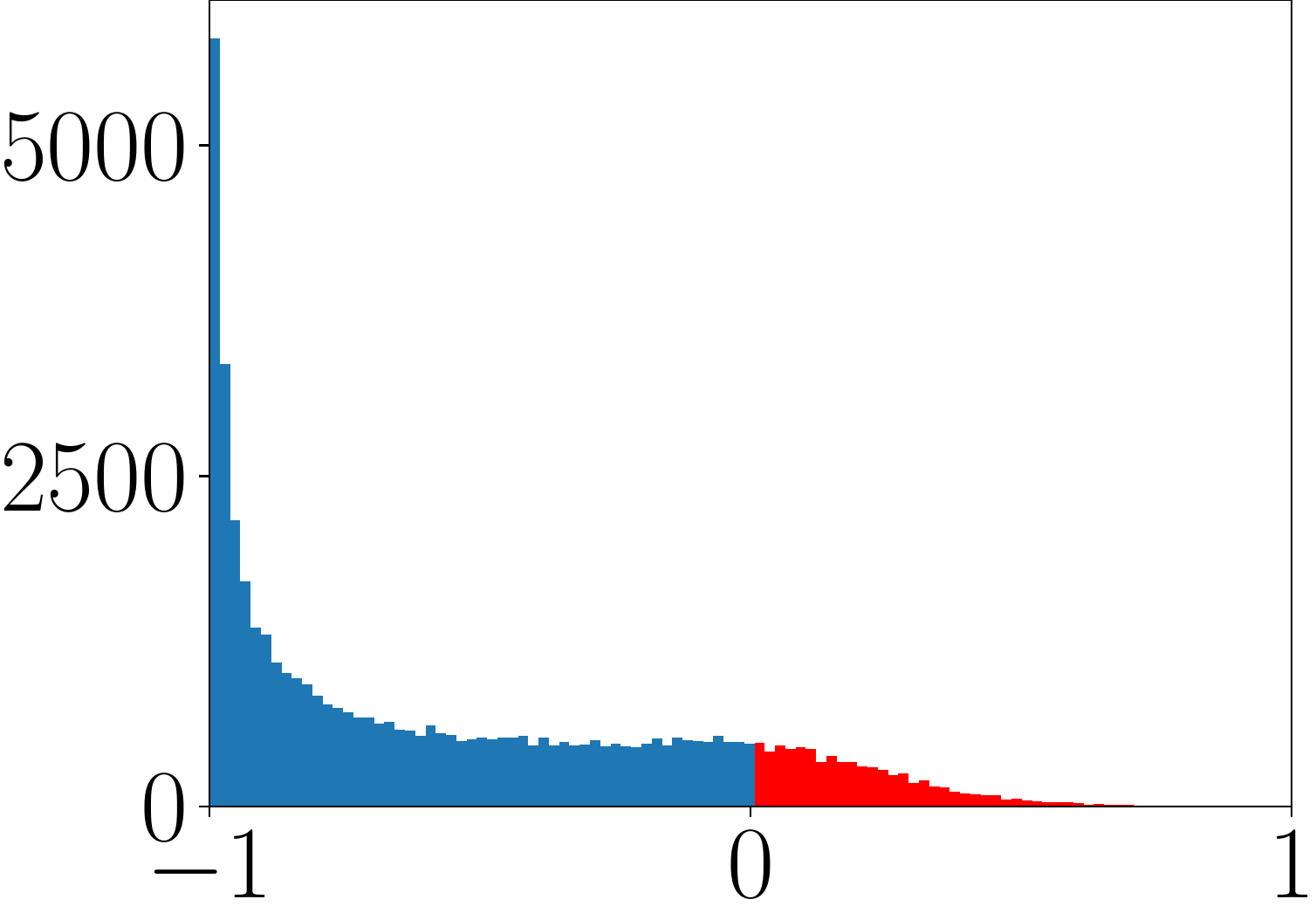}
       \caption{AT}\label{PMHist-AT}
   \end{subfigure}
   \begin{subfigure}[t]{0.16\linewidth}
     \includegraphics[width=\linewidth]{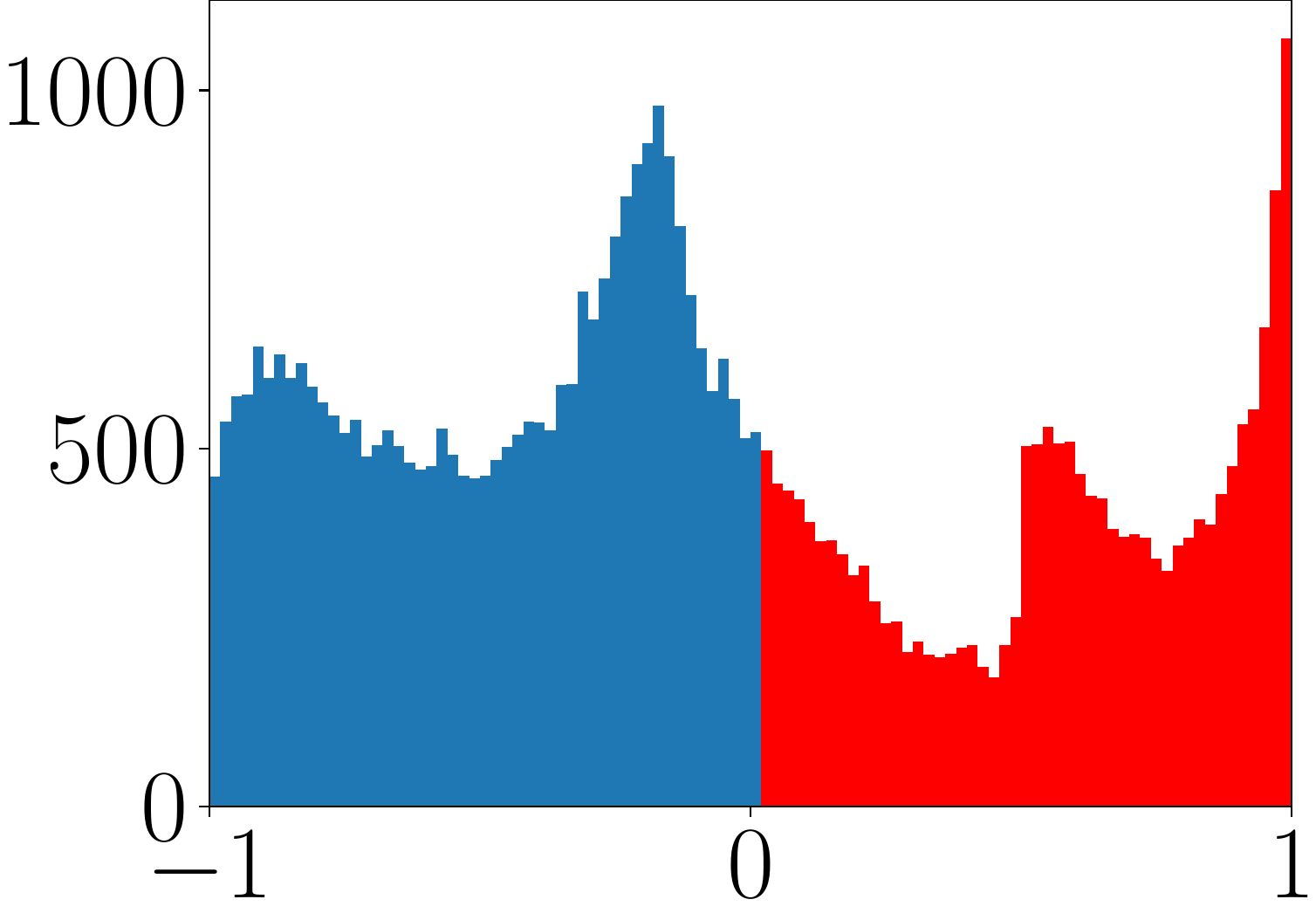}
     \caption{MMA}\label{PMHist-MMA}
   \end{subfigure}
   \begin{subfigure}[t]{0.16\linewidth}
     \includegraphics[width=\linewidth]{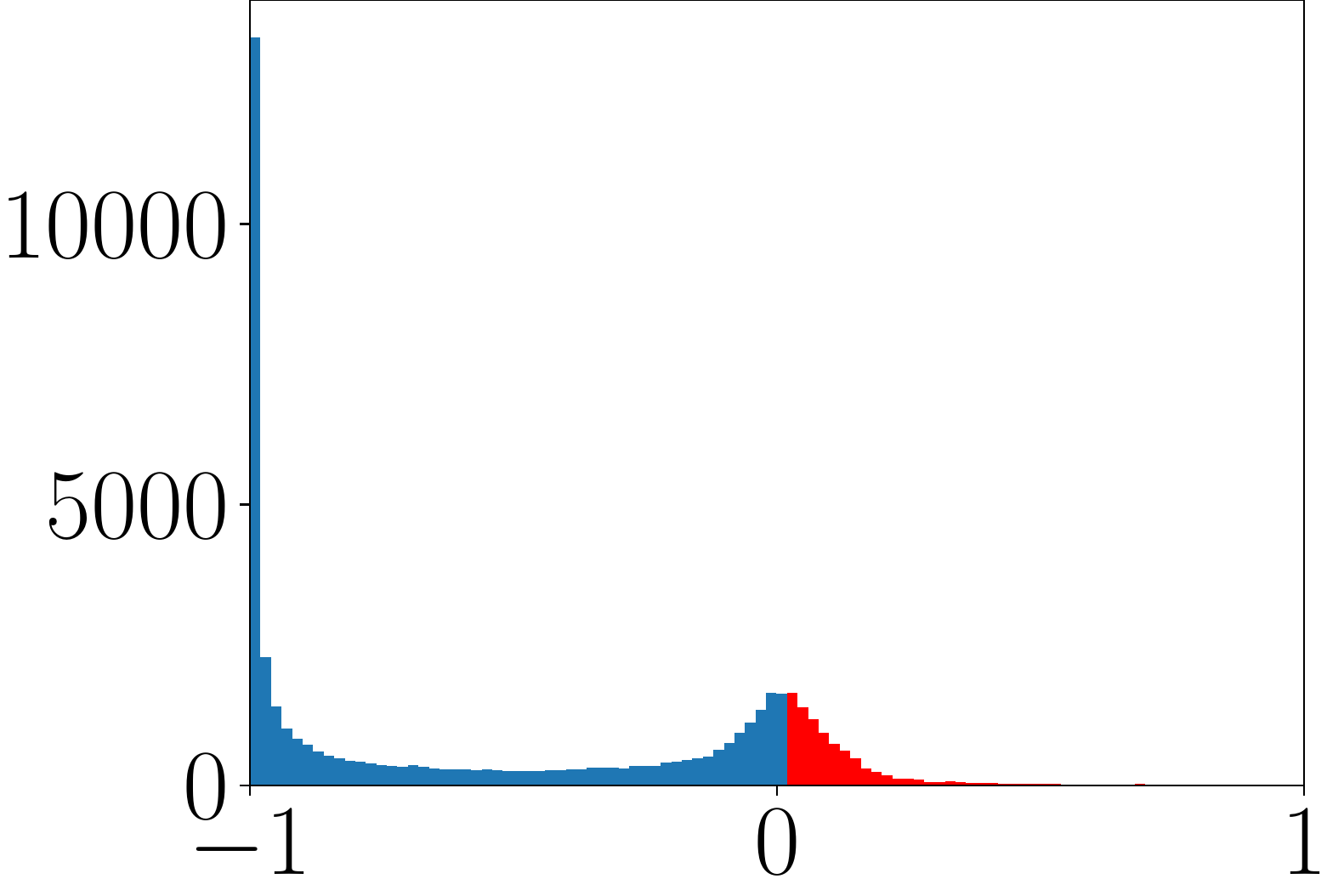}
     \caption{MART}\label{PMHist-MART}
   \end{subfigure}
   \begin{subfigure}[t]{0.163\linewidth}
     \includegraphics[width=\linewidth]{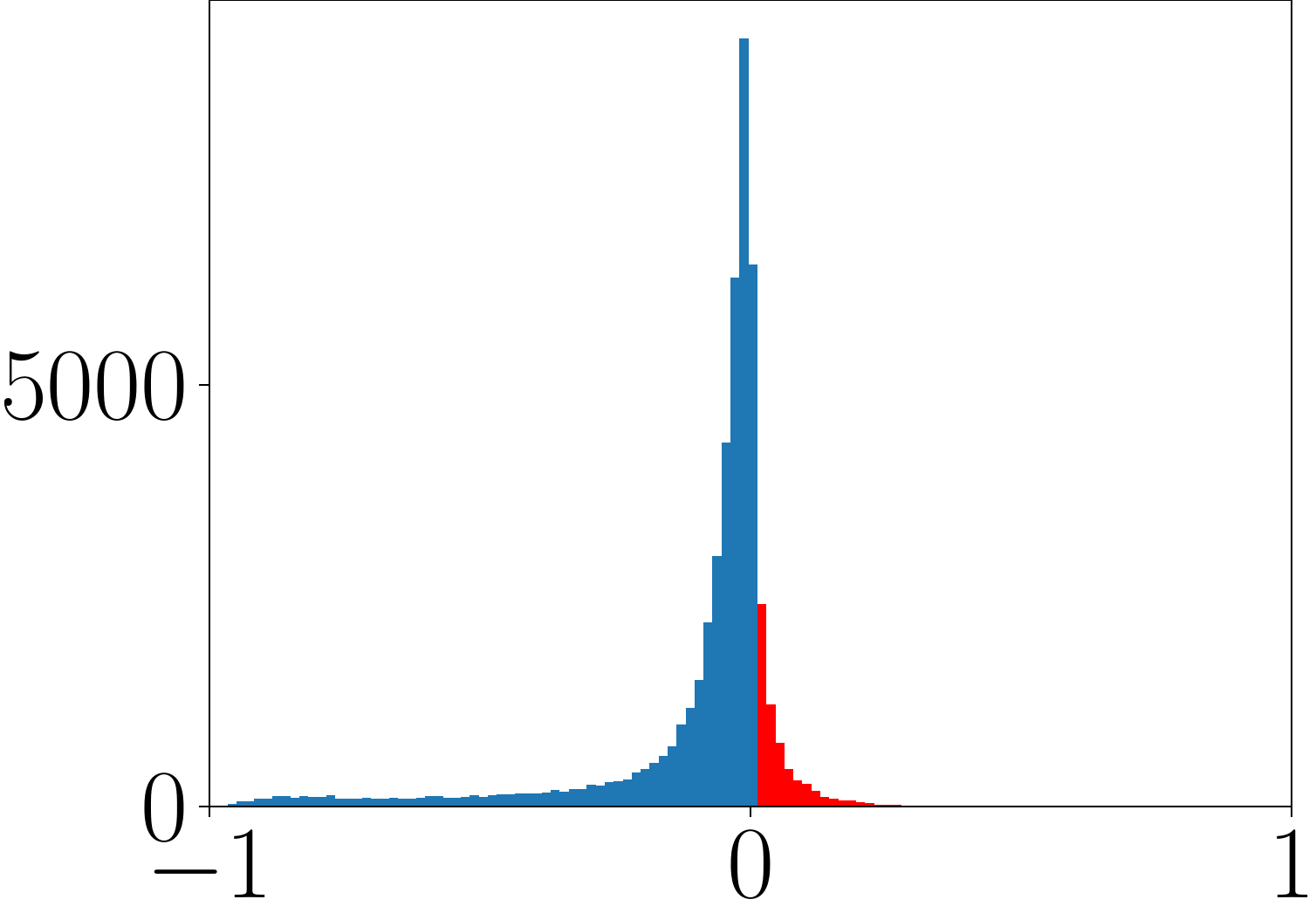}
     \caption{GAIRAT}\label{PMHist-GAIRAT}
   \end{subfigure}
   \begin{subfigure}[t]{0.16\linewidth}
     \includegraphics[width=\linewidth]{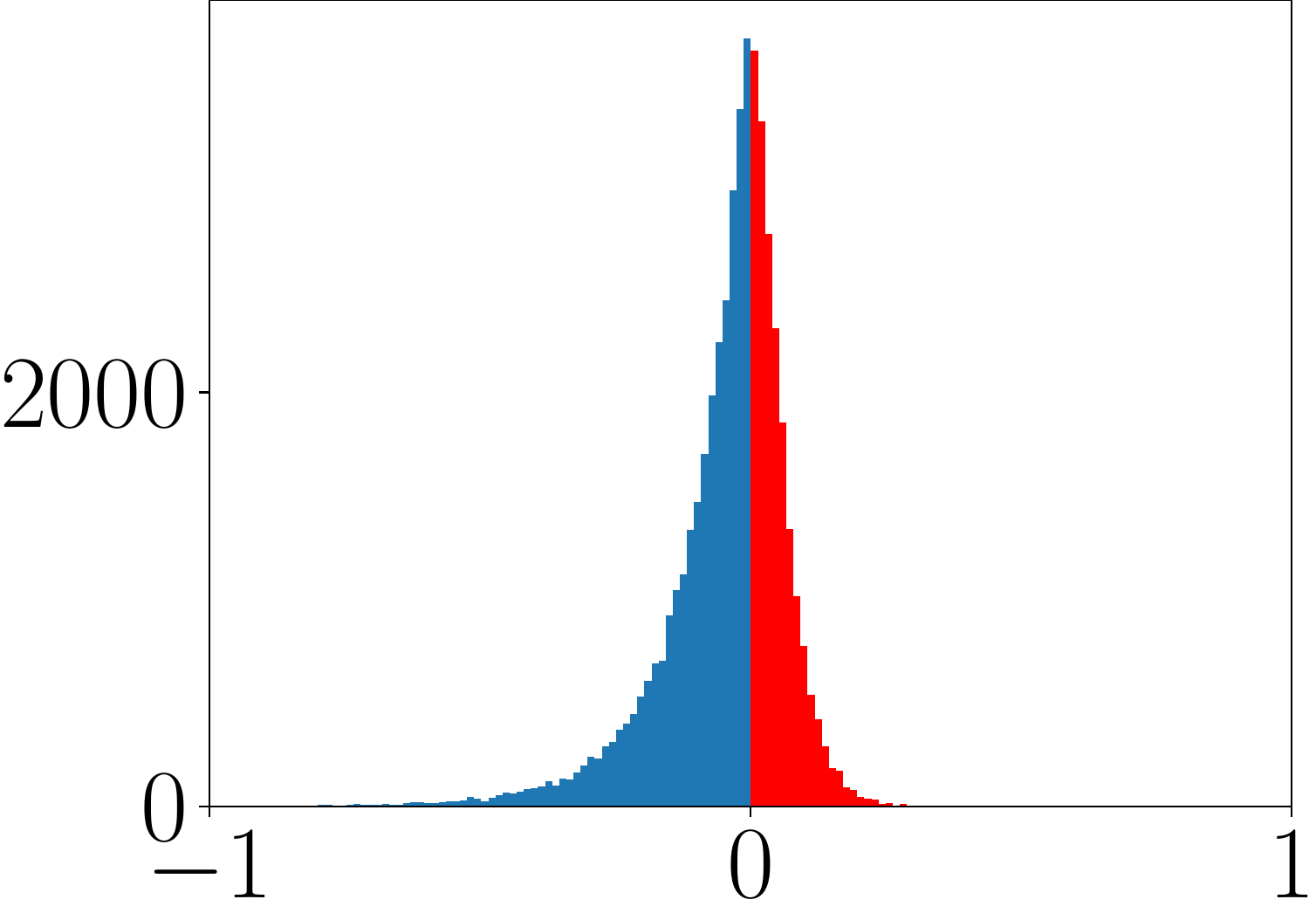}
     \caption{MAIL}\label{PMHist-MAIL}
   \end{subfigure}
   \begin{subfigure}[t]{0.16\linewidth}
     \includegraphics[width=\linewidth]{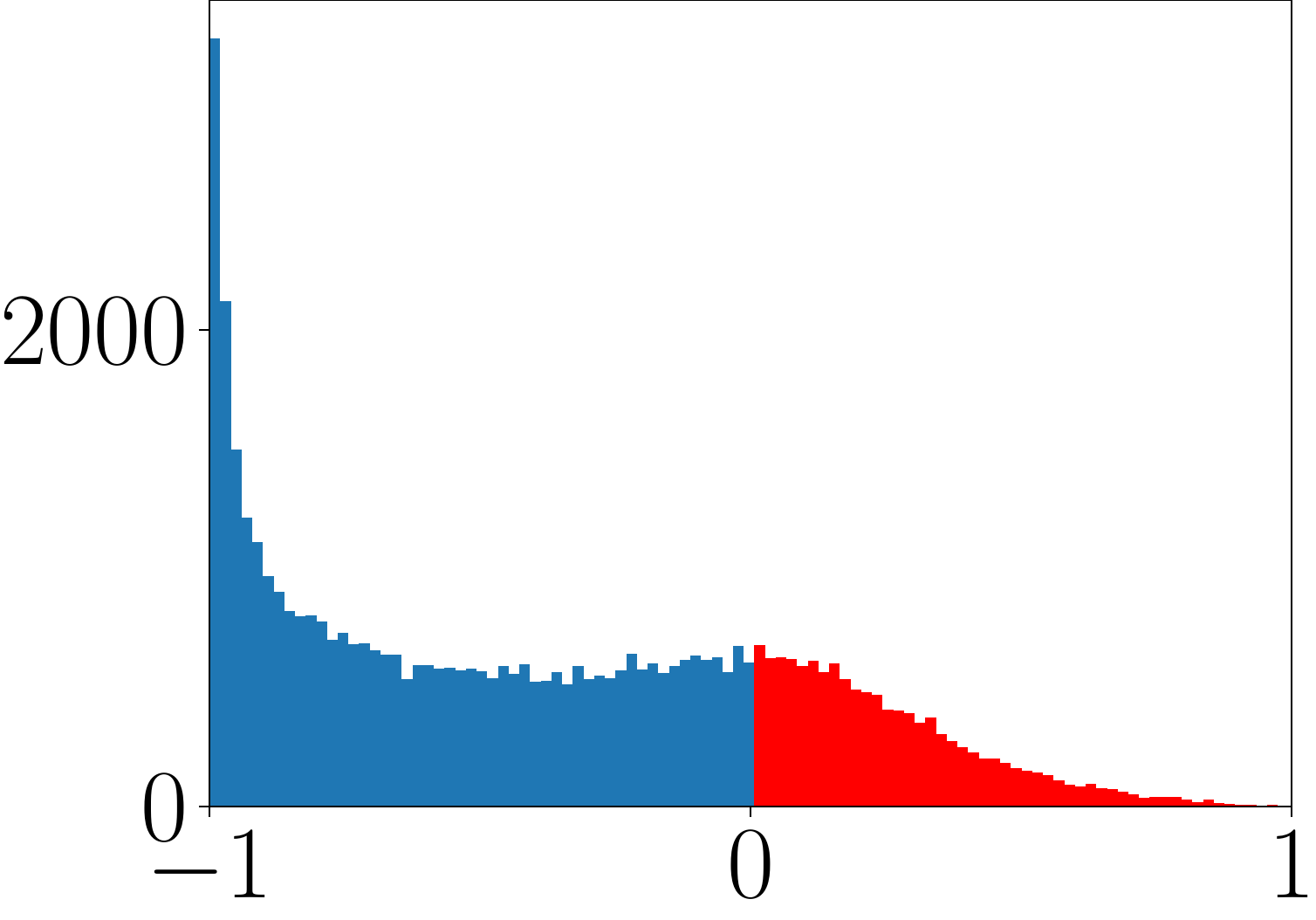}
     \caption{EWAT}\label{PMHist-EWAT}
   \end{subfigure}
   \begin{subfigure}[t]{0.16\linewidth}
     \includegraphics[width=\linewidth]{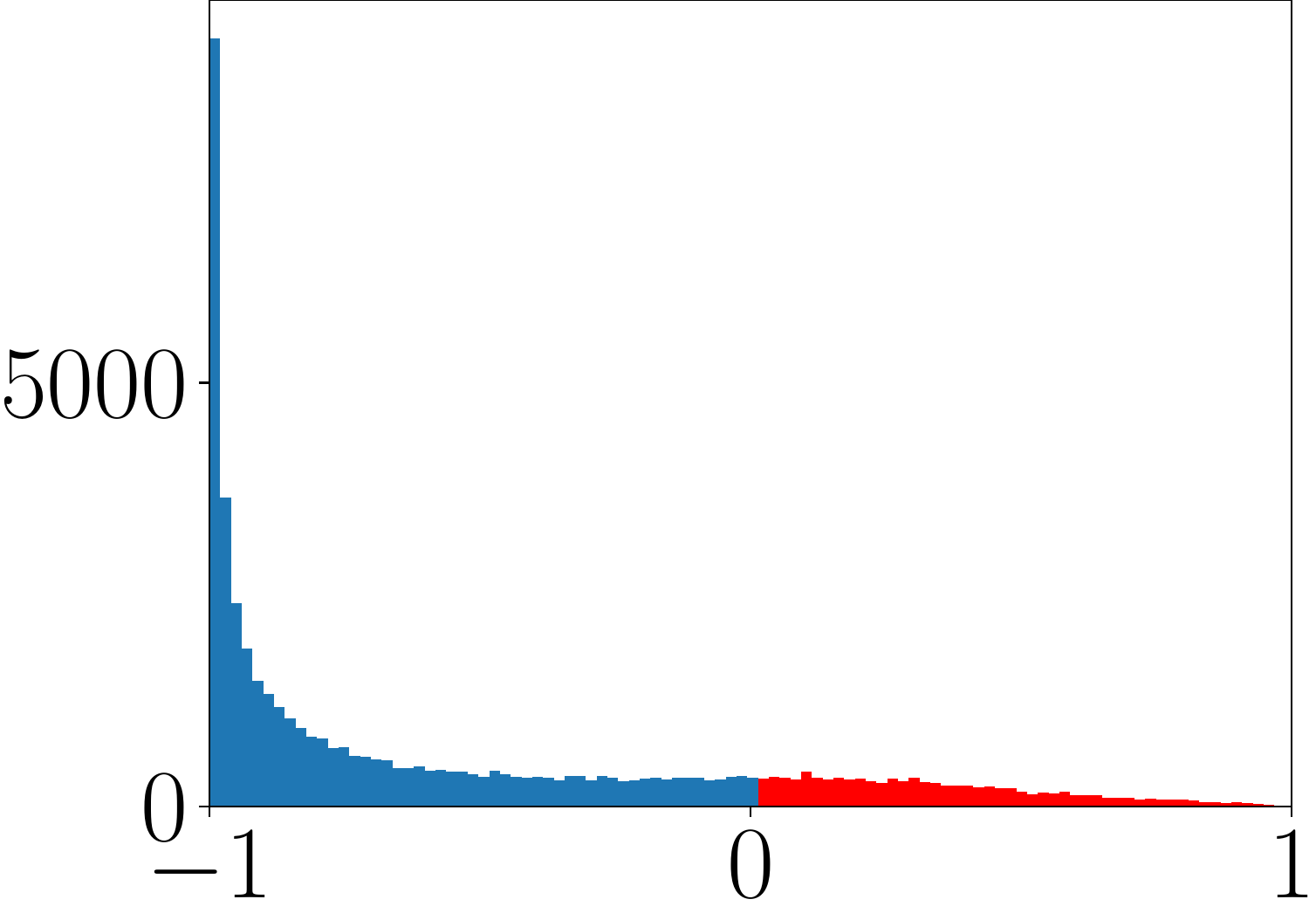}
     \caption{SOVR}\label{PMHist-SOVR}
   \end{subfigure}
   \caption{Histogram of probabilistic margin losses for training data of CIFAR10 with PreActResNet18 at the last epoch. ST denotes standard training, i.e., training on clean data.
     For standard training, we use $PM$ on clean data $\bm{x}$,
     while we plot that on adversarial examples $\bm{x}^\prime$ for the other methods. Blue bins correspond to the correctly classified data points,
     and red bins are misclassified samples.}
     \label{PMHist}
   \end{figure}

    While our study focuses on the logit margin loss, MAIL~\citep{MAIL} uses the probabilistic margin,
    \begin{align}
      \textstyle PM_n&\textstyle =f_{y_n}(\bm{x}_n^\prime,\bm{\theta})-\max_{k\neq {y_n}}f_k(\bm{x}_n^\prime,\bm{\theta}),
    \end{align}
    to evaluate the difficulty of data point.
    In the same way as Fig.~\ref{LMHist}, \rfig{PMHist} plots the histograms of probabilistic margins on CIFAR10 with PreActResNet18.
    Since softmax output is bounded in $[0,1]$, $PM$ is bounded in [-1,1]. As a result, most correctly classified data points concentrate near -1.
    In addition, softmax uses exponential functions, distributions of $PM$ are
    similar to the exponential distributions. Due to these effects, histograms of $PM$ make it more difficult to discover the fact 
    that there are two types of data points (easy samples and difficult samples).
    Since softmax preserves the order of logit, and a classifier infers the label by using the largest logit,
    the analysis by using $PM$ can under-estimate the distribution of difficult samples.
    Thus, logit margin losses are more suitable to empirically analyze trained models.
    Since softmax preserves the order of logit, the probabilistic margin can be used to determine $\mathbb{L}$ and $\mathbb{S}$ in SOVR.
 \end{document}